\documentclass[final]{cvpr}

\usepackage{times}
\usepackage{epsfig}
\usepackage{graphicx}
\usepackage{amsmath}
\usepackage{amssymb}

\usepackage{latexsym}
\usepackage{gensymb}
\usepackage[ruled]{algorithm2e}
\usepackage{amsfonts}       
\usepackage{adjustbox}
\usepackage{multirow}
\usepackage{booktabs}
\usepackage{nicefrac}       
\usepackage{microtype}      
\usepackage[T1]{fontenc}
\usepackage[utf8]{inputenc} 
\usepackage{comment}
\usepackage[caption=false]{subfig}
\usepackage{float}
\usepackage{makecell}
\usepackage{appendix}
\usepackage[hyphens]{url}

\usepackage[pagebackref=true,breaklinks=true,colorlinks,bookmarks=false]{hyperref}
\usepackage{cleveref}

\def\cvprPaperID{769} 
\def\confYear{CVPR 2021}

\begin{document}

\title{SSLayout360: Semi-Supervised Indoor Layout Estimation from 360{\degree} Panorama}

\author{Phi Vu Tran\\
Flyreel AI Research\\
{\tt\small vuptran@flyreel.co}}

\maketitle

\begin{abstract}
Recent years have seen flourishing research on both semi-supervised learning and 3D room layout reconstruction. In this work, we explore the intersection of these two fields to advance the research objective of enabling more accurate 3D indoor scene modeling with less labeled data. We propose the first approach to learn representations of room corners and boundaries by using a combination of labeled and unlabeled data for improved layout estimation in a 360{\degree} panoramic scene. Through extensive comparative experiments, we demonstrate that our approach can advance layout estimation of complex indoor scenes using as few as 20 labeled examples. When coupled with a layout predictor pre-trained on synthetic data, our semi-supervised method matches the fully supervised counterpart using only 12\% of the labels. Our work takes an important first step towards robust semi-supervised layout estimation that can enable many applications in 3D perception with limited labeled data.

\end{abstract}

\section{Introduction}
The task of inferring room layout from a single view 360{\degree} panoramic image has been gaining much attention from the computer vision community in the past several years. The problem addresses an important step towards holistic indoor scene understanding that can enable structured 3D modeling of the physical environment. Recent state-of-the-art methods \cite{horizonnet,dula-net,layoutnet} have made substantial progress towards accurate 3D room layout reconstruction by adopting large and powerful neural network architectures for representation learning. However, there are several challenges associated with acquiring vast quantities of high-quality room layout annotations to supervise deep neural networks. For one, it is difficult to consistently annotate cluttered scenes with ambiguous wall boundaries, especially in rooms with complex layouts that contain many corners. The lack of large-scale labeled data with precise layout annotations for panoramic scenes further hinders the progress of this important problem that has many pertinent applications in 3D computer vision.

\begin{figure}[t]
    \subfloat{%
        \includegraphics[width=\columnwidth]{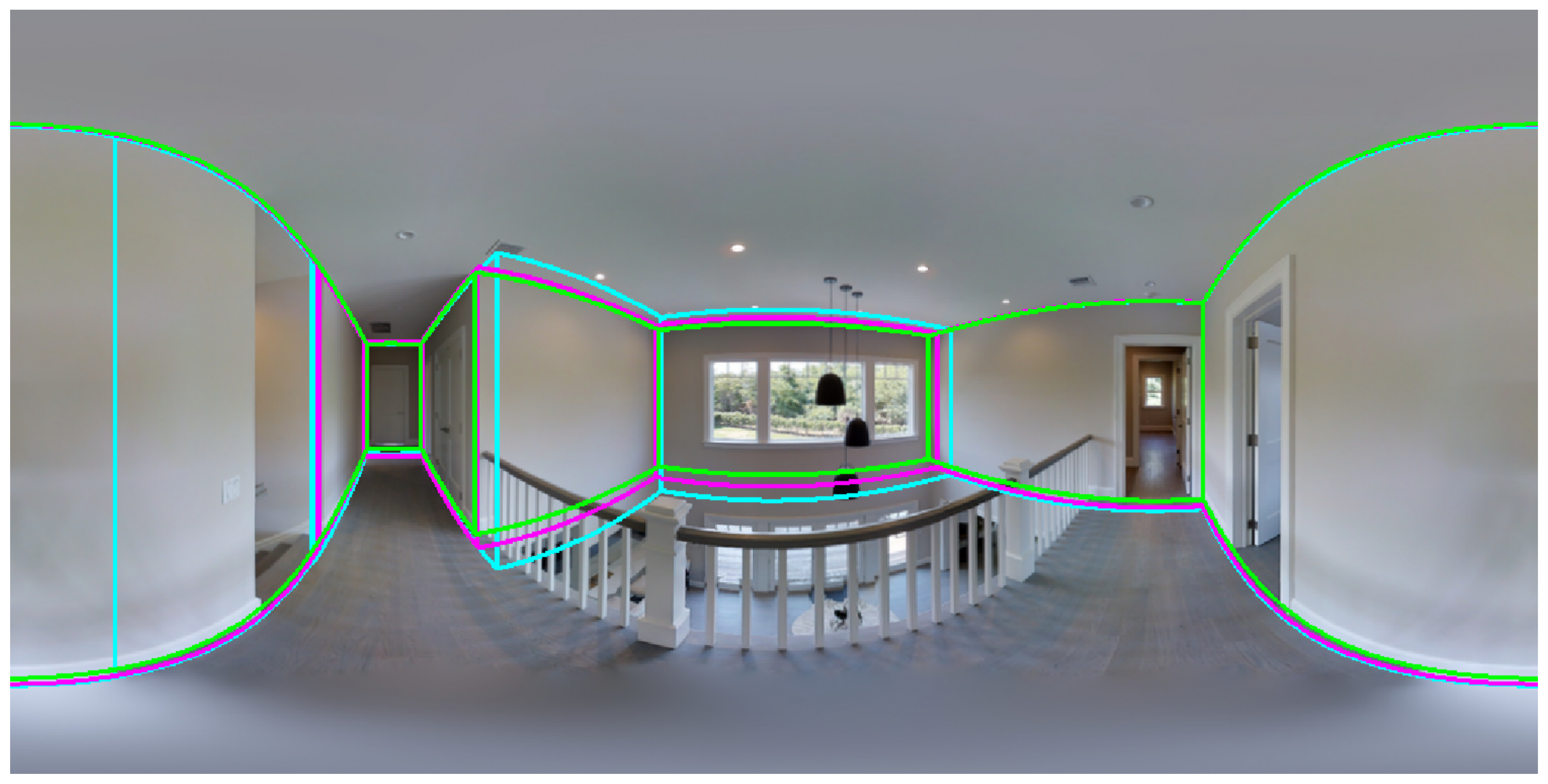}%
    }
    \\[-3ex]
    \subfloat{%
        \includegraphics[width=\columnwidth]{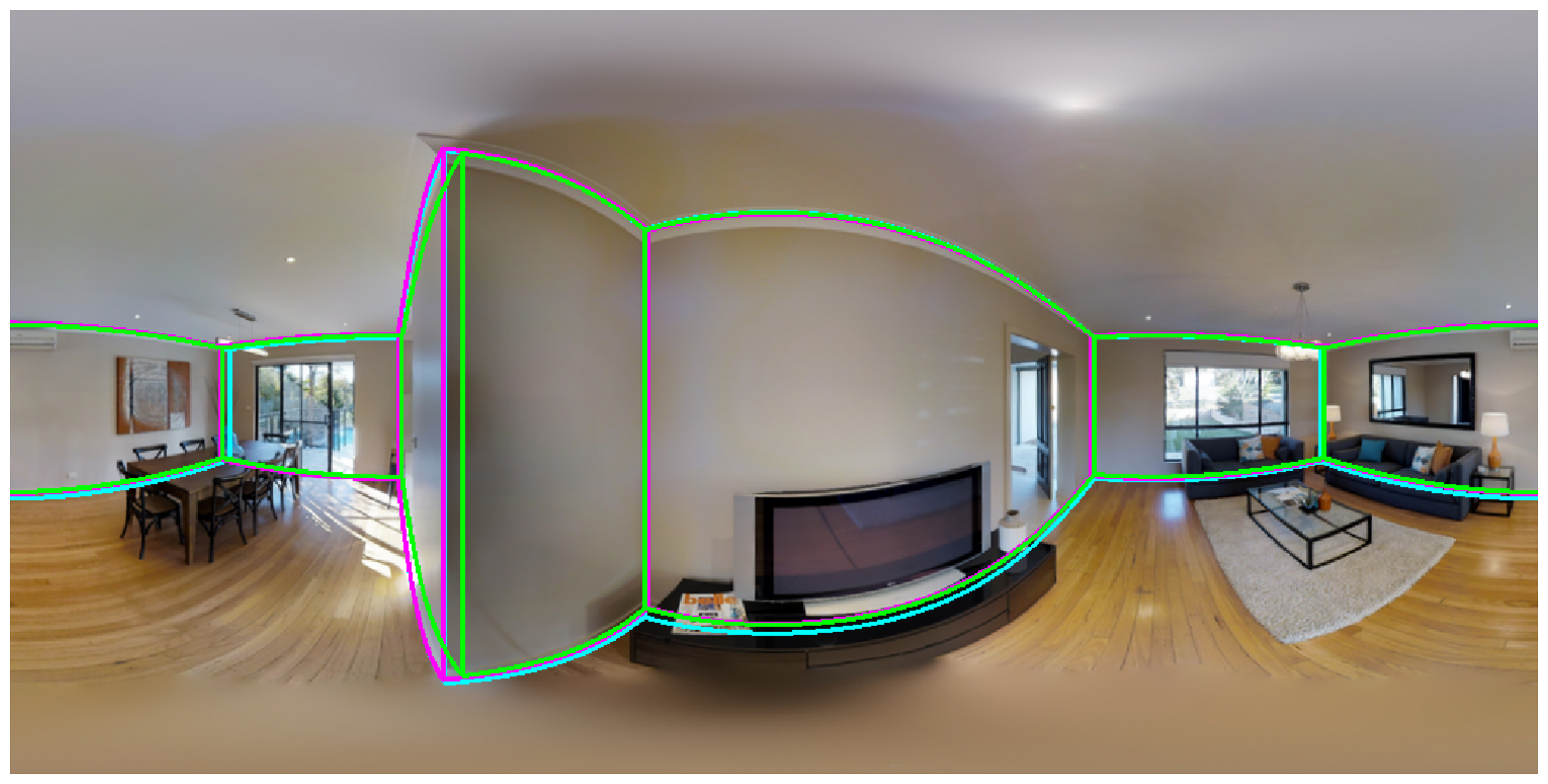}%
    }
    \caption{The effective use of unlabeled data improves complex 3D layout estimation with limited labels. We compare predicted layout boundary lines from a state-of-the-art supervised model \cite{horizonnet} trained on 1,650 labels (cyan) with our proposed semi-supervised model (magenta) and show that our model's predictions follow more closely to the ground truth (green lines) using just 100 labels.}
    \label{teaser}
\end{figure}

At the same time, recent years have seen flourishing research on deep semi-supervised learning (SSL) \cite{remixmatch,mean-teacher,s4l} that can leverage abundant unlabeled data for enhanced learning and generalization in the limited labeled data setting. The success of deep SSL has been mostly demonstrated on the relatively simple task of image classification, where the labeling procedure is the binary indication for the presence or absence of an object class. To our knowledge, the principles of SSL have not been studied in conjunction with room layout estimation, a complex and challenging task that depends on fine-grained human annotations, which presents an effective and efficient opportunity to improve learning with few annotated examples, as illustrated in \Cref{teaser}. In this work, we propose to explore and evaluate the potential contribution of unlabeled data for 3D room layout estimation, with the goal of promoting directed research towards the intersection of semi-supervised learning and 3D perception.

\medskip

\noindent
\textbf{Summary of Contributions} ~ We present SSLayout360, a neural architecture capable of learning representations of floor-wall, ceiling-wall, and wall-wall boundaries from a combination of labeled and unlabeled data for improved room layout estimation in a 360{\degree} panoramic scene. Our work is the first substantive attempt at semi-supervised 3D layout reconstruction of complex indoor scenes using as few as 20 labeled examples.

Through extensive comparative experiments, we show SSLayout360 achieves new state-of-the-art results on a number of benchmarks for both simple and complex room layouts. Coupled with a layout predictor pre-trained on synthetic data, SSLayout360 matches the best-in-class fully supervised baseline using only 12\% of the required labels.

We establish the first comprehensive set of semi-supervised benchmarks to measure the contribution of unlabeled data for indoor layout estimation. As part of this contribution, we propose a rigorous evaluation protocol to encourage the use of error bounds as standard practice and demonstrate the utility of unlabeled data across many experimental settings. We hope that our results serve as a strong baseline to inspire future research towards even more robust semi-supervised 3D layout reconstruction.

\section{Related Work}
\noindent
\textbf{Deep Semi-Supervised Learning} ~
The overarching goal of SSL is to make effective use of unlabeled data, without relying on any human supervision, to augment conventional supervised learning where labeled training data is scarce \cite{ssl}. One set of approaches involves pre-training neural models on large-scale unlabeled data, by way of unsupervised \cite{pretrain1,pretrain2} or self-supervised \cite{rotations,revisit-ssl} representation learning, followed by supervised fine-tuning on downstream tasks with limited ground truth information (\eg, detection, segmentation).

Another set of methods produces proxy targets for unlabeled data to be jointly trained end-to-end with ground truth labels. The training protocol for this class of SSL algorithms imposes an additional loss term to \emph{regularize} the objective function of the supervised algorithm. Recent examples of self-supervised regularization \cite{tran-ssl,s4l} improve supervised image classification performance by jointly training with an auxiliary self-supervised loss component based on the pretext task of image rotation recognition.

A third set of SSL methods based on consistency regularization \cite{pea,consistency} largely follows the student-teacher framework \cite{distillation}. As the student, the model learns from labeled data in the conventional supervised manner. As the teacher,
it generates soft unsupervised targets by enforcing consistent ensembles of predictions on unlabeled training samples under random perturbations. The consistency constraint encourages the student to learn representations from unlabeled data for enhanced SSL. Existing research on student-teacher SSL formulates clever ways to generate good unsupervised targets. Rasmus \etal~\cite{ladder-net} showed the effectiveness of random noise in regularizing the targets. Miyato \etal~\cite{vat2,vat} further explored this idea and adopted adversarial noise as an implicit teacher to improve the quality of the targets. Laine and Aila \cite{tempens} reduced teacher prediction variance by using an exponential moving average (EMA) to accumulate the predictions over training epochs. Tarvainen and Valpola \cite{mean-teacher} used an EMA of model weights to obtain an explicit ``mean teacher'', a simple but effective approach that was shown to achieve among the best SSL performances for image classification \cite{ssl-eval}. More recent extensions of the student-teacher framework have been demonstrated to surpass state-of-the-art fully supervised baselines using a fraction of the required labeled examples \cite{remixmatch,mixmatch}.

\medskip

\noindent
\textbf{3D Room Layout Estimation} ~
Room layout reconstruction has been an active research topic for over a decade \cite{survey}, dating back to Delage \etal~\cite{delage} fitting floor-wall boundaries in a perspective image under ``Manhattan world'' assumptions \cite{manhattan-world}. In this paper, we focus our review on modern, state-of-the-art approaches that recover room layout from a single RGB panorama represented in equirectangular projection covering a 360{\degree} horizontal and 180{\degree} vertical field-of-view.

Existing methods for layout estimation include PanoContext~\cite{panocontext}, LayoutNet~\cite{layoutnet}, DuLa-Net~\cite{dula-net}, CFL~\cite{CFL}, and HorizonNet~\cite{horizonnet}. These methods were tested on datasets with strong Manhattan assumptions, \ie, the wall-wall boundaries form right angles and are orthogonal to the horizontal floor plane. The recent work of AtlantaNet~\cite{atlantanet} is not constrained to Manhattan scenes, and can recover room layout with walls that do not form right angles or are curved. All of these methods utilize a deep encoder-decoder neural network to predict layout elements, such as floor-wall and ceiling-wall boundaries and corner positions (LayoutNet, CFL, HorizonNet) or a semantic 2D floor plan in the ceiling view (DuLa-Net and AtlantaNet), and then fit the predicted elements to a 3D layout via a post-processing step \cite{layoutnetv2}.

Although the aforementioned methods have made substantial contributions towards accurate layout estimation, these models have all been trained in a fully supervised manner, on relatively small sets of annotated layout examples, while leaving an abundant amount of unlabeled panoramic indoor images completely untapped. While effective data augmentation strategies, such as panoramic horizontal rotation, horizontal flipping, and the recently introduced Pano Stretch \cite{horizonnet}, have been used to improve supervised learning, their utilization in conventional supervised training cannot exploit unlabeled data in a principled way.

Recent work in \cite{interiornet,structured3d} proposed the use of well-crafted, photo-realistic synthetic data with detailed ground truth structure annotations as a measure to alleviate costly hand-labeling efforts. The physically-based rendering process requires data modeling with extensive domain expertise, but generating synthetic data at scale is cheaper than the traditional approach of collecting, curating, and annotating panoramas from the real world. While synthetic data offers many advantages, there is still a challenge to learn transferable representations between real and synthetic domains in order to overcome the dataset bias \cite{domain-gap}.

In this work, we bridge the gap by combining unlabeled data with available labeled data from both real and synthetic contexts in a semi-supervised setting to further push the performance envelope of 3D room layout reconstruction.

\begin{figure}[t]
\centering
\includegraphics[width=\columnwidth]{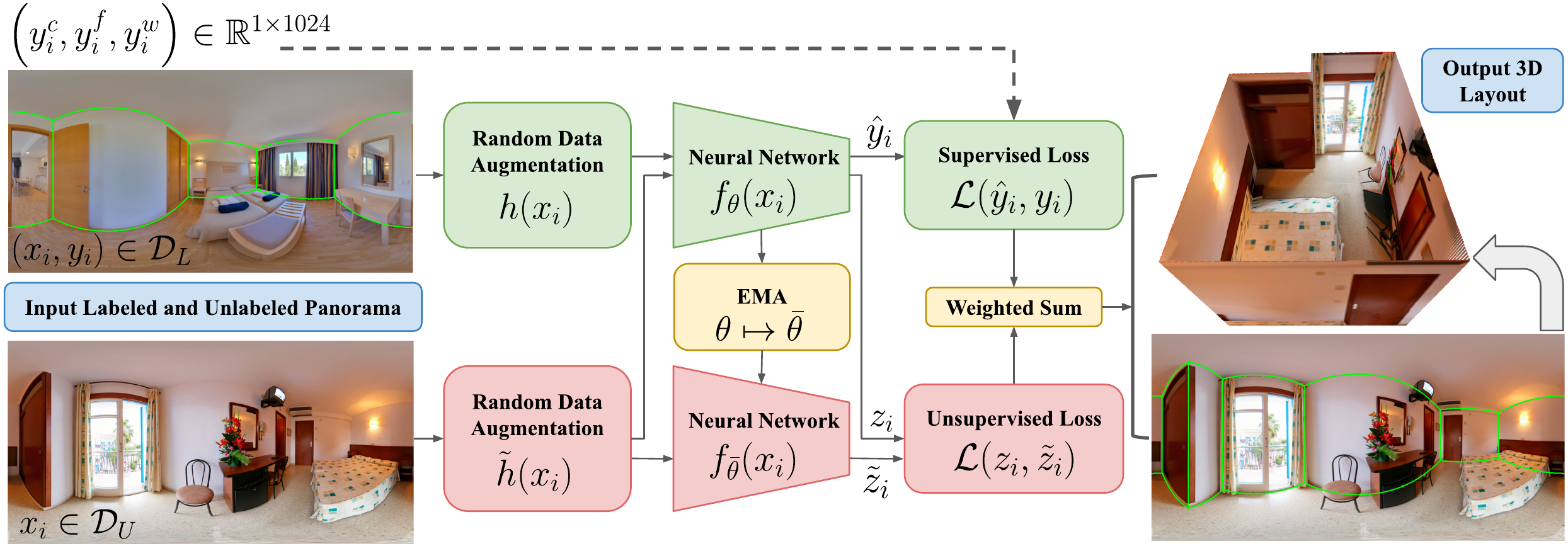}
\caption{An illustration of the SSLayout360 architecture for semi-supervised indoor layout estimation from a 360{\degree} panoramic scene.}
\label{schematic}
\end{figure}

\section{Approach}
The goal of this work is to learn 3D room layout from a single view 360{\degree} RGB panorama in the semi-supervised setting. The design and algorithmic overview of SSLayout360 are depicted in \Cref{schematic} and \Cref{algorithm}, with more details given in \Cref{schematic-appendix} and Section \ref{perturbation} of the appendix. The input is a set of labeled input-target pairs $(x_l,y_l) \in \mathcal{D}_L$ and a set of unlabeled examples $x_u \in \mathcal{D}_U$. As is common in prior work, we assume $\mathcal{D}_L$ and $\mathcal{D}_U$ are sampled from the same underlying data distribution (\eg, indoor scenes), in which case $\mathcal{D}_L$ is a labeled subset of $\mathcal{D}_U$. In real-world applications, however, $\mathcal{D}_L$ and $\mathcal{D}_U$ often come from different, but somewhat related, data distributions (\eg, indoor $+$ outdoor panoramic scenes), and it is desirable for the SSL algorithm to appropriately learn from such distribution mismatch.

We train a neural network $f_\theta (x)$, a stochastic prediction function parametrized by $\theta$, to learn room layout boundaries by using a combination of $\mathcal{D}_L$ and $\mathcal{D}_U$. Our work builds on a successful, state-of-the-art variant of the student-teacher model, Mean Teacher \cite{mean-teacher}, originally formulated for image classification and extends it to the more challenging task of layout estimation from complex panoramic indoor scenes. We consider the following compound objective for SSL:
\begin{equation}
\label{canonical}
\min_\theta \mathcal{L}_l (\mathcal{D}_L, \theta) + \lambda\mathcal{L}_u(\mathcal{D}_U, \theta),
\end{equation}
\noindent
where $\mathcal{L}_l$ is the supervised loss over labeled examples and $\mathcal{L}_u$ is the unsupervised loss defined for unlabeled data. The learning objective treats $\mathcal{L}_u$ as a regularizer, and $\lambda > 0$ is a hyper-parameter controlling the strength of regularization.

\begin{algorithm}[t]
\caption{SSLayout360 training procedure.}
\label{algorithm}
\DontPrintSemicolon
\SetAlgoNoLine
\begin{flushleft}
\nl \textbf{Input:} Training set of labeled inputs $(x_l, y_l) \in \mathcal{D}_L$. \\
\nl Training set of unlabeled inputs $x_u \in \mathcal{D}_U$. \\
\nl Data augmentation functions $h(x)$ and $\smash{\tilde{h}(x)}$. \\
\nl Student network $f_\theta(x)$ with trainable parameters $\theta$. \\
\nl Teacher network $f_{\bar{\theta}}(x)$ with parameters $\bar{\theta} = \theta$. \\
\nl Distance function $d$ (\eg, $L_1$ and $L_2$).
\end{flushleft}
\vspace{-8pt}
\nl \textbf{for} \emph{each epoch over} $\mathcal{D}_U$ \textbf{do} \\[2pt]
\nl $~~ b_l \hspace{2pt} \leftarrow h\left(x_l\right) \hspace{2pt} \quad \triangleright ~ \text{\small Mini-batches of labeled input.}$ \\[2pt]
\nl $~~ b_u \leftarrow \tilde{h}\left(x_u\right) \quad \triangleright ~ \text{\small Mini-batches of unlabeled input.}$ \\[5pt]
\nl ~ \textbf{for} \emph{each mini-batch} \textbf{do} \\[2pt]
\nl    $\quad \hat{y}_l \hspace{2pt} \leftarrow f_\theta\left(b_l\right) \quad \hspace{-1.5pt} \triangleright ~ \text{\small Forward pass on labeled input.}$ \\[2pt]
\nl    $\quad z_u \leftarrow f_\theta\left(b_u\right) ~~ \triangleright ~ \text{\small Forward pass on unlabeled input.}$ \\[2pt]
\nl    $\quad \tilde{z}_u \leftarrow f_{\bar{\theta}}\left(b_u\right) ~~ \triangleright ~ \text{\small Again using $\bar{\theta}$.}$ \\[2pt]
\nl    $\quad \mathcal{L} \leftarrow \frac{1}{\vert b_l \vert}\sum_{i \in b_l}d(\hat{y}_{il}, y_{il}) \quad \hspace{2pt} \triangleright ~ \text{\small Supervised loss.}$ \\[2pt]
\nl    $\qquad ~ + \frac{\lambda}{\vert b_u \vert}\sum_{i \in b_u}d(z_{iu}, \tilde{z}_{iu}) ~ \triangleright ~ \text{\small Unsupervised loss.}$ \\[2pt]
\nl    $\quad \lambda \leftarrow e^{-5 \left(1 - T\right)^2} ~~ \triangleright ~ \text{\small Ramp up $\lambda$ for $T \in [0,1]$}.$ \\[2pt]
\nl    $\quad \theta \leftarrow \theta - \nabla_\theta\mathcal{L} \quad \triangleright ~ \text{\small Update $\theta$ via gradient descent.}$ \\[2pt]
\nl    $\quad \bar{\theta} \leftarrow \alpha \bar{\theta} + (1 - \alpha)\theta \quad  \triangleright ~ \text{\small Update $\bar{\theta}$ via EMA.}$ \\[2pt]
\nl ~ \textbf{end} \\[2pt]
\nl \textbf{end} \\[2pt]
\nl \Return{$\theta, \bar{\theta}$}
\end{algorithm}

\subsection{Semi-Supervised Layout Estimation}
The task of room layout estimation essentially boils down to inferring the floor-wall and ceiling-wall boundaries and wall-wall (or corner) positions. In this work, we derive insight from HorizonNet \cite{horizonnet} to regress layout boundaries and corners to the ground truth for each column of the input image in the semi-supervised setting. In principle, other layout prediction methods based on pixel-wise classification (\eg, LayoutNet, AtlantaNet) could be extended to the semi-supervised setting. But a comparative investigation of alternative layout prediction methods under the semi-supervised setting is beyond the scope of this paper, and would be an interesting research direction for future work.

\medskip

\noindent
\textbf{HorizonNet} ~ We choose HorizonNet as our prediction function $f_\theta(x)$ for its simplicity, efficient computation, and state-of-the-art performance on room layout estimation. The input to HorizonNet is an RGB panorama with shape $3 \times 512 \times 1024$ (for channel, height, width) along with a 3-channel target vector of size $3 \times 1 \times 1024$ representing the ceiling-wall ($y_c$), floor-wall ($y_f$), and wall-wall ($y_w$) boundary position of each image column. The values of $y_c$ and $y_f$ are normalized in $[-\pi/2, \pi/2]$, and $y_w$ is scaled to $[0,1]$.

HorizonNet follows an encoder-decoder approach to learn whole-room layout from a panoramic scene, similar to other competing methods. The encoder is the ResNet-50 architecture \cite{resnet} pre-trained on ImageNet \cite{imagenet}, combined with a sequence of convolution layers followed by ReLU \cite{relu} activation, to compute an abstract $1024 \times 1 \times 256$ dimensional feature representation from the input image. The decoder is a bidirectional recurrent neural network \cite{bi-rnn} that predicts $(\hat{y}_c$, $\hat{y}_f$, $\hat{y}_w) \in \mathbb{R}^{1\times1024}$ column by column. Next, we formulate HorizonNet as a student-teacher model, and describe the resulting SSLayout360 architecture as a semi-supervised learner for 3D layout reconstruction.

\medskip

\noindent
\textbf{SSLayout360} ~ Our approach treats HorizonNet as a stochastic predictor with the dual role of being both the student and teacher. Given a batch $b_l$ of labeled examples, and their real-valued target vectors $y_l \in \mathbb{R}^{3\times1\times1024}$, and a batch $b_u$ of unlabeled examples at each training step, we forward propagate HorizonNet three times: (1) on the batch of labeled examples as the student $f_\theta(x)$ using parameters $\theta$ to produce real-valued prediction vectors $\hat{y}_l \in \mathbb{R}^{3\times1\times1024}$, (2) on the unlabeled batch using the same parameters $\theta$ to compute $z_u \in \mathbb{R}^{3\times1\times1024}$, and (3) on the unlabeled batch as the teacher $f_{\bar{\theta}}(x)$ using parameters $\bar{\theta}$ to output $\tilde{z}_u \in \mathbb{R}^{3\times1\times1024}$. Here, $\bar{\theta}$ is an exponential moving average (EMA) of the student's parameters $\theta$ after each training step $t$:
\begin{equation}
    \bar{\theta}_t = \alpha \bar{\theta}_{t-1} + (1 - \alpha)\theta_t,
\end{equation}
where $\alpha \in [0,1]$ is a decay hyper-parameter. The intuition for setting $\bar{\theta} = \textsc{EMA}(\theta)$ is to obtain a good teacher that provides stable unsupervised targets for the student to imitate, and was the main result of Mean Teacher. As is common practice, we do not back-propagate gradients through the teacher and keep its prediction fixed at each training step \cite{mixmatch,mean-teacher}. The alternative case is to set $\bar{\theta} = \theta$ with $\alpha=0$ and back-propagate gradients through both student and teacher models, which was the formulation of $\Pi$ model \cite{tempens}, and has been shown to produce less stable unsupervised targets and overall inferior SSL performance to Mean Teacher \cite{mean-teacher}. \Cref{ablation} provides an ablation experiment where we evaluate both settings for comparative semi-supervised layout estimation.

\subsection{Loss Function}
The real-valued prediction vectors $\hat{y}_l$ are regressed to target vectors $y_l$ using $L_1$ distance for $(y_c,y_f)$ and squared $L_2$ distance for $y_w$. The supervised loss component, evaluated over a mini-batch of labeled examples, is computed as:
\begin{equation}
\label{supervised-loss}
    \mathcal{L}_l = \frac{1}{|b_l|}\sum_{i \in b_l} \lVert\hat{y}_{ic} - y_{ic}\rVert_1
    + \lVert\hat{y}_{if} - y_{if}\rVert_1
    + \lVert\hat{y}_{iw} - y_{iw}\rVert_2^2.
\end{equation}
\indent Our supervised objective is different from HorizonNet, in that we use the squared $L_2$ loss, or the Brier score \cite{brier}, instead of binary cross-entropy loss for the wall-wall corner $y_w$. The Brier score is commonly used in the SSL literature because it is bounded and does not heavily penalize predicted probabilities far away from the ground truth \cite{mixmatch,tempens,mean-teacher}. Our initial experiments showed that the squared $L_2$ loss gave slightly better accuracy performance than cross-entropy loss.

For the unsupervised loss component, we constrain $z_u$ and $\tilde{z}_u$ to be close by computing their $L_1$ and squared $L_2$ distances over a mini-batch of unlabeled examples, similar to \Cref{supervised-loss}:
\begin{equation}
\label{unsupervised-loss}
    \mathcal{L}_u = \frac{1}{|b_u|}\sum_{i \in b_u} \lVert z_{ic} - \tilde{z}_{ic} \rVert_1
    + \lVert z_{if} - \tilde{z}_{if} \rVert_1
    + \lVert z_{iw} - \tilde{z}_{iw} \rVert_2^2.
\end{equation}
\indent It is a reasonable objective to enforce consistency on $z_u$ and $\tilde{z}_u$ because $f_\theta(x)$ and $f_{\bar{\theta}}(x)$ are stochastic predictors with random dropout \cite{dropout} and input data augmentation at each forward pass. By minimizing the discrepancy between student $z_u$ and teacher $\tilde{z}_u$ predictions on unlabeled instances, we encourage the student to learn additional layout representations from unlabeled data via the unsupervised targets provided by the teacher. The compound objective for training SSLayout360 on both labeled and unlabeled data is the weighted sum of the supervised and unsupervised losses:
\setlength{\belowdisplayskip}{5pt} \setlength{\belowdisplayshortskip}{5pt}
\setlength{\abovedisplayskip}{5pt} \setlength{\abovedisplayshortskip}{5pt}
\begin{equation}
    \label{loss}
    \mathcal{L} = \mathcal{L}_l + \lambda\mathcal{L}_u.
\end{equation}

Our formulation of the SSLayout360 objective works well for layout estimation and stands out from previous SSL methods in that we \emph{maintain compatibility} between the supervised and unsupervised terms by applying $L_1$ and $L_2$ losses to both. This has the intended benefit of removing the need to tune the weight hyper-parameter $\lambda$ during training; we simply set $\lambda = 1$ to obtain reliable results in all experiments across all datasets under consideration. By contrast, Mean Teacher and other SSL methods used different loss functions for the supervised and unsupervised terms (\eg, cross entropy $+$ $L_2$), resulting in the need to carefully tune $\lambda$ to manage the balance between the two objectives.

\begin{figure*}[t]
    \begin{minipage}[t]{0.33\textwidth}
        \subfloat{%
            \includegraphics[width=\columnwidth]{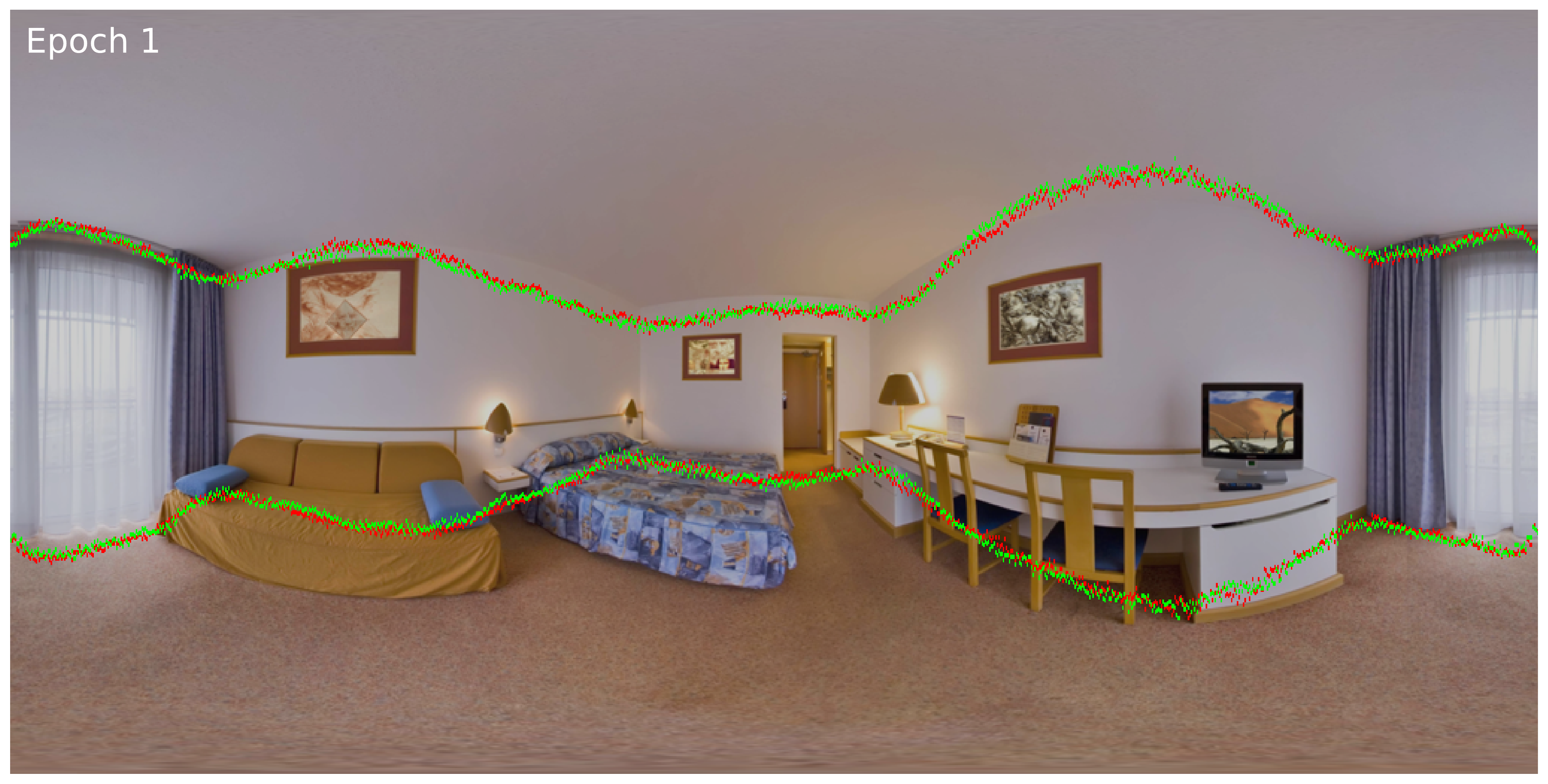}
        }
    \end{minipage}%
    \hfill
    \begin{minipage}[t]{0.33\textwidth}
        \subfloat{%
            \includegraphics[width=\columnwidth]{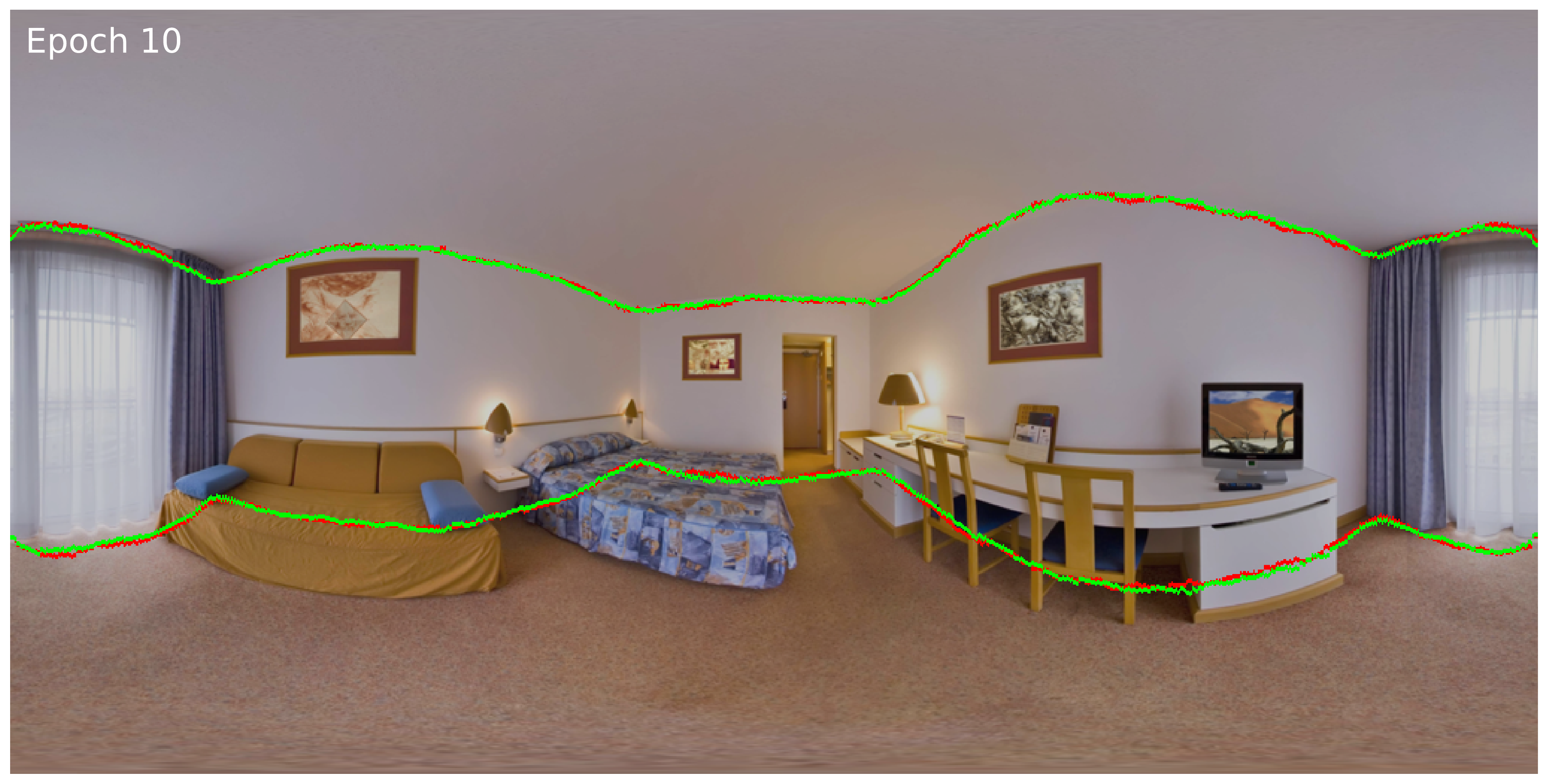}
        }
    \end{minipage}%
    \hfill
    \begin{minipage}[t]{0.33\textwidth}
        \subfloat{%
            \includegraphics[width=\columnwidth]{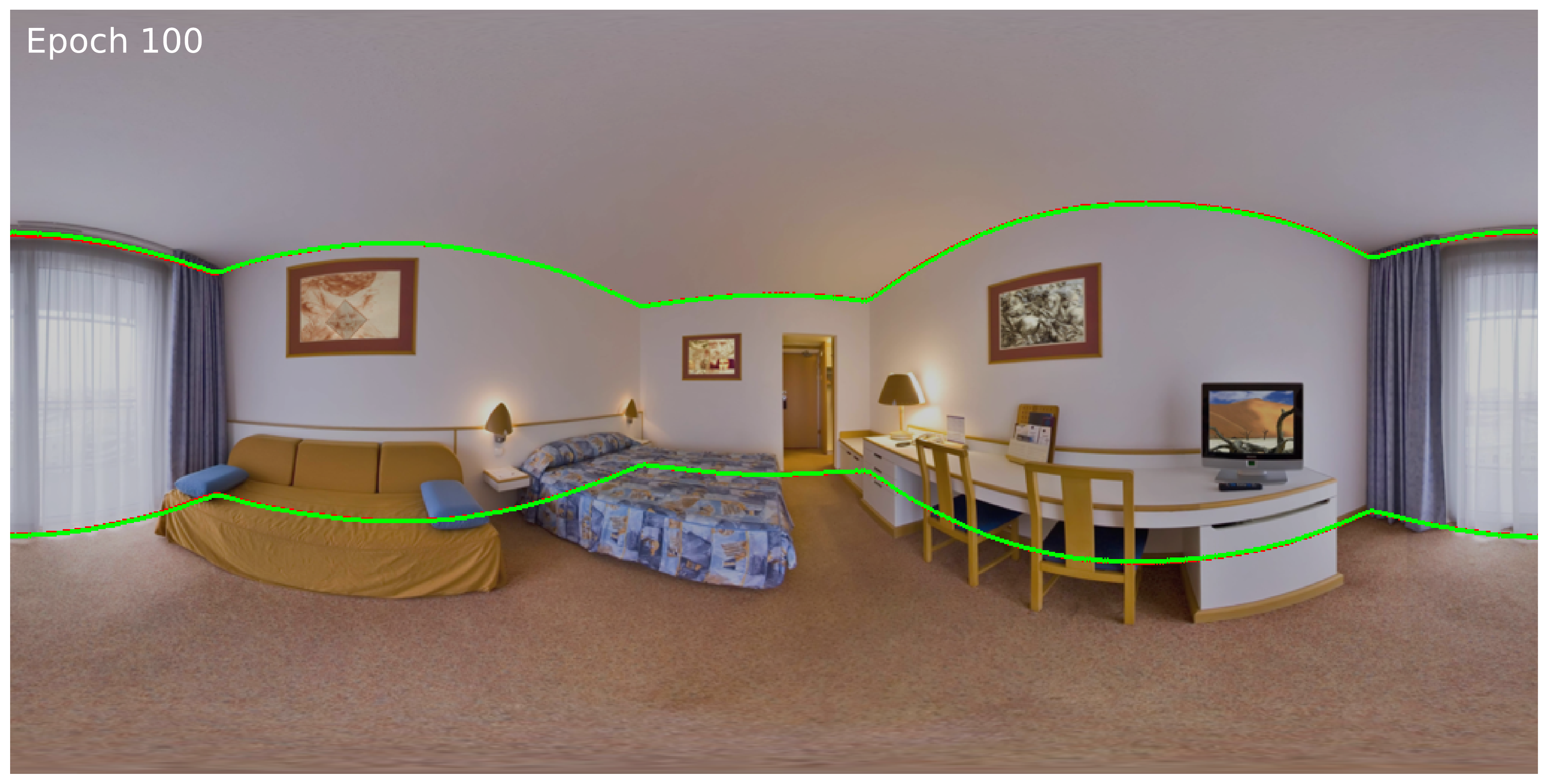}
        }
    \end{minipage}%
    \smallskip
    \caption{The evolution of HorizonNet as both the student (red lines) and teacher (green lines) over the course of training on 100 labeled and 1,009 unlabeled images. We encourage the student to learn layout representations of floor, ceiling, and wall boundaries from unlabeled data by enforcing consistency (or minimizing discrepancy) between student and teacher predictions perturbed by random model and data noise.}
    \label{evolution}
\end{figure*}

Our approach to SSLayout360 assumes that the teacher provides good unsupervised targets for the student to imitate. As illustrated in \Cref{evolution}, at the beginning of model training, both student and teacher models are likely to produce incorrect and inconsistent predictions, especially when few labels are available. Similar to SSL for image classification \cite{tempens,mean-teacher}, we mitigate a potentially degenerate solution by ramping up the unsupervised loss weight from 0 to 1 according to the sigmoid-shaped function: $\lambda (t) = e^{-5 (1 - \nicefrac{t}{T})^2}$, where $t$ is the current training iteration and $T$ is the number of iterations at which to stop the ramp-up. We define $T$ to be a percentage of the maximum number of iterations, which is the product of training epochs and the cardinality of mini-batches in the unlabeled dataset. For example, $T@30\%$ means that if we train SSLayout360 for 100 epochs over an unlabeled set of 2,000 images using a mini-batch of 4, then $T@30\% = 0.30 \times 100 \times 2000/4 = 15,000$ iterations. The ramp-up period ensures that student learning progresses predominantly via the supervised loss on labeled data at the beginning of training up to $T$ iterations, after which the teacher becomes reliable to provide good stable targets.

\section{Experimental Setup}
\subsection{Datasets}
We train and evaluate our SSLayout360 algorithm on three challenging benchmark datasets covering both simple cuboid and complex non-cuboid indoor layouts. For cuboid layout estimation, we use PanoContext \cite{panocontext} and Stanford 2D-3D \cite{stanford-2d3d}, which consist of 512 and 550 RGB panoramic images, respectively. For non-cuboid layout estimation, we train and evaluate on MatterportLayout \cite{layoutnetv2}, which comprise 2,295 RGB panoramic images of indoor scenes having up to 22 corners. We use the standard training, validation, and test splits provided by Zou \etal~\cite{layoutnet,layoutnetv2} for all three datasets. See Section \ref{data-details} in the appendix for details.

MatterportLayout is a labeled subset of the larger Matterport3D \cite{matterport3d} dataset comprising 10,912 panoramas of general indoor and outdoor environments. We use the auxiliary Matterport3D dataset (minus 458 test instances) of 10,454 panoramas as a source of extra unlabeled data to augment our SSL experiments on MatterportLayout, and to evaluate SSLayout360 under the condition of data distribution mismatch that includes a mixture of indoor and outdoor scenes.

We also experiment with Structured3D \cite{structured3d}, a large photo-realistic \emph{synthetic} dataset comprising 21,835 panoramas of rooms in 3,500 diverse indoor scenes with ground truth cuboid and non-cuboid layout annotations. We pre-train HorizonNet on 18,362 synthetic images and perform transfer learning on the MatterportLayout dataset via fine-tuning.

\begin{table*}[t]
\begin{minipage}[t]{0.51\textwidth}
\centering
\resizebox{\columnwidth}{!}{
	\begin{tabular} {lrrrrrr}
	\toprule
	\multicolumn{2}{l}{\textbf{PanoContext}}
	& \multicolumn{3}{c}{3D IoU (\%) $\uparrow$} \\
	\cmidrule{2-6}
	\multicolumn{1}{l}{\multirow{2}{*}{Method}} &
	\multicolumn{1}{c}{\multirow{1}{*}{~~~~20 labels}} &
	\multicolumn{1}{c}{\multirow{1}{*}{~~~~50 labels}} &
	\multicolumn{1}{c}{\multirow{1}{*}{~~~100 labels}} &
	\multicolumn{1}{c}{\multirow{1}{*}{~~~200 labels}} &
	\multicolumn{1}{c}{\multirow{1}{*}{~~~963 labels}} \\
    & \multicolumn{1}{c}{\multirow{1}{*}{~~1,009 images}}
    & \multicolumn{1}{c}{\multirow{1}{*}{~~1,009 images}}
    & \multicolumn{1}{c}{\multirow{1}{*}{~~1,009 images}}
    & \multicolumn{1}{c}{\multirow{1}{*}{~~1,009 images}}
    & \multicolumn{1}{c}{\multirow{1}{*}{~~1,009 images}} \\
    \midrule
    HorizonNet
                & $61.48 \pm 2.07$
				& $63.84 \pm 2.87$
				& $65.43 \pm 1.30$
				& $75.76 \pm 0.62$
				& \bfseries 83.55 $\pm$ 0.31 \\
	SSLayout360
                & \bfseries 63.05 $\pm$ 0.65
				& \bfseries 68.41 $\pm$ 1.02
				& \bfseries 72.86 $\pm$ 1.07
				& \bfseries 78.64 $\pm$ 0.72
				& \bfseries 83.30 $\pm$ 0.53 \\
	\end{tabular}
	}
\resizebox{\columnwidth}{!}{
	\begin{tabular} {lrrrrrr}
	\toprule
	& \multicolumn{5}{c}{Corner Error (\%) $\downarrow$} \\
	\cmidrule{2-6}
	\multicolumn{1}{l}{\multirow{2}{*}{Method}} &
	\multicolumn{1}{c}{\multirow{1}{*}{~~~~20 labels}} &
	\multicolumn{1}{c}{\multirow{1}{*}{~~~~50 labels}} &
	\multicolumn{1}{c}{\multirow{1}{*}{~~~100 labels}} &
	\multicolumn{1}{c}{\multirow{1}{*}{~~~200 labels}} &
	\multicolumn{1}{c}{\multirow{1}{*}{~~~963 labels}} \\
    & \multicolumn{1}{c}{\multirow{1}{*}{~~1,009 images}}
    & \multicolumn{1}{c}{\multirow{1}{*}{~~1,009 images}}
    & \multicolumn{1}{c}{\multirow{1}{*}{~~1,009 images}}
    & \multicolumn{1}{c}{\multirow{1}{*}{~~1,009 images}}
    & \multicolumn{1}{c}{\multirow{1}{*}{~~1,009 images}} \\
    \midrule
    HorizonNet
                & $3.51 \pm 0.79$
				& $2.78 \pm 0.90$
				& $3.17 \pm 0.27$
				& \bfseries 1.07 $\pm$ 0.15
				& \bfseries 0.70 $\pm$ 0.02 \\
	SSLayout360
                & \bfseries 2.57 $\pm$ 0.58
				& \bfseries 1.76 $\pm$ 0.51
				& \bfseries 1.42 $\pm$ 0.31
				& \bfseries 0.96 $\pm$ 0.06
				& \bfseries 0.69 $\pm$ 0.01 \\
	\end{tabular}
	}
\resizebox{\columnwidth}{!}{
	\begin{tabular} {lrrrrrr}
	\toprule
	& \multicolumn{5}{c}{Pixel Error (\%) $\downarrow$} \\
	\cmidrule{2-6}
	\multicolumn{1}{l}{\multirow{2}{*}{Method}} &
	\multicolumn{1}{c}{\multirow{1}{*}{~~~~20 labels}} &
	\multicolumn{1}{c}{\multirow{1}{*}{~~~~50 labels}} &
	\multicolumn{1}{c}{\multirow{1}{*}{~~~100 labels}} &
	\multicolumn{1}{c}{\multirow{1}{*}{~~~200 labels}} &
	\multicolumn{1}{c}{\multirow{1}{*}{~~~963 labels}} \\
    & \multicolumn{1}{c}{\multirow{1}{*}{~~1,009 images}}
    & \multicolumn{1}{c}{\multirow{1}{*}{~~1,009 images}}
    & \multicolumn{1}{c}{\multirow{1}{*}{~~1,009 images}}
    & \multicolumn{1}{c}{\multirow{1}{*}{~~1,009 images}}
    & \multicolumn{1}{c}{\multirow{1}{*}{~~1,009 images}} \\
    \midrule
    HorizonNet
                & $5.68 \pm 0.52$
				& $5.03 \pm 0.45$
				& $4.75 \pm 0.06$
				& $3.17 \pm 0.17$
				& $1.97 \pm 0.03$ \\
	SSLayout360
                & \bfseries 4.84 $\pm$ 0.32
				& \bfseries 3.73 $\pm$ 0.26
				& \bfseries 3.47 $\pm$ 0.24
				& \bfseries 2.72 $\pm$ 0.08
				& \bfseries 1.90 $\pm$ 0.02 \\
    \bottomrule
	\end{tabular}
	}
\end{minipage}
\begin{minipage}[t]{0.49\textwidth}
\centering
\resizebox{\columnwidth}{!}{
	\begin{tabular} {lrrrrrr}
	\toprule
	\multicolumn{2}{l}{\textbf{Stanford 2D-3D}}
	& \multicolumn{3}{c}{3D IoU (\%) $\uparrow$} \\
	\cmidrule{2-6}
	\multicolumn{1}{l}{\multirow{2}{*}{Method}} &
	\multicolumn{1}{c}{\multirow{1}{*}{~~20 labels}} &
	\multicolumn{1}{c}{\multirow{1}{*}{~~50 labels}} &
	\multicolumn{1}{c}{\multirow{1}{*}{~~100 labels}} &
	\multicolumn{1}{c}{\multirow{1}{*}{~~200 labels}} &
	\multicolumn{1}{c}{\multirow{1}{*}{~~916 labels}} \\
    & \multicolumn{1}{c}{\multirow{1}{*}{~~~949 images}}
    & \multicolumn{1}{c}{\multirow{1}{*}{~~~949 images}}
    & \multicolumn{1}{c}{\multirow{1}{*}{~~~~949 images}}
    & \multicolumn{1}{c}{\multirow{1}{*}{~~~~949 images}}
    & \multicolumn{1}{c}{\multirow{1}{*}{~~~~949 images}} \\
    \midrule
    HorizonNet
                & $62.20 \pm 3.98$
                & $68.27 \pm 1.45$
                & $69.94 \pm 3.64$
                & $74.95 \pm 3.69$
				& $82.79 \pm 0.90$ \\
	SSLayout360
                & \bfseries 71.60 $\pm$ 2.04
				& \bfseries 73.86 $\pm$ 1.65
				& \bfseries 76.96 $\pm$ 1.20
				& \bfseries 79.78 $\pm$ 0.83
				& \bfseries 84.66 $\pm$ 0.57 \\
	\end{tabular}
	}
\resizebox{\columnwidth}{!}{
	\begin{tabular} {lrrrrrr}
	\toprule
	& \multicolumn{5}{c}{Corner Error (\%) $\downarrow$} \\
	\cmidrule{2-6}
	\multicolumn{1}{l}{\multirow{2}{*}{Method}} &
	\multicolumn{1}{c}{\multirow{1}{*}{~~20 labels}} &
	\multicolumn{1}{c}{\multirow{1}{*}{~~50 labels}} &
	\multicolumn{1}{c}{\multirow{1}{*}{~~100 labels}} &
	\multicolumn{1}{c}{\multirow{1}{*}{~~200 labels}} &
	\multicolumn{1}{c}{\multirow{1}{*}{~~916 labels}} \\
    & \multicolumn{1}{c}{\multirow{1}{*}{~~~949 images}}
    & \multicolumn{1}{c}{\multirow{1}{*}{~~~949 images}}
    & \multicolumn{1}{c}{\multirow{1}{*}{~~~~949 images}}
    & \multicolumn{1}{c}{\multirow{1}{*}{~~~~949 images}}
    & \multicolumn{1}{c}{\multirow{1}{*}{~~~~949 images}} \\
    \midrule
    HorizonNet
                & $2.70 \pm 0.60$
                & $1.64 \pm 0.12$
                & $1.66 \pm 0.20$
                & $1.50 \pm 0.18$
				& $0.64 \pm 0.02$ \\
	SSLayout360
                & \bfseries 1.69 $\pm$ 0.22
                & \bfseries 1.32 $\pm$ 0.04
                & \bfseries 1.15 $\pm$ 0.24
                & \bfseries 1.04 $\pm$ 0.13
				& \bfseries 0.60 $\pm$ 0.01 \\
	\end{tabular}
	}
\resizebox{\columnwidth}{!}{
	\begin{tabular} {lrrrrrr}
	\toprule
	& \multicolumn{5}{c}{Pixel Error (\%) $\downarrow$} \\
	\cmidrule{2-6}
	\multicolumn{1}{l}{\multirow{2}{*}{Method}} &
	\multicolumn{1}{c}{\multirow{1}{*}{~~20 labels}} &
	\multicolumn{1}{c}{\multirow{1}{*}{~~50 labels}} &
	\multicolumn{1}{c}{\multirow{1}{*}{~~100 labels}} &
	\multicolumn{1}{c}{\multirow{1}{*}{~~200 labels}} &
	\multicolumn{1}{c}{\multirow{1}{*}{~~916 labels}} \\
    & \multicolumn{1}{c}{\multirow{1}{*}{~~~949 images}}
    & \multicolumn{1}{c}{\multirow{1}{*}{~~~949 images}}
    & \multicolumn{1}{c}{\multirow{1}{*}{~~~~949 images}}
    & \multicolumn{1}{c}{\multirow{1}{*}{~~~~949 images}}
    & \multicolumn{1}{c}{\multirow{1}{*}{~~~~949 images}} \\
    \midrule
    HorizonNet
                & $5.03 \pm 0.51$
                & $3.95 \pm 0.22$
                & $3.77 \pm 0.39$
                & $3.69 \pm 0.26$
				& $2.13 \pm 0.05$ \\
	SSLayout360
                & \bfseries 3.50 $\pm$ 0.17
                & \bfseries 3.24 $\pm$ 0.18
                & \bfseries 3.01 $\pm$ 0.26
                & \bfseries 2.91 $\pm$ 0.27
				& \bfseries 1.97 $\pm$ 0.06 \\
    \bottomrule
	\end{tabular}
	} \smallskip
\end{minipage}
{\caption{Quantitative cuboid layout results evaluated on the \textbf{PanoContext (left)} and \textbf{Stanford 2D-3D (right)} test sets over four randomized trials. The semi-supervised SSLayout360 settings outperform the supervised HorizonNet baselines across most metrics under consideration.}
\label{cuboid}}
\end{table*}

\subsection{Evaluation Protocol}
Existing evaluation procedures for layout estimation suffer from statistical unreliability, given the relatively small sample sizes of training, validation, and test instances. Case in point, our findings in \Cref{cuboid} show that supervised results can vary as much as 4\% points. Moreover, prior work only reported point estimates without standard error bounds for performance evaluations, making a comparison of published methods difficult when considering for statistical significance. We adopt the rigorous evaluation protocol previously used for semi-supervised image classification \cite{ssl-eval} and extend it to this work for semi-supervised layout estimation. 

We conduct our supervised and semi-supervised experiments over four runs using different random seeds, and report the mean and standard deviation to assess statistical significance. This helps to evaluate the contributions of unlabeled data instead of confounding statistical noise inherent in deep neural networks (\eg, dropout, weight initialization). For SSL, we follow the standard practice of randomly sampling varying amounts of the training data as labeled examples while treating the combined training and validation sets, discarding all label information, as the source of unlabeled data \cite{ssl-eval,mean-teacher,tran-ssl}. We take extra care not to include any test instances as unlabeled data. We train SSLayout360 with both labeled and unlabeled data according to \Cref{algorithm} and compare its performance to that of HorizonNet trained using only the labeled portion in the traditional supervised manner.

We extend our rigorous evaluation protocol to include experiments with extra unlabeled and synthetic data, which have not been explored for layout estimation. In experiments using synthetic data, we ask the question: given a strong predictor pre-trained on synthetic data, can our model bridge the performance gap between synthetic and real data by using a combination of both in the semi-supervised setting?

\subsection{Implementation Details}
We implement SSLayout360 using PyTorch \cite{pytorch} and train on 2 NVIDIA TITAN X GPUs each with 12GB of video memory. Our SSL experiments take between 6 and 65 hours to complete, depending on the number of training epochs and how much unlabeled data is used in combination with labeled data. The supervised HorizonNet baselines are trained using the original authors' publicly available source code for direct comparison. See Section \ref{training} in the appendix for a detailed breakdown of our training protocol.

\medskip

\noindent
\textbf{Data Processing and Augmentation} ~ We separate the input data source into labeled and unlabeled branches. All images are pre-processed by the panorama alignment algorithm described in \cite{layoutnet} to enforce the Manhattan constraint, owing to the use of HorizonNet as the prediction function. We follow standard panorama augmentation techniques \cite{horizonnet,dula-net,layoutnet}, and apply random stretching in both $(k_x, k_z) \in [0.5, 1.5]$ directions, horizontal rotation $r \in (0\degree, 360\degree]$, left-right flipping with probability $0.5$, and gamma correction with $\gamma \in [0.5, 2.0]$ to each branch independently.

\medskip

\noindent
\textbf{Hyper-parameters} ~ We start with the hyper-parameter configuration for the HorizonNet baseline, and only tune hyper-parameters specific to SSLayout360 which include: the consistency loss weight $\lambda$, the ramp-up period $T$, and EMA decay $\alpha$. In our implementation, we tune hyper-parameters on the PanoContext validation set and fix them constant for experiments on Stanford 2D-3D and MatterportLayout. We make a conscientious effort to keep our hyper-parameter configuration general to avoid overfitting on a per-dataset or per-experiment basis, which can limit the real-world applicability of our method. \Cref{ablation} discusses our hyper-parameter choices with detailed ablation experiments.

\begin{table}[t]
\centering
\resizebox{\columnwidth}{!}{
	\begin{tabular} {lrrrrrr}
	\toprule
	\multicolumn{2}{l}{\textbf{MatterportLayout (4-18 Corners)}}
	& \multicolumn{3}{c}{~~3D IoU (\%) $\uparrow$} \\
	\cmidrule{2-6}
	\multicolumn{1}{l}{\multirow{2}{*}{Method}} &
	\multicolumn{1}{c}{\multirow{1}{*}{~~~~50 labels}} &
	\multicolumn{1}{c}{\multirow{1}{*}{~~~100 labels}} &
	\multicolumn{1}{c}{\multirow{1}{*}{~~~200 labels}} &
	\multicolumn{1}{c}{\multirow{1}{*}{~~~400 labels}} &
	\multicolumn{1}{c}{\multirow{1}{*}{~~1,650 labels}} \\
    & \multicolumn{1}{c}{\multirow{1}{*}{~~1,837 images}}
    & \multicolumn{1}{c}{\multirow{1}{*}{~~1,837 images}}
    & \multicolumn{1}{c}{\multirow{1}{*}{~~1,837 images}}
    & \multicolumn{1}{c}{\multirow{1}{*}{~~1,837 images}}
    & \multicolumn{1}{c}{\multirow{1}{*}{~~~~1,837 images}} \\
    \midrule
    HorizonNet
                & $63.44 \pm 0.56$
				& $68.79 \pm 0.49$
				& $72.25 \pm 0.50$
				& $74.46 \pm 0.35$
				& $79.12 \pm 0.37$ \\
	SSLayout360
                & \bfseries 67.42 $\pm$ 0.24
				& \bfseries 72.37 $\pm$ 0.35
				& \bfseries 75.31 $\pm$ 0.37
				& \bfseries 77.09 $\pm$ 0.41
				& \bfseries 80.33 $\pm$ 0.48 \\
	\end{tabular}
	}
\resizebox{\columnwidth}{!}{
	\begin{tabular} {lrrrrrr}
	\toprule
	& \multicolumn{5}{c}{2D IoU (\%) $\uparrow$} \\
	\cmidrule{2-6}
	\multicolumn{1}{l}{\multirow{2}{*}{Method}} &
	\multicolumn{1}{c}{\multirow{1}{*}{~~~~50 labels}} &
	\multicolumn{1}{c}{\multirow{1}{*}{~~~100 labels}} &
	\multicolumn{1}{c}{\multirow{1}{*}{~~~200 labels}} &
	\multicolumn{1}{c}{\multirow{1}{*}{~~~400 labels}} &
	\multicolumn{1}{c}{\multirow{1}{*}{~~1,650 labels}} \\
    & \multicolumn{1}{c}{\multirow{1}{*}{~~1,837 images}}
    & \multicolumn{1}{c}{\multirow{1}{*}{~~1,837 images}}
    & \multicolumn{1}{c}{\multirow{1}{*}{~~1,837 images}}
    & \multicolumn{1}{c}{\multirow{1}{*}{~~1,837 images}}
    & \multicolumn{1}{c}{\multirow{1}{*}{~~~~1,837 images}} \\
    \midrule
    HorizonNet
                & $67.17 \pm 0.65$
				& $72.06 \pm 0.49$
				& $75.16 \pm 0.53$
				& $77.15 \pm 0.36$
				& $81.54 \pm 0.31$ \\
	SSLayout360
                & \bfseries 71.03 $\pm$ 0.28
				& \bfseries 75.46 $\pm$ 0.36
				& \bfseries 78.05 $\pm$ 0.33
				& \bfseries 79.67 $\pm$ 0.40
				& \bfseries 82.54 $\pm$ 0.51 \\
	\end{tabular}
	}
\resizebox{\columnwidth}{!}{
	\begin{tabular} {lrrrrrr}
	\toprule
	& \multicolumn{5}{c}{$\delta_1$ $\uparrow$} \\
	\cmidrule{2-6}
	\multicolumn{1}{l}{\multirow{2}{*}{Method}} &
	\multicolumn{1}{c}{\multirow{1}{*}{~~~~50 labels}} &
	\multicolumn{1}{c}{\multirow{1}{*}{~~~100 labels}} &
	\multicolumn{1}{c}{\multirow{1}{*}{~~~200 labels}} &
	\multicolumn{1}{c}{\multirow{1}{*}{~~~400 labels}} &
	\multicolumn{1}{c}{\multirow{1}{*}{~~1,650 labels}} \\
    & \multicolumn{1}{c}{\multirow{1}{*}{~~1,837 images}}
    & \multicolumn{1}{c}{\multirow{1}{*}{~~1,837 images}}
    & \multicolumn{1}{c}{\multirow{1}{*}{~~1,837 images}}
    & \multicolumn{1}{c}{\multirow{1}{*}{~~1,837 images}}
    & \multicolumn{1}{c}{\multirow{1}{*}{~~~~1,837 images}} \\
    \midrule
    HorizonNet
                & $0.76 \pm 0.01$
				& $0.84 \pm 0.01$
				& $0.89 \pm 0.01$
				& $0.91 \pm 0.01$
				& \bfseries 0.94 $\pm$ 0.01 \\
	SSLayout360
                & \bfseries 0.81 $\pm$ 0.01
				& \bfseries 0.89 $\pm$ 0.01
				& \bfseries 0.91 $\pm$ 0.01
				& \bfseries 0.93 $\pm$ 0.01
				& \bfseries 0.95 $\pm$ 0.01 \\
	\end{tabular}
	}
\resizebox{\columnwidth}{!}{
	\begin{tabular} {lrrrrrr}
	\toprule
	& \multicolumn{5}{c}{RMSE $\downarrow$} \\
	\cmidrule{2-6}
	\multicolumn{1}{l}{\multirow{2}{*}{Method}} &
	\multicolumn{1}{c}{\multirow{1}{*}{~~~~50 labels}} &
	\multicolumn{1}{c}{\multirow{1}{*}{~~~100 labels}} &
	\multicolumn{1}{c}{\multirow{1}{*}{~~~200 labels}} &
	\multicolumn{1}{c}{\multirow{1}{*}{~~~400 labels}} &
	\multicolumn{1}{c}{\multirow{1}{*}{~~1,650 labels}} \\
    & \multicolumn{1}{c}{\multirow{1}{*}{~~1,837 images}}
    & \multicolumn{1}{c}{\multirow{1}{*}{~~1,837 images}}
    & \multicolumn{1}{c}{\multirow{1}{*}{~~1,837 images}}
    & \multicolumn{1}{c}{\multirow{1}{*}{~~1,837 images}}
    & \multicolumn{1}{c}{\multirow{1}{*}{~~~~1,837 images}} \\
    \midrule
    HorizonNet
                & $0.41 \pm 0.01$
				& $0.34 \pm 0.01$
				& $0.30 \pm 0.01$
				& $0.28 \pm 0.01$
				& \bfseries 0.23 $\pm$ 0.01 \\
	SSLayout360
                & \bfseries 0.35 $\pm$ 0.01
				& \bfseries 0.29 $\pm$ 0.01
				& \bfseries 0.27 $\pm$ 0.01
				& \bfseries 0.25 $\pm$ 0.01
				& \bfseries 0.22 $\pm$ 0.01 \\
    \bottomrule
	\end{tabular}
	} \smallskip
{\caption{Quantitative non-cuboid layout results evaluated on the MatterportLayout test set over four runs. SSLayout360 surpasses the supervised HorizonNet baselines across all settings and metrics.}
\label{non-cuboid}}
\end{table}

\medskip

\noindent
\textbf{Model Selection} ~ We use the same underlying architecture and training protocol for both supervised and SSL experiments, with the exception of tuning hyper-parameters specific to SSLayout360. This is to ensure that any performance boost in the SSL setting is directly attributed to unlabeled data and not to changes in model configuration.

We employ the Adam optimizer \cite{adam} to train HorizonNet and SSLayout360. Following HorizonNet's training protocol, we use learning rate 0.0003 and batch size 8 for PanoContext and Stanford 2D-3D experiments; for MatterportLayout experiments, we use learning rate 0.0001 and batch size 4 to achieve the best results on both HorizonNet and SSLayout360. Similar to Tran~\cite{tran-ssl}, we anneal the learning rate hyper-parameter after each training step $t$ according to the polynomial schedule: $\text{lr} \times \left(1 - \nicefrac{t}{t_\text{max}}\right)^{0.5}$.

We check-point the best models for testing based on their best performances on the validation sets. At test time, SSLayout360 produces two sets of model parameters $\theta$ and $\bar{\theta} = \textsc{EMA}(\theta)$, both of which are expected to have comparable predictive accuracy. For simplicity, we take the average of both predictions on each test instance and report results, but otherwise do not perform any test-time augmentation.

\medskip

\noindent
\textbf{Performance Metrics} ~ We assess layout estimation performance using six standard metrics to maintain parity with previous work \cite{horizonnet,layoutnet,layoutnetv2}. For evaluation between predicted layout and the ground truth, we use 3D intersection over union (IoU), 2D IoU, corner error, and pixel error. For evaluation between predicted and ground truth layout depth, we use root mean squared error (RMSE) and $\delta_1$, defined by Zou \etal~\cite{layoutnetv2} as ``the percentage of pixels where the ratio (or its reciprocal) between the prediction and the label is within a threshold of 1.25.'' We evaluate cuboid layout estimation using 3D IoU, corner error, pixel error, and non-cuboid layout using 3D IoU, 2D IoU, RMSE, and $\delta_1$.

\begin{figure}[t]
\centering
\includegraphics[width=0.87\columnwidth]{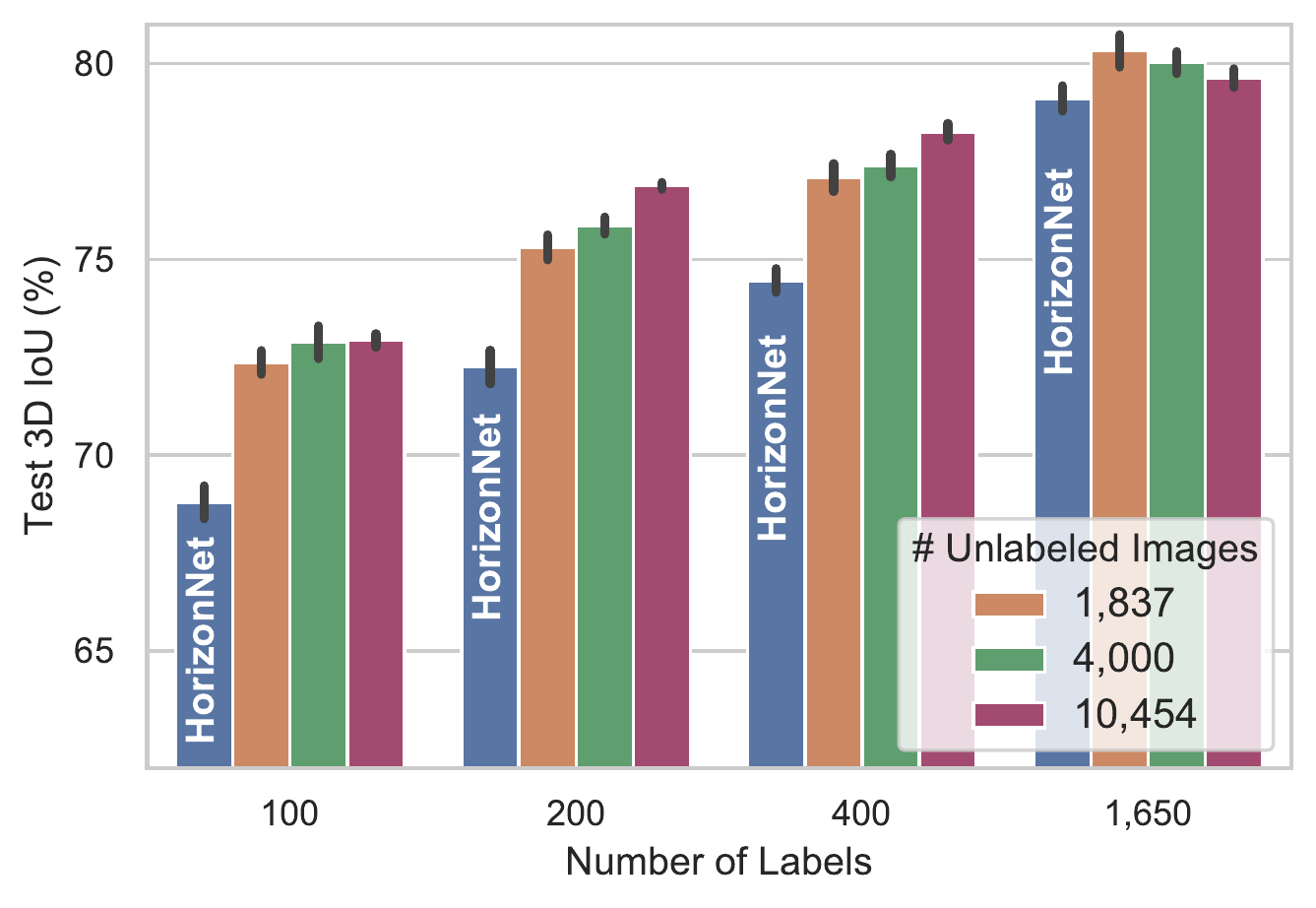}
\caption{SSLayout360 results averaged over four runs on MatterportLayout with increasing amount of unlabeled data, evaluated for rooms with 4-18 corners. More unlabeled data improves semi-supervised layout estimation when the number of labels is 400 or less. The effect of unlabeled data diminishes when all labels are used, but performance still remains above the supervised baseline.}
\label{extra}
\end{figure}

\section{Results and Analysis}

\subsection{Quantitative Evaluation}

\noindent
\textbf{Cuboid Layout Estimation} ~ Quantitative evaluations on the PanoContext and Stanford 2D-3D test sets are presented in \Cref{cuboid}. We show SSLayout360 surpasses the supervised HorizonNet baselines across most settings under consideration. For the fully supervised setting with all labeled images, SSLayout360 achieves similar performance to HorizonNet on the 3D IoU metric, and outperforms HorizonNet on the corner and pixel error metrics, indicating the benefit of learning with consistency regularization for layout estimation.

\medskip

\noindent
\textbf{Non-Cuboid Layout Estimation} ~ We present quantitative non-cuboid results on the challenging MatterportLayout test set in \Cref{non-cuboid}. For space considerations, we report the overall performances of rooms having a mixture of 4-18 corners. Detailed results for rooms having 4, 6, 8, 10 or more corners are reported in Section \ref{more-mp3d} of the appendix. The trend is clear: the effective use of unlabeled data, in combination with labeled data, improves layout estimation across all settings and metrics under investigation, most notably in the scenarios with only 50 and 100 labeled images.

\medskip

\noindent
\textbf{SSLayout360 without Unlabeled Data} ~ In light of the results observed in Tables \ref{cuboid} -- \ref{non-cuboid} for the fully supervised setting, we perform experiments to clarify the utility of SSLayout360 without unlabeled data, and report them in Section \ref{no-unlabeled} of the appendix due to space limitation. Our findings show that SSLayout360 without unlabeled data produces slightly better results than the HorizonNet counterpart across most settings and metrics. Our results corroborate previous SSL literature that regularization can slightly improve supervised learning without unlabeled data \cite{vat2,tran-ssl,s4l}. In scenarios with additional unlabeled data, SSLayout360 provides a significant boost in accuracy performance over the supervised baselines.

\medskip

\noindent
\textbf{SSLayout360 with Extra Unlabeled Data} ~ We randomly sample 2,163 and 8,617 extra unlabeled images from the Matterport3D dataset and combine them with the MatterportLayout training and validation sets (1,837 images) for the total of 4,000 and 10,454 unlabeled images, respectively. \Cref{extra} shows that extra unlabeled data helps improve semi-supervised layout estimation by an average of about 1\% point when the number of labels is 400 or less. When the full labeled set is used with 4,000 and 10,454 unlabeled images, test performance dips a bit but remains above the fully supervised HorizonNet baseline. Our encouraging results suggest that SSLayout360 is capable of learning additional supervisory signals from extra unlabeled data, even when there is a data distribution mismatch.

\medskip

\begin{figure}[t]
\centering
\includegraphics[width=0.9\columnwidth]{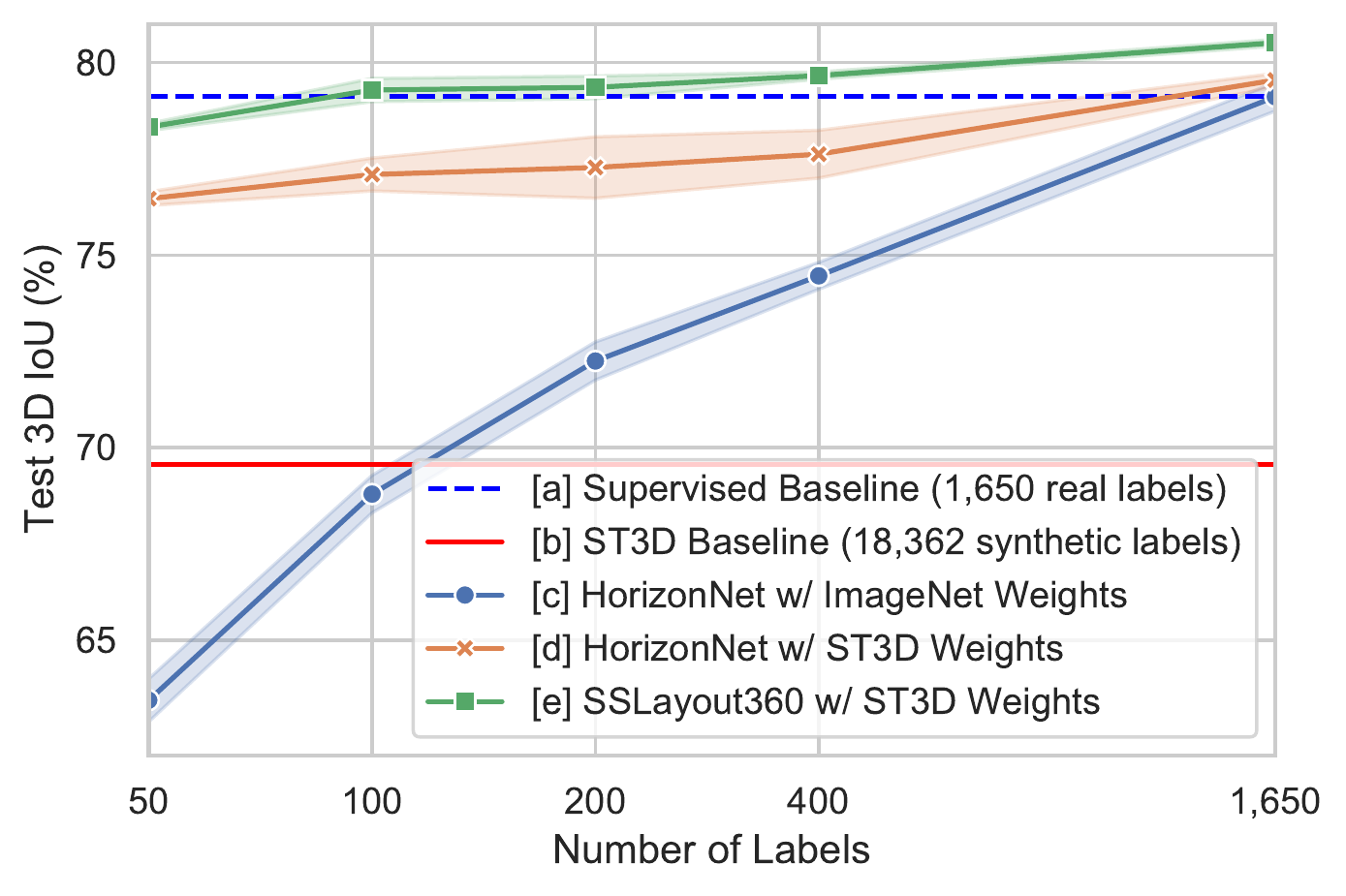} 
\caption{Supervised and semi-supervised fine-tuning experiments on MatterportLayout for rooms with 4-18 corners using Structured3D (ST3D) synthetic data. Shaded regions denote standard deviation over four runs. The $x$-axis is shown on the log scale. Coupled with HorizonNet pre-trained on ST3D \textbf{[d]}, SSLayout360 matches the fully supervised counterpart with only 200 labels \textbf{[e]}.}
\label{st3d}
\end{figure}

\noindent
\textbf{SSLayout360 with Synthetic Data} ~ \Cref{st3d} summarizes the following findings when utilizing Structured3D synthetic data to evaluate on MatterportLayout: there is a large performance gap between HorizonNet models trained on real and synthetic data \textbf{[a] vs. [b]}; HorizonNet pre-trained on Structured3D and fine-tuned on MatterportLayout \textbf{[d]} always outperforms HorizonNet initialized with ImageNet weights \textbf{[c]}; and SSLayout360 with pre-trained HorizonNet matches the fully supervised results using only 200 labels \textbf{[e]}, effectively bridging the performance gap between real and synthetic domains in the semi-supervised setting.

\begin{figure*}[t]
  \centering
  \setlength\tabcolsep{-0.5pt}
  \begin{tabular}{cccc}
    \makecell{\includegraphics[width=0.25\linewidth]{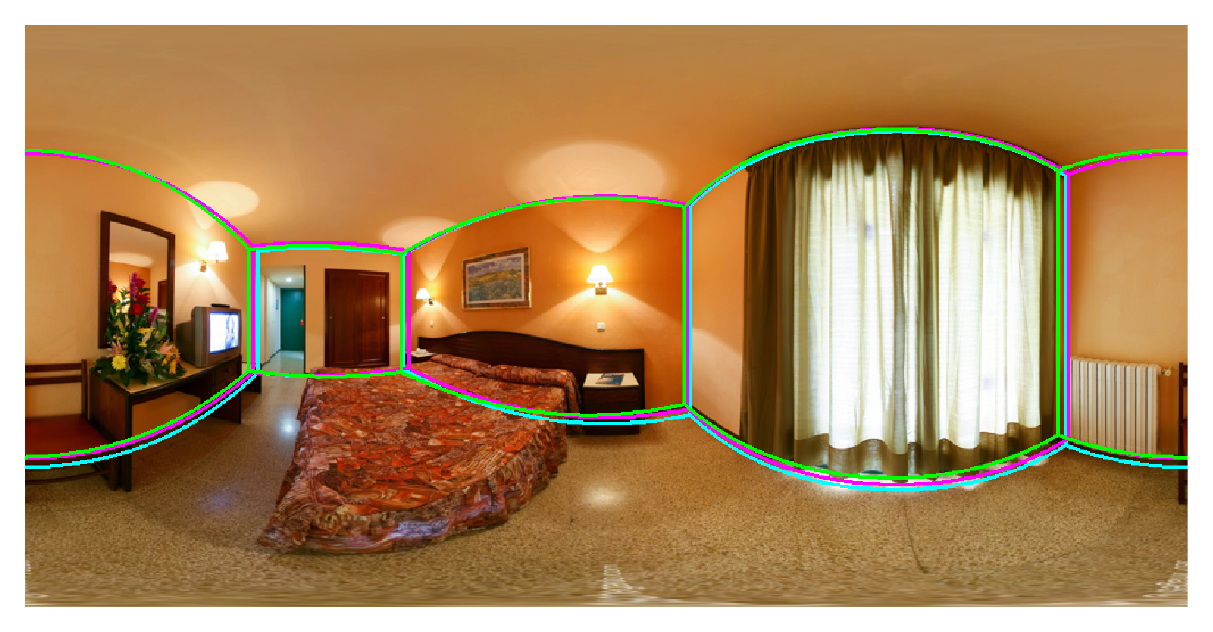}}&
    \makecell{\includegraphics[width=0.25\linewidth]{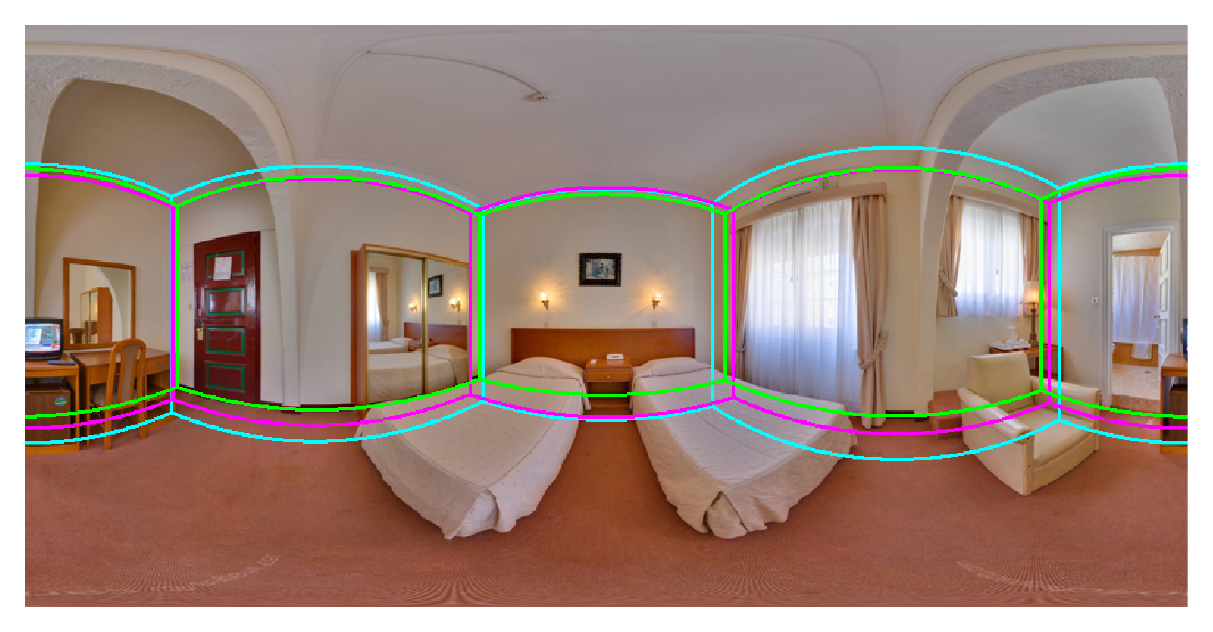}}&
    \makecell{\includegraphics[width=0.25\linewidth]{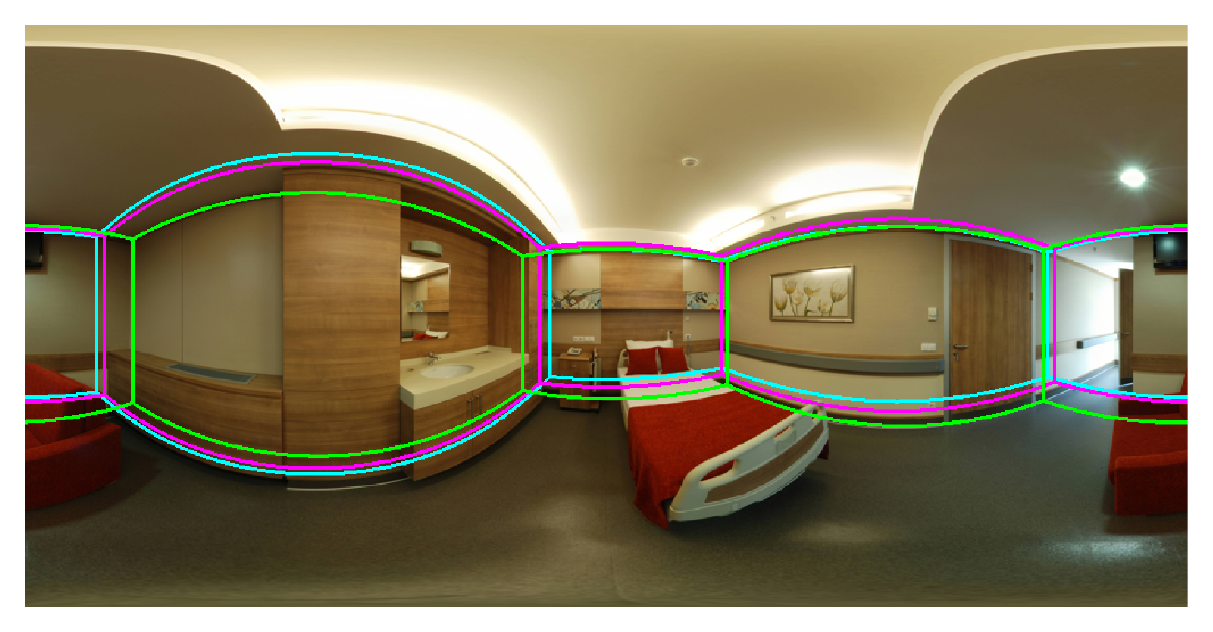}}&
    \makecell{\includegraphics[width=0.25\linewidth]{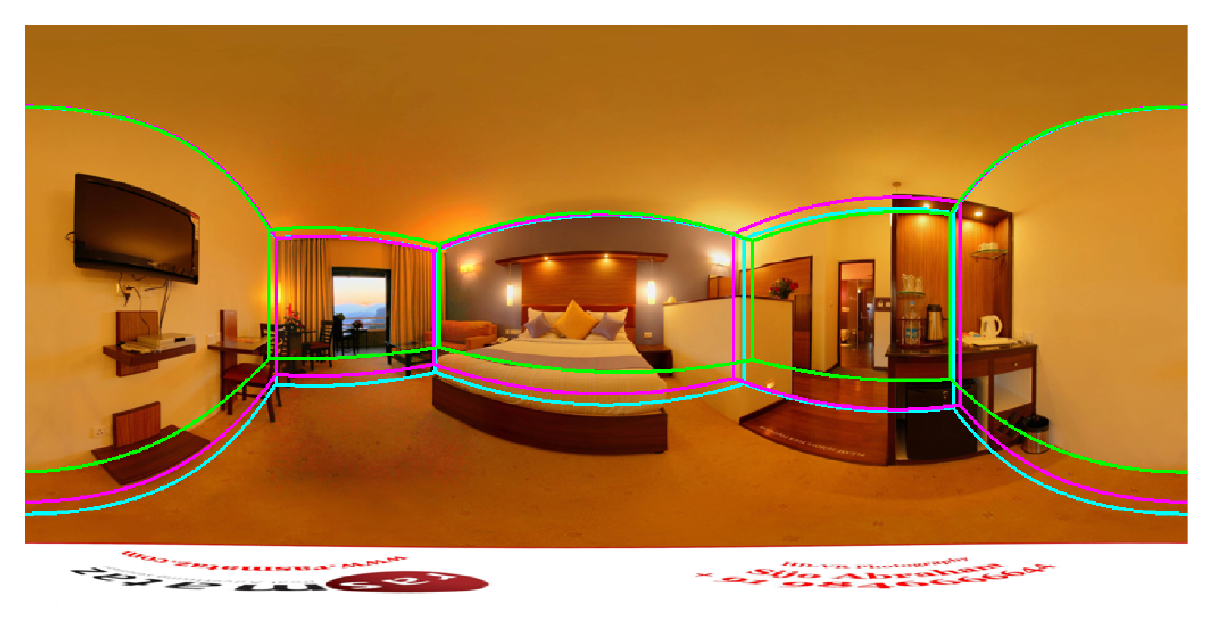}}\\[-5pt]
    \makecell{\includegraphics[width=0.25\linewidth]{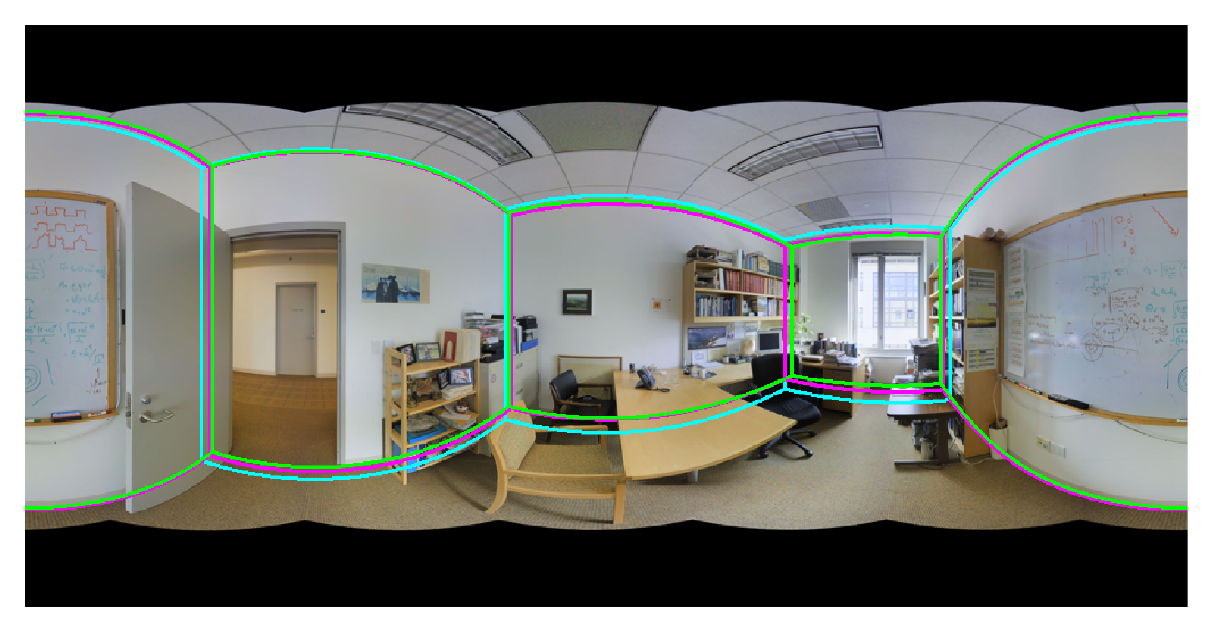}}&
    \makecell{\includegraphics[width=0.25\linewidth]{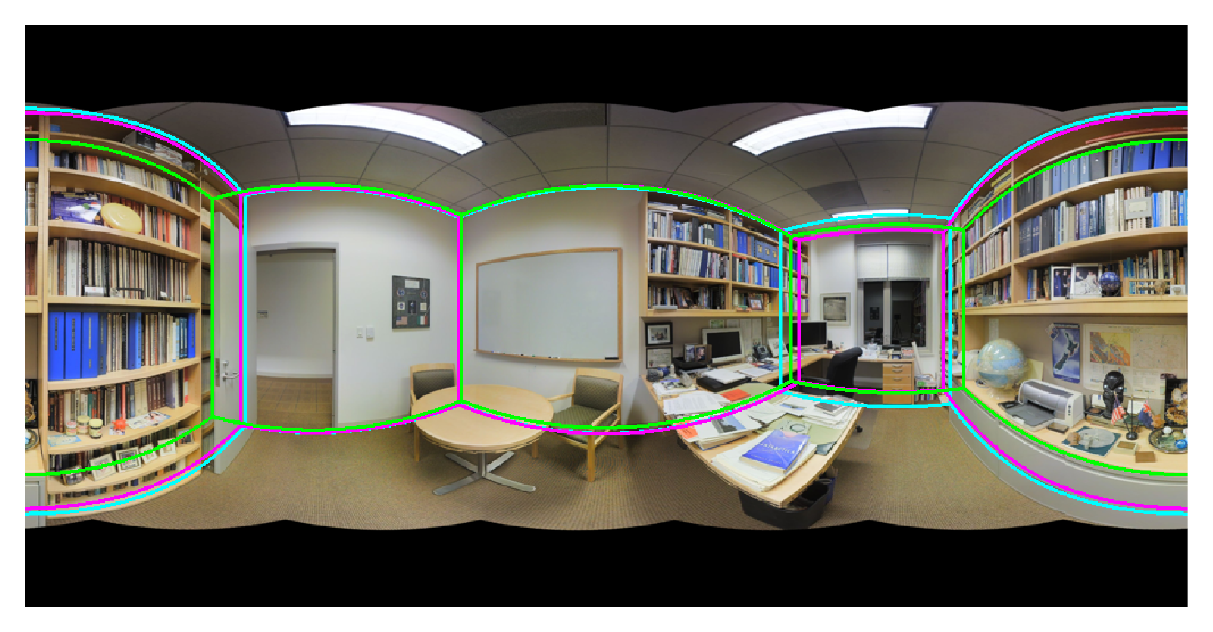}}&
    \makecell{\includegraphics[width=0.25\linewidth]{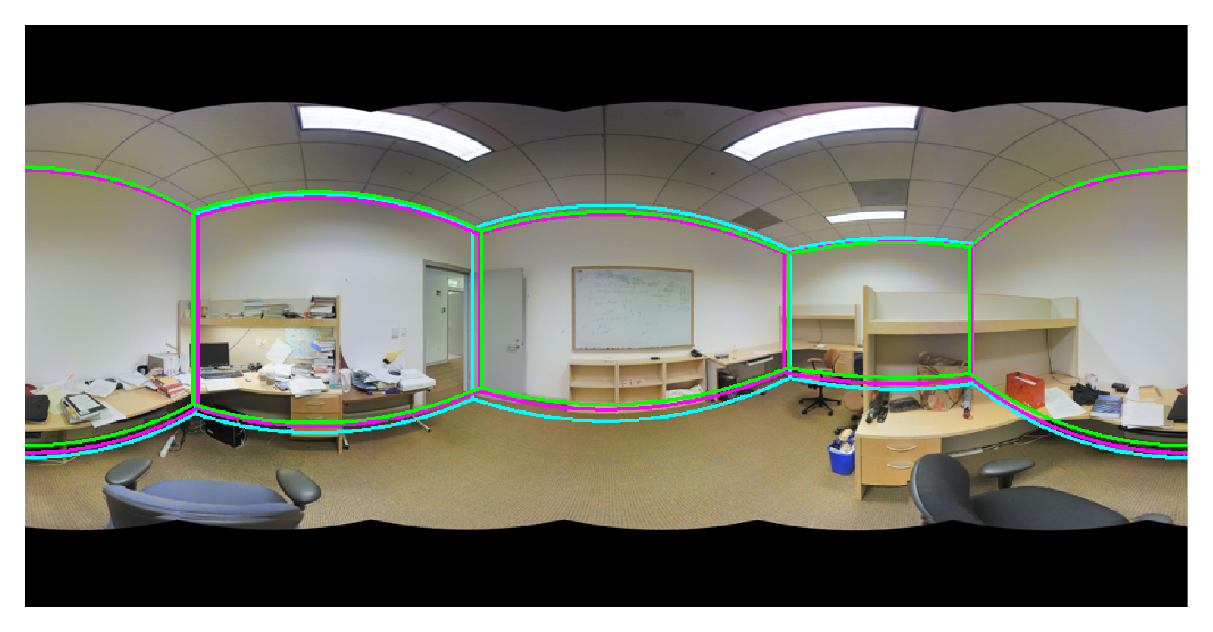}}&
    \makecell{\includegraphics[width=0.25\linewidth]{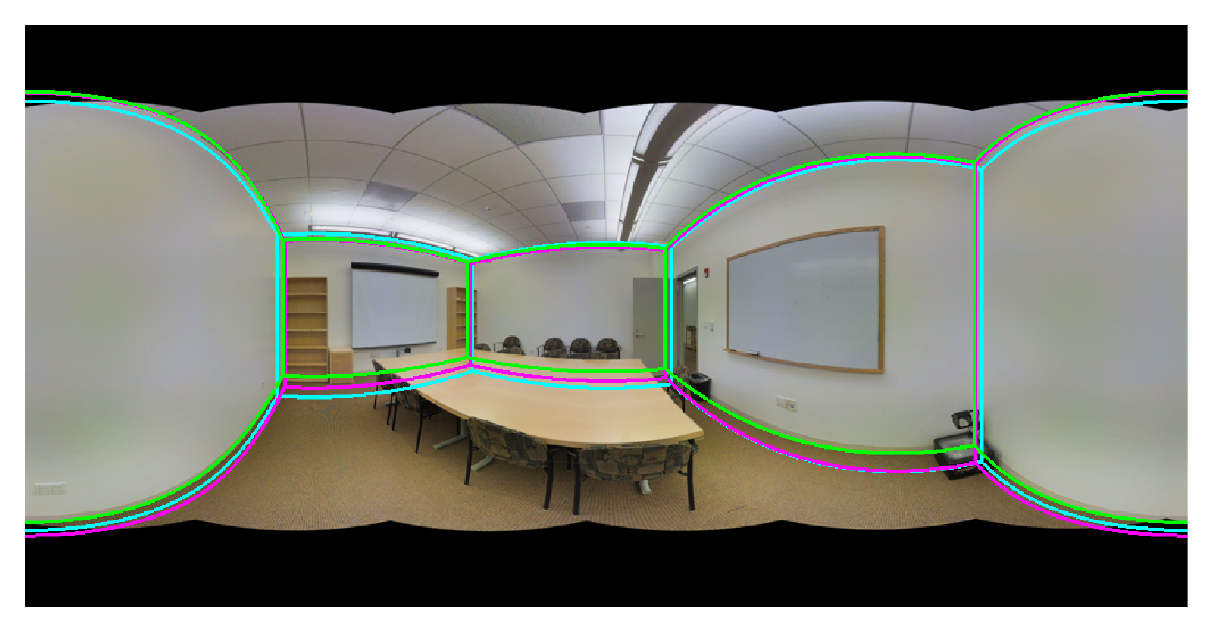}}\\[-5pt]
    \makecell{\includegraphics[width=0.25\linewidth]{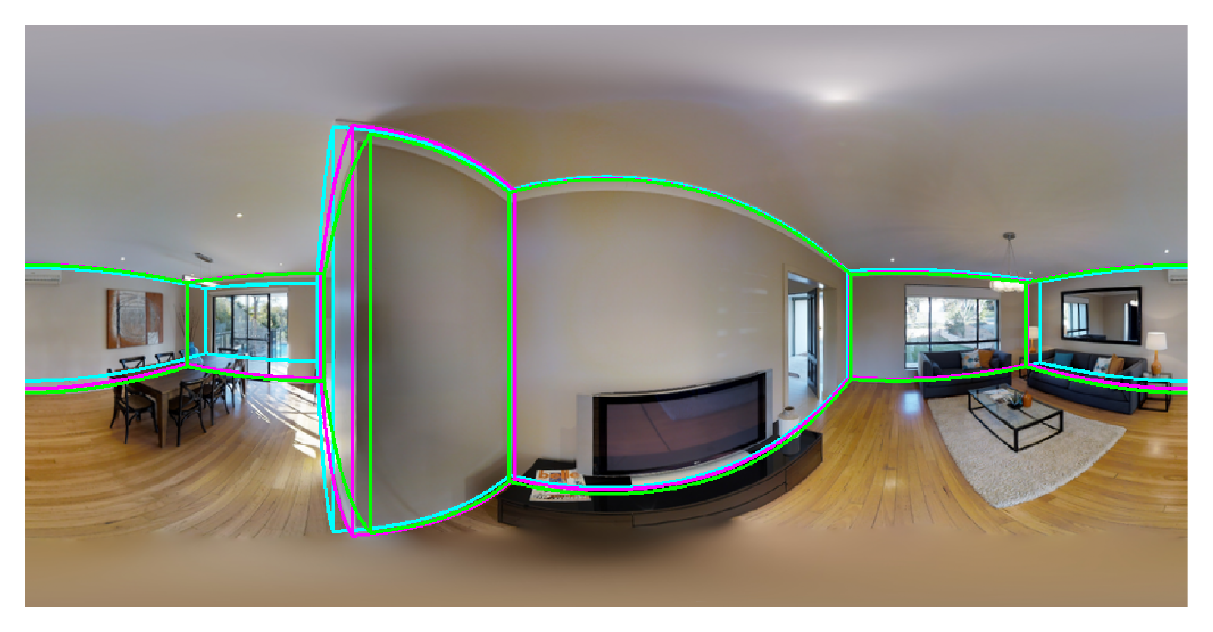}}&
    \makecell{\includegraphics[width=0.25\linewidth]{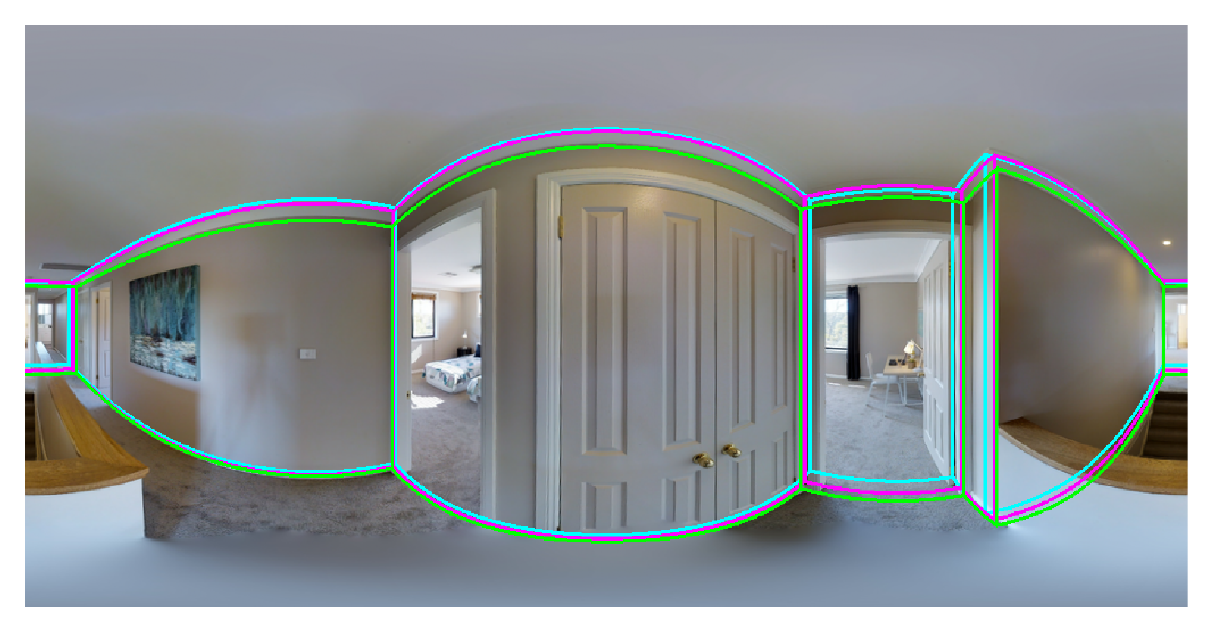}}&
    \makecell{\includegraphics[width=0.25\linewidth]{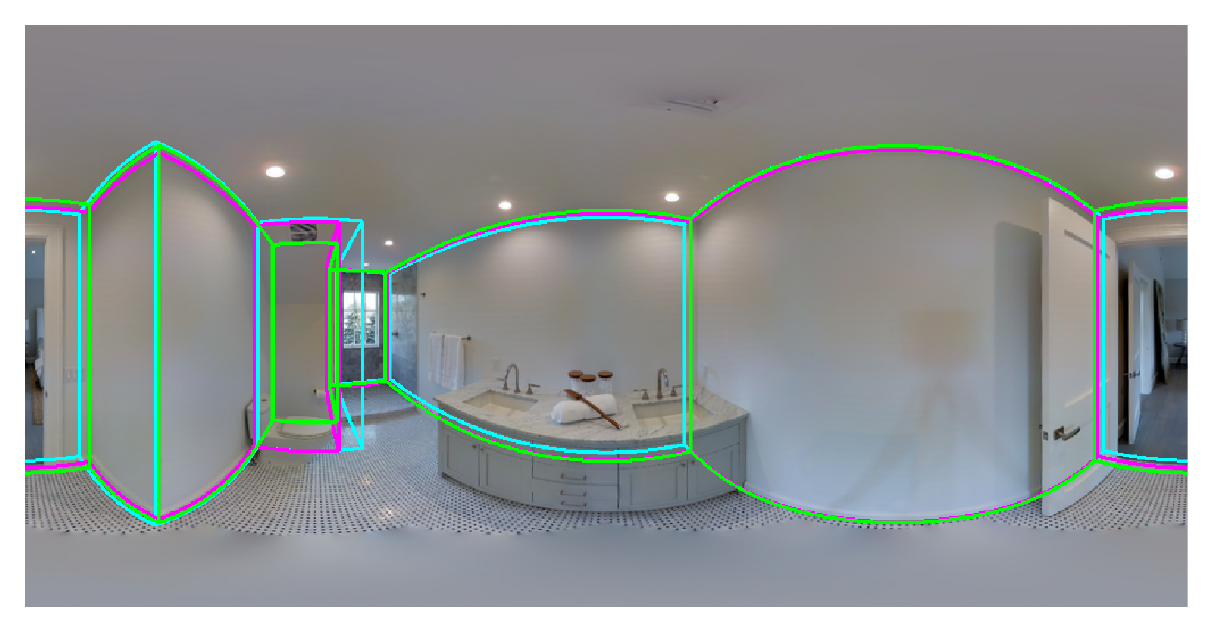}}&
    \makecell{\includegraphics[width=0.25\linewidth]{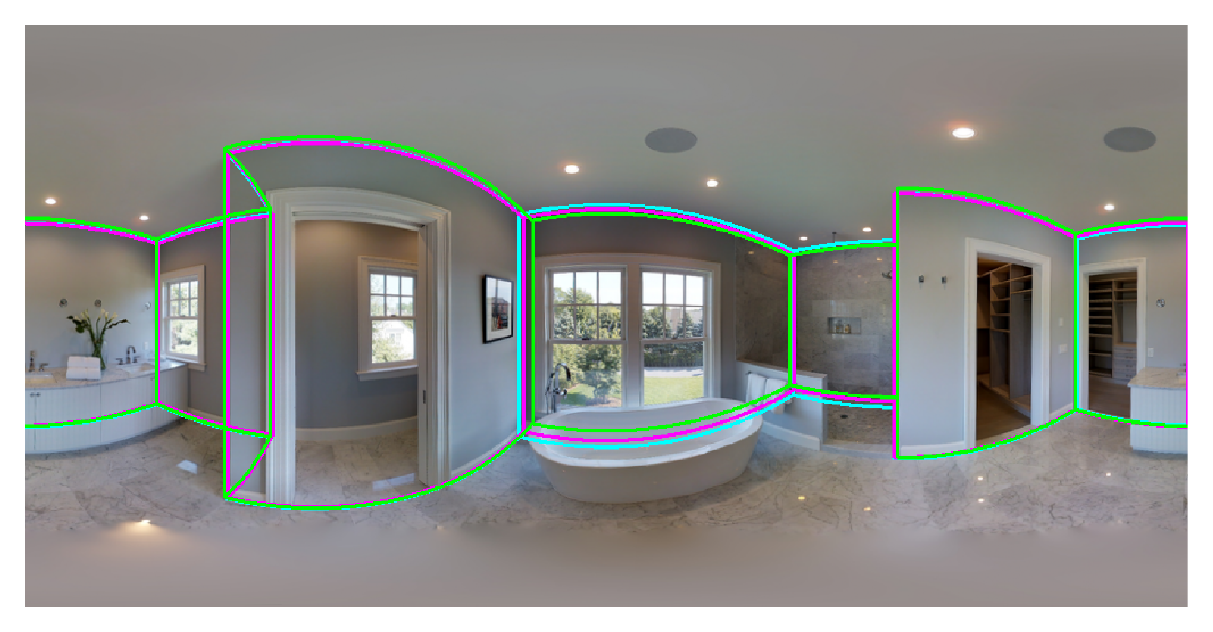}}
  \end{tabular}
  \caption{Exemplar qualitative test results of cuboid and non-cuboid layout estimation under equirectangular view. Best viewed electronically. We compare supervised HorizonNet trained on 100 labels with our SSLayout360 model trained on the same 100 labels along with unlabeled images for \textbf{PanoContext (top)}, \textbf{Stanford 2D-3D (middle)}, and \textbf{MatterportLayout (bottom)}. Layout boundary lines for HorizonNet are shown in cyan, SSLayout360 in magenta, and ground truth in green. We observe that SSLayout360 predicts layout boundary lines following more closely to the ground truth than HorizonNet, which explains the performance gap between the supervised and semi-supervised models.}
  \label{qualitative}
\end{figure*}

\begin{figure*}[t]
    \begin{minipage}[t]{0.25\textwidth}
        \subfloat{%
            \includegraphics[width=\columnwidth]{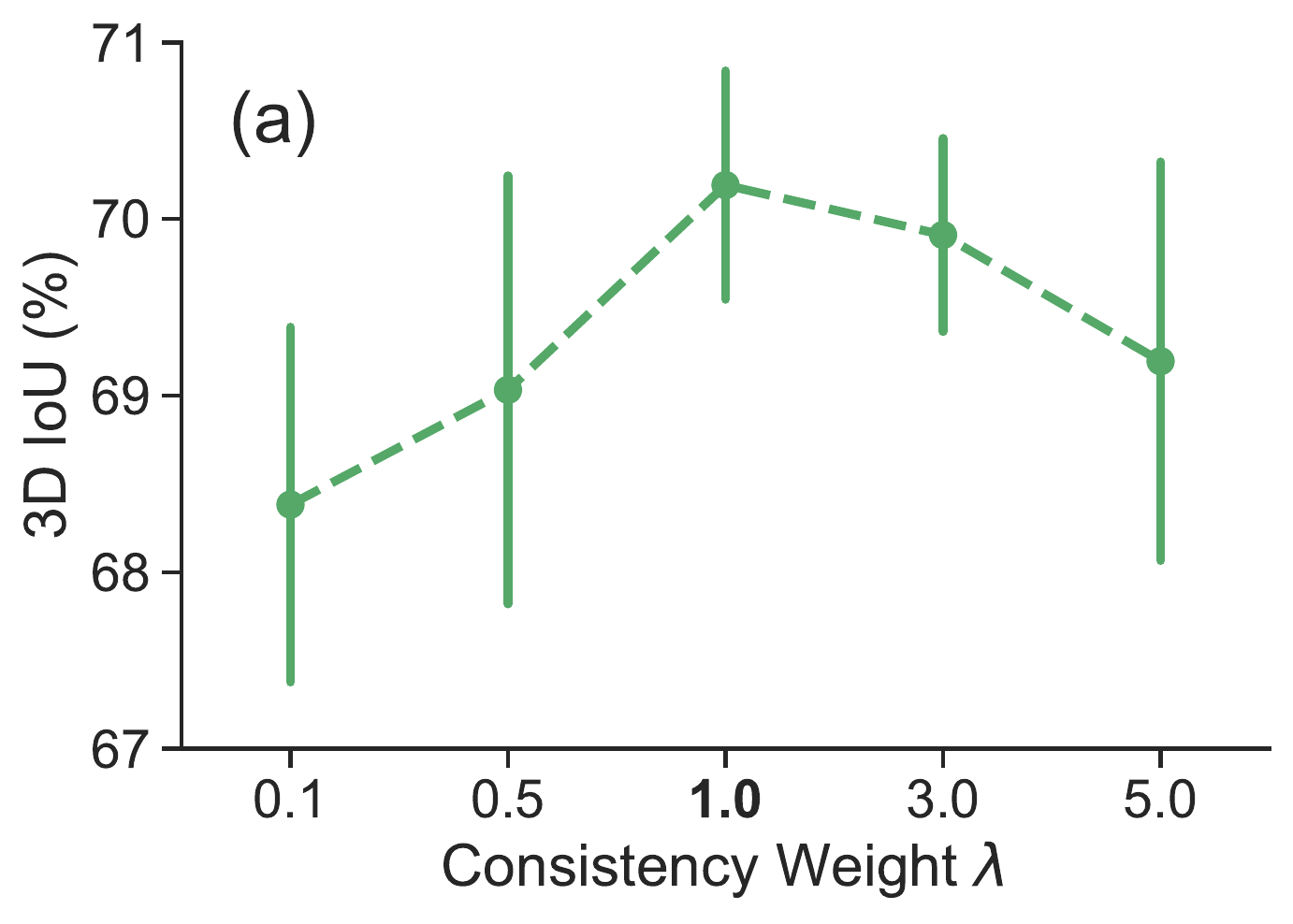}
        }
    \end{minipage}%
    \hfill
    \begin{minipage}[t]{0.25\textwidth}
        \subfloat{%
            \includegraphics[width=\columnwidth]{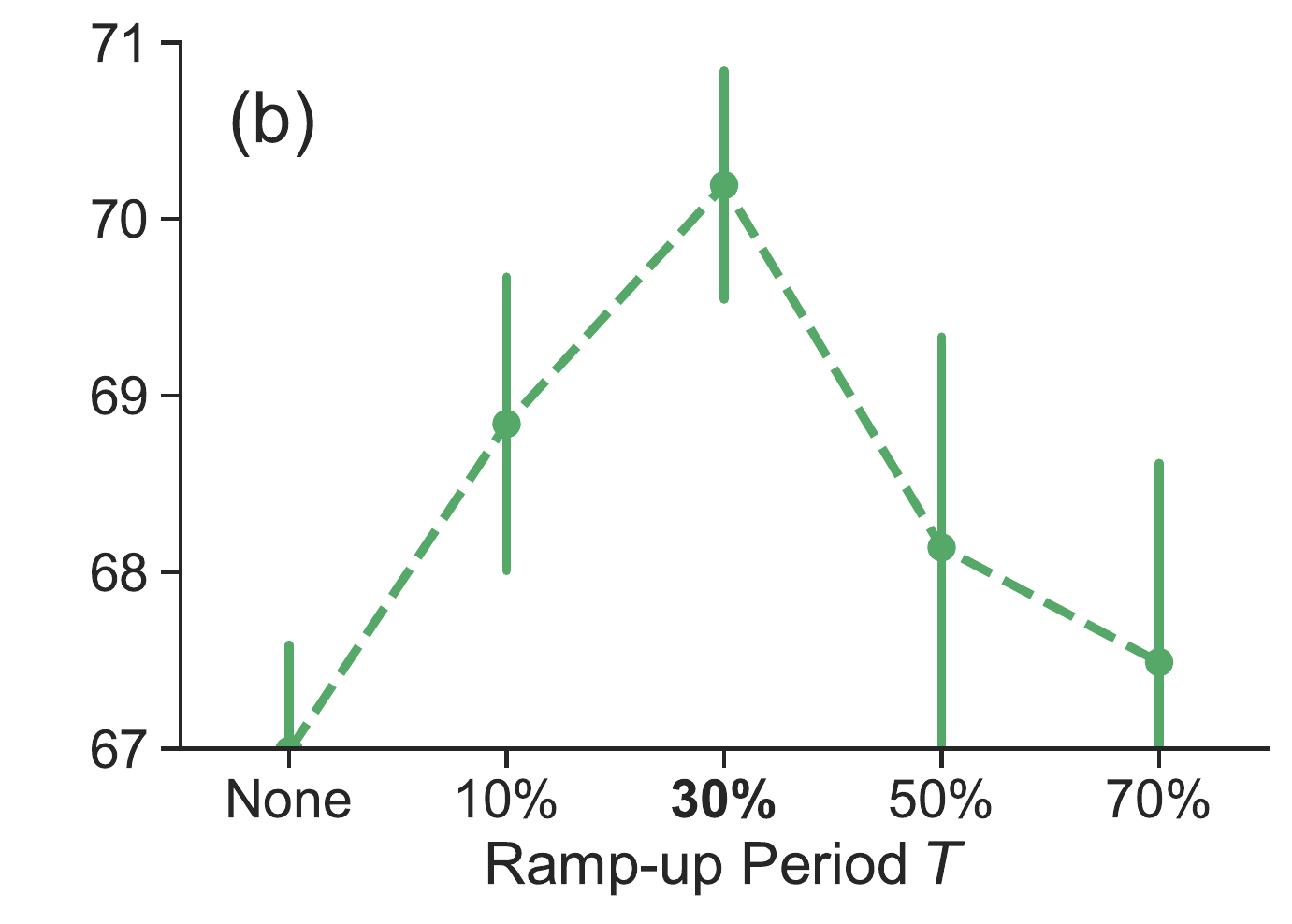}
        }
    \end{minipage}%
    \hfill
    \begin{minipage}[t]{0.25\textwidth}
        \subfloat{%
            \includegraphics[width=\columnwidth]{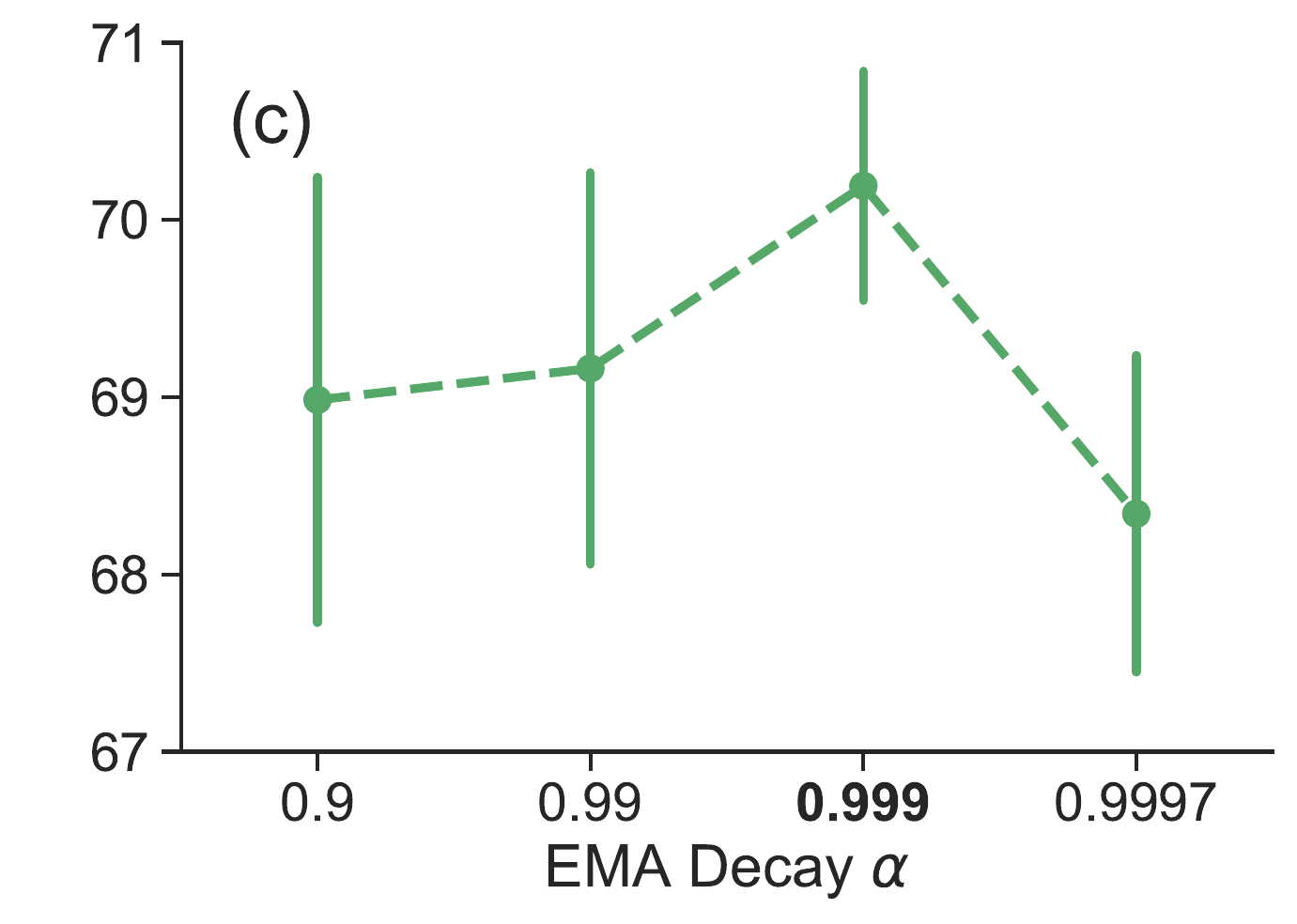}
        }
    \end{minipage}%
    \hfill
    \begin{minipage}[t]{0.25\textwidth}
        \subfloat{%
            \includegraphics[width=\columnwidth]{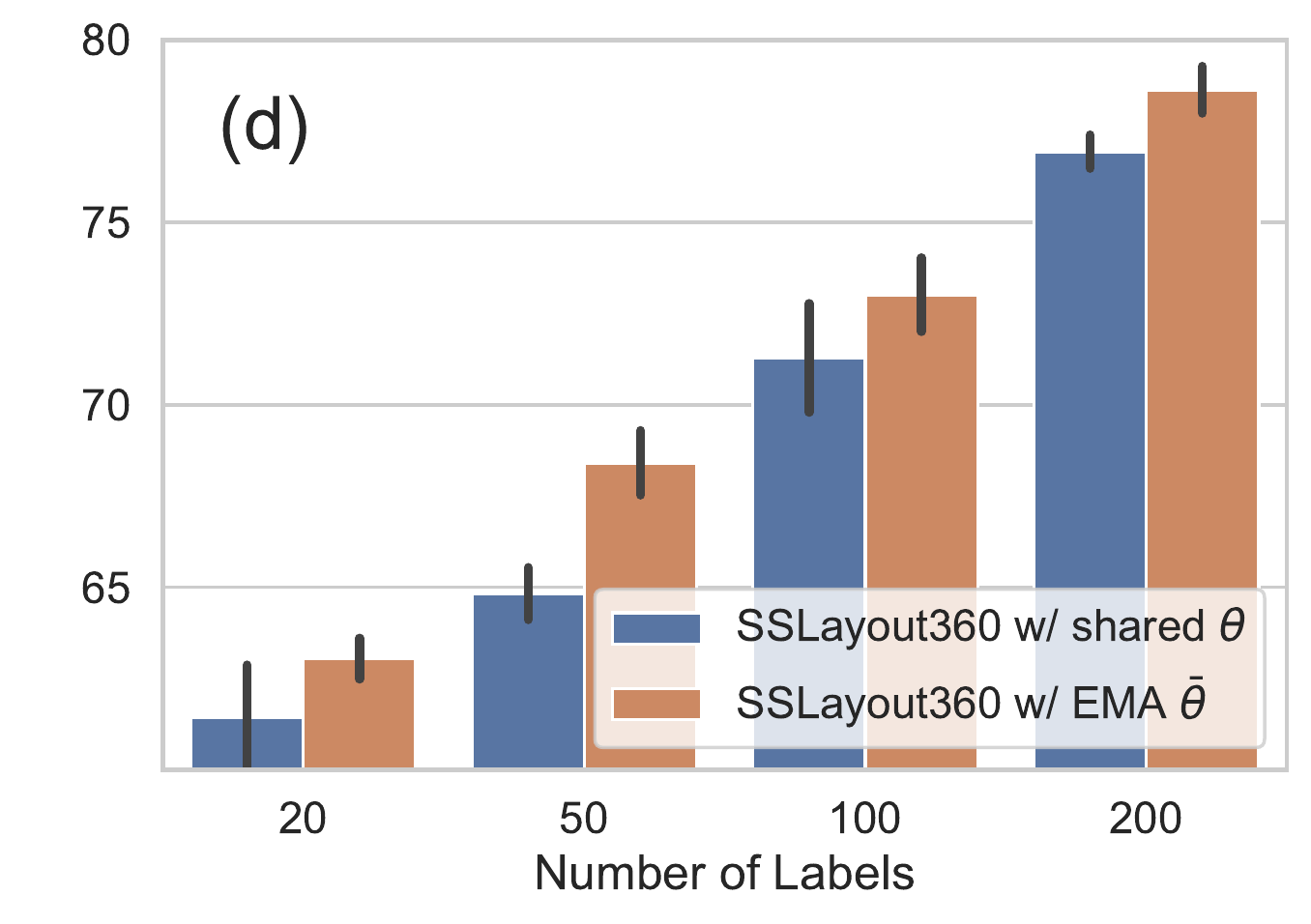}
        }
    \end{minipage}
    \smallskip
    \caption{Ablation experiments on PanoContext using 100 labeled and 1,009 unlabeled images over four randomized runs. For each hyper-parameter \textbf{(a) -- (c)}, we find the optimal value (in boldface) via the ``knee in the curve'' that produces the best 3D IoU performance on the validation set. For the experiment in \textbf{(d)}, we show SSLayout360 achieves uniformly better performance with EMA $\bar{\theta}$ as the teacher.}
    \label{ablation-charts}
\end{figure*}

\subsection{Qualitative Evaluation}
\Cref{qualitative} compares qualitative test results between HorizonNet and SSLayout360 trained on 100 labels for PanoContext, Stanford 2D-3D, and MatterportLayout under equirectangular view. We report additional qualitative results for 3D layout reconstruction utilizing the post-processing algorithm of HorizonNet in Section \ref{3d-layout} of the appendix.

\subsection{Ablation Study} \label{ablation}
Figures \ref{ablation-charts}(a) -- \ref{ablation-charts}(c) show ablation experiments where we systematically search for good values of the three hyper-parameters essential to the SSLayout360 algorithm. In each experiment, we vary one hyper-parameter while keeping the other two constant. We confirm our hypothesis in the formulation of \Cref{loss} that $\lambda = 1$ is the principled choice for the consistency weight. We observe that the ramp-up period $T@30\%$ provides a good balance between student learning and teacher soft supervision that results in the highest validation accuracy. The intuition for the ramp-up is that if it is too short, then the teacher provides unstable targets; and if ramp-up takes too long, then the teacher's supervisory contributions to performance are delayed. We also find the EMA decay coefficient $\alpha = 0.999$ gives the best validation accuracy with the lowest standard deviation. Lastly, Figure \ref{ablation-charts}(d) shows that SSLayout360 with teacher parameters $\bar{\theta}$, which is the main driver for all experiments in this paper, uniformly outperforms SSLayout360 with shared $\bar{\theta} = \theta$.

\section{Conclusion}
We presented SSLayout360, an approach that combines the strengths of HorizonNet and Mean Teacher, and extends them both to enable semi-supervised layout estimation from complex 360{\degree} panoramic indoor scenes. A distinct modification that allows our algorithm to work well is the loss formulation to regress real-valued prediction vectors to ground truth via $L_1$ and $L_2$ distances. We evaluated our approach on three challenging benchmarks across six metrics and reported steady gains in semi-supervised layout estimation from the state-of-the-art supervised baseline, utilizing only unlabeled data as the additional source of soft ``supervisory'' information. Our work takes an important first step towards robust semi-supervised layout estimation with exciting applications related to 3D scene modeling and understanding.

\section*{Acknowledgments}
The author thanks Cole Winans and Brian Keller for their continued support, Victor Palmer for technical assistance with Matlab and fruitful discussions, Mike Procopio and anonymous reviewers for their thoughtful and constructive feedback on this paper.

\clearpage

{\small
\bibliographystyle{ieee_fullname}
\bibliography{refs}
}

\clearpage

\appendix

\section{SSLayout360 Architecture Details}
\Cref{schematic-appendix} shows an expanded version of the SSLayout360 architecture depicted in \Cref{schematic}.

\renewcommand{\thefigure}{A1}
\begin{figure*}[t]
\centering
\includegraphics[width=\linewidth]{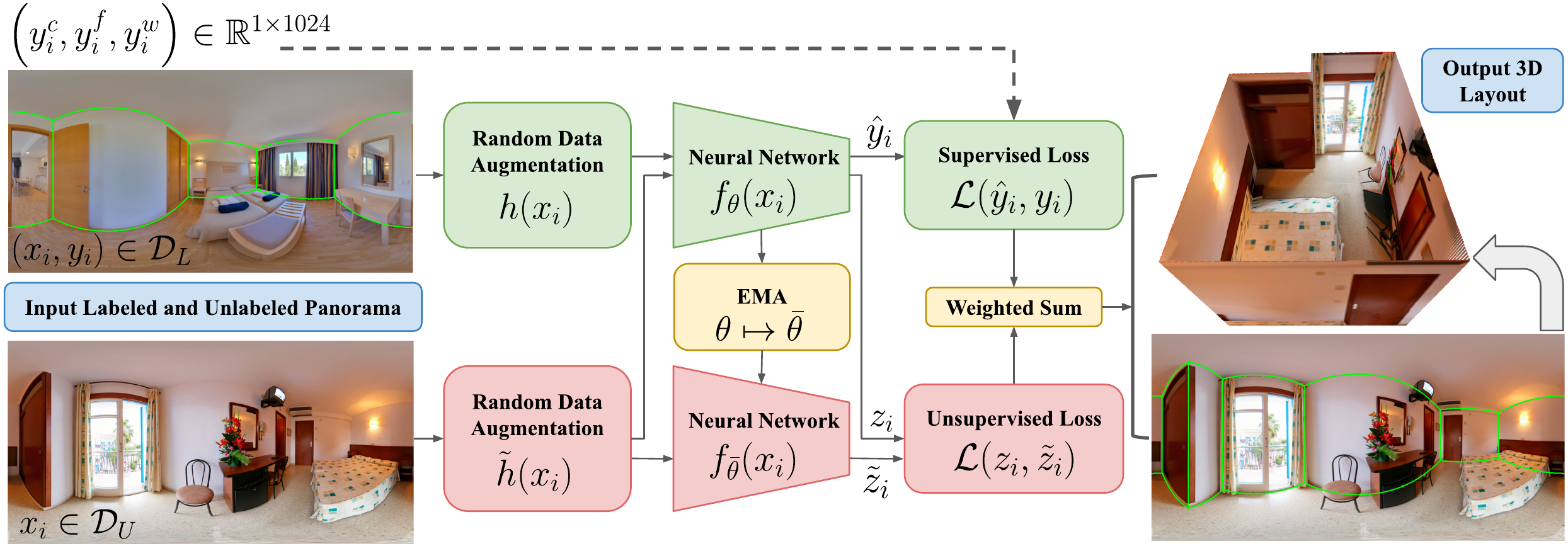}%
\caption{An illustration of the SSLayout360 architecture for semi-supervised indoor layout estimation from a 360{\degree} panoramic scene.}
\label{schematic-appendix}
\end{figure*}

\section{Implementation Details}
\subsection{Dataset Details}
\label{data-details}
\Cref{datasets} provides summary statistics of the datasets used in the empirical evaluation of this paper. We utilize the standard training, validation, and test splits provided by Zou \etal~\cite{layoutnet,layoutnetv2} for PanoContext, Stanford 2D-3D, and MatterportLayout. For PanoContext evaluation, we follow the authors' protocol of combining the training split of PanoContext with the entire Stanford 2D-3D dataset for a total of 963 labeled training examples. Similarly, we augment the Stanford 2D-3D training split with the entire PanoContext dataset, resulting in a combined total of 916 labeled training instances, for the evaluation of the Stanford 2D-3D dataset. In our semi-supervised experiments, we follow the standard practice of combining the training and validation dataset splits, discarding all label information, as the source of unlabeled data \cite{ssl-eval,mean-teacher,tran-ssl}. Thus, we use 1,009 unlabeled instances for semi-supervised experiments on PanoContext, 949 on Stanford 2D-3D, and 1,837 on MatterportLayout.

Structured3D is a large photo-realistic \emph{synthetic} dataset comprising 21,835 panoramas of rooms in 3,500 diverse indoor scenes with ground truth cuboid and non-cuboid layout annotations. We set aside the first 3,000 scenes (corresponding to 18,362 images) for training, and equally split the last 500 scenes into 250 validation (1,776 images) and 250 test (1,697 images) sets. We pre-train HorizonNet on 18,362 synthetic images and perform transfer learning on MatterportLayout via supervised fine-tuning. In our semi-supervised experiments, we leverage HorizonNet pre-trained on Structured3D as a strong layout predictor initialized with synthetic data to further evaluate SSLayout360 on MatterportLayout.

\subsection{Training Epochs}
\label{training}
\Cref{epochs} summarizes the number of training epochs we run in our supervised and semi-supervised experiments, expressed as a function of labeled examples for each dataset under consideration. In the supervised setting, we define an epoch as one pass over available labeled examples in $\mathcal{D}_L$. In the semi-supervised setting, an epoch is defined as one pass over all unlabeled examples in $\mathcal{D}_U$. In short, we train our models between 50 and 300 epochs, depending on the setting and how many unlabeled instances are used in combination with labeled examples.

For the \texttt{Structured3D FT} row, we first train HorizonNet on 18,362 synthetic images for 50 epochs, and then transfer the learned weights to evaluate on MatterportLayout via supervised fine-tuning with an additional 100 epochs when using 50, 100, 200, 400, or 1,650 labels. In the semi-supervised setting, we leverage pre-trained HorizonNet as a strong predictor and train SSLayout360 to convergence with only 50 epochs when using 50, 100, 200, 400, or 1,650 labels. The reduced number of training epochs is an expected benefit when performing transfer learning.

Lastly, (not shown in \Cref{epochs}) in MatterportLayout experiments with 4,000 unlabeled images, we train SSLayout360 for 100 epochs when using 100, 200, 400 labels, and 300 epochs using 1,650 labels. For MatterportLayout experiments with 10,454 unlabeled images, we train SSLayout360 for 100 epochs when using 100, 200, 400, or 1,650 labels.

\renewcommand{\thetable}{B1}
\begin{table}[t]
\centering
\resizebox{\columnwidth}{!}{
    \begin{tabular} {lrrrr}
	\toprule
	Dataset & 
	\# Training &
    \# Validation &
	\# Testing &
	\# Unlabeled \\
    \midrule
    PanoContext \cite{panocontext}
	            & $413$
				& $46$
                & $53$
				& $1,009$ \\
    Stanford 2D-3D \cite{stanford-2d3d}
				& $404$
				& $33$
                & $113$
				& $949$ \\
	MatterportLayout \cite{layoutnetv2}
				& $1,650$
				& $187$
                & $458$
				& $1,837$ \\
	Structured3D \cite{structured3d}
				& $18,362$
				& $1,776$
                & $1,697$
				& -- \\
    \bottomrule
	\end{tabular}
	} \smallskip
{\caption{Summary statistics of the datasets used in the empirical evaluation. See text for more details.}
\label{datasets}}
\end{table}

\renewcommand{\thetable}{B2}
\begin{table}[t]
\centering
\resizebox{\columnwidth}{!}{
	\begin{tabular} {lrrrrrr}
	\toprule
	\multicolumn{2}{l}{Cuboid Layout}
	& \multicolumn{3}{c}{Training Epochs} \\
	\cmidrule{2-6}
	\multicolumn{1}{l}{\multirow{1}{*}{Dataset}} &
	\multicolumn{1}{r}{\multirow{1}{*}{20 labels}} &
	\multicolumn{1}{r}{\multirow{1}{*}{50 labels}} &
	\multicolumn{1}{r}{\multirow{1}{*}{100 labels}} &
	\multicolumn{1}{r}{\multirow{1}{*}{200 labels}} &
	\multicolumn{1}{r}{\multirow{1}{*}{All labels}} \\
    \midrule
    PanoContext
                & 200 (50)
				& 200 (100)
				& 200 (100)
				& 200 (100)
				& 300 (300) \\
	Stanford 2D-3D
                & 200 (50)
				& 200 (100)
				& 200 (100)
				& 200 (100)
				& 300 (300) \\
	\end{tabular}
	}
\resizebox{\columnwidth}{!}{
	\begin{tabular} {lrrrrrr}
	\toprule
	\multicolumn{2}{l}{Non-Cuboid Layout}
	& \multicolumn{3}{c}{Training Epochs} \\
	\cmidrule{2-6}
	\multicolumn{1}{l}{\multirow{1}{*}{Dataset}} &
	\multicolumn{1}{r}{\multirow{1}{*}{50 labels}} &
	\multicolumn{1}{r}{\multirow{1}{*}{100 labels}} &
	\multicolumn{1}{r}{\multirow{1}{*}{200 labels}} &
	\multicolumn{1}{r}{\multirow{1}{*}{400 labels}} &
	\multicolumn{1}{r}{\multirow{1}{*}{1,650 labels}} \\
    \midrule
    Structured3D FT
                & 100 (50)
				& 100 (50)
				& 100 (50)
				& 100 (50)
				& 100 (50) \\
	MatterportLayout
                & 200 (50)
				& 200 (100)
				& 200 (100)
				& 300 (100)
				& 300 (300) \\
    \bottomrule
	\end{tabular}
	} \smallskip
{\caption{The number of training epochs as a function of labeled examples for each dataset under consideration across both supervised and semi-supervised settings. The number in parenthesis indicates the semi-supervised counterpart. \texttt{All labels} refers to 963 labels for PanoContext and 916 labels for Stanford 2D-3D. \texttt{Structured3D FT} refers to supervised and semi-supervised fine-tuning experiments using the Structured3D synthetic dataset.}
\label{epochs}}
\end{table}

\subsection{Hyper-Parameters}
\Cref{parameters} summarizes the shared hyper-parameters as a function of dataset for both supervised and semi-supervised settings. Not shown in \Cref{parameters} are three hyper-parameters specific to our SSLayout360 algorithm, which are kept constant in all semi-supervised experiments: the unsupervised (or consistency) loss weight $\lambda = 1$, the ramp-up period $T@30\%$, and EMA decay coefficient $\alpha = 0.999$. We anneal the learning rate hyper-parameter after each training step $t$ according to the polynomial schedule: $\text{lr} \times \left(1 - \nicefrac{t}{t_\text{max}}\right)^{0.5}$. In general, we fix the hyper-parameters constant and do not attempt to tune them on a per-dataset or per-experiment basis, which can limit the real-world applicability of our method.

\renewcommand{\thetable}{B3}
\begin{table}[t]
\centering
\resizebox{\columnwidth}{!}{
    \begin{tabular} {lrrr}
	\toprule
	Dataset & 
	Batch Size &
    Learning Rate &
	Adam $\beta_1, \beta_2$ \cite{adam} \\
    \midrule
    PanoContext
	            & 8 (8)
				& 0.0003
                & 0.9, 0.999 \\
    Stanford 2D-3D
				& 8 (8)
				& 0.0003
                & 0.9, 0.999 \\
	Structured3D
				& 8 (--)
				& 0.0003
                & 0.9, 0.999 \\
	MatterportLayout
				& 4 (4)
				& 0.0001
                & 0.9, 0.999 \\
    Extra Unlabeled
                & 4 (12)
                & 0.0001
                & 0.9, 0.999 \\
    \bottomrule
	\end{tabular}
	} \smallskip
{\caption{Shared training hyper-parameters as a function of dataset for both supervised and semi-supervised settings. The \texttt{Batch Size} column indicates the blend of labeled (unlabeled) images for semi-supervised experiments or just labeled images for supervised experiments. In the \texttt{Extra Unlabeled} experiments, we use a mini-batch with a mixture of 4 labeled and 12 unlabeled examples to accelerate model training.}
\label{parameters}}
\end{table}

\subsection{Model and Data Perturbation}
\label{perturbation}
SSLayout360 relies on two sources of noise perturbation for its success: random model dropout \cite{dropout} and input data augmentation. In conjunction with \Cref{algorithm}, we employ the following procedure to obtain consistent student-teacher predictions for semi-supervised indoor layout estimation:
\begin{enumerate}
    \itemsep0em
    \item Separate the input data source into labeled and unlabeled branches. The unlabeled branch consists of all available training and validation examples, but without ground truth label information.
    \item Apply random data augmentation consisting of panoramic stretching \cite{horizonnet}, horizontal rotation, left-right flipping, and gamma correction to the labeled branch as input to the \emph{student} model.
    \item Apply the same set of data augmentation to the unlabeled branch, but without gamma correction and with a different random seed, as another set of input to the \emph{student} model.
    \item Perturb the resulting output of Step 3 with gamma correction as (noisy) input to the \emph{teacher} model.
    \item Enforce the consistency constraint on the student-teacher model outputs from Steps 3-4 per \Cref{algorithm} and \Cref{unsupervised-loss}.
\end{enumerate}

With this procedure, we introduce noise perturbation to the teacher's unlabeled input via gamma correction with $\gamma \in [0.5, 2]$. We also rely on random dropout with probability $0.5$ at each forward pass for additional model perturbation to help regularize the teacher's unsupervised targets.

\renewcommand{\thetable}{C1}
\begin{table*}[t]
\begin{minipage}[t]{0.5\textwidth}
\centering
\resizebox{\columnwidth}{!}{
	\begin{tabular} {lrrrrrr}
	\toprule
	\multicolumn{2}{l}{\textbf{PanoContext}}
	& \multicolumn{3}{c}{3D IoU (\%) $\uparrow$} \\
	\cmidrule{2-6}
	\multicolumn{1}{l}{\multirow{2}{*}{Method}} &
	\multicolumn{1}{c}{\multirow{1}{*}{20 labels}} &
	\multicolumn{1}{c}{\multirow{1}{*}{50 labels}} &
	\multicolumn{1}{c}{\multirow{1}{*}{100 labels}} &
	\multicolumn{1}{c}{\multirow{1}{*}{200 labels}} &
	\multicolumn{1}{c}{\multirow{1}{*}{963 labels}} \\
    & \multicolumn{1}{c}{\multirow{1}{*}{~~20 images}}
    & \multicolumn{1}{c}{\multirow{1}{*}{~~50 images}}
    & \multicolumn{1}{c}{\multirow{1}{*}{~~100 images}}
    & \multicolumn{1}{c}{\multirow{1}{*}{~~200 images}}
    & \multicolumn{1}{c}{\multirow{1}{*}{~~963 images}} \\
    \midrule
    HorizonNet
                & $61.48 \pm 2.07$
				& $63.84 \pm 2.87$
				& $65.43 \pm 1.30$
				& $75.76 \pm 0.62$
				& $83.55 \pm 0.31$ \\
	SSLayout360
                & $62.21 \pm 1.65$
				& $67.15 \pm 1.25$
				& $69.14 \pm 1.23$
				& $77.55 \pm 0.89$
				& $82.56 \pm 1.05$ \\
	\end{tabular}
	}
\resizebox{\columnwidth}{!}{
	\begin{tabular} {lrrrrrr}
	\toprule
	& \multicolumn{5}{c}{Corner Error (\%) $\downarrow$} \\
	\cmidrule{2-6}
	\multicolumn{1}{l}{\multirow{2}{*}{Method}} &
	\multicolumn{1}{c}{\multirow{1}{*}{20 labels}} &
	\multicolumn{1}{c}{\multirow{1}{*}{50 labels}} &
	\multicolumn{1}{c}{\multirow{1}{*}{100 labels}} &
	\multicolumn{1}{c}{\multirow{1}{*}{200 labels}} &
	\multicolumn{1}{c}{\multirow{1}{*}{963 labels}} \\
    & \multicolumn{1}{c}{\multirow{1}{*}{~~20 images}}
    & \multicolumn{1}{c}{\multirow{1}{*}{~~50 images}}
    & \multicolumn{1}{c}{\multirow{1}{*}{~~100 images}}
    & \multicolumn{1}{c}{\multirow{1}{*}{~~200 images}}
    & \multicolumn{1}{c}{\multirow{1}{*}{~~963 images}} \\
    \midrule
    HorizonNet
                & $3.51 \pm 0.79$
				& $2.78 \pm 0.90$
				& $3.17 \pm 0.27$
				& $1.07 \pm 0.15$
				& $0.70 \pm 0.02$ \\
	SSLayout360
                & $3.09 \pm 0.55$
				& $2.37 \pm 0.56$
				& $2.02 \pm 0.40$
				& $1.06 \pm 0.27$
				& $0.75 \pm 0.09$ \\
	\end{tabular}
	}
\resizebox{\columnwidth}{!}{
	\begin{tabular} {lrrrrrr}
	\toprule
	& \multicolumn{5}{c}{Pixel Error (\%) $\downarrow$} \\
	\cmidrule{2-6}
	\multicolumn{1}{l}{\multirow{2}{*}{Method}} &
	\multicolumn{1}{c}{\multirow{1}{*}{20 labels}} &
	\multicolumn{1}{c}{\multirow{1}{*}{50 labels}} &
	\multicolumn{1}{c}{\multirow{1}{*}{100 labels}} &
	\multicolumn{1}{c}{\multirow{1}{*}{200 labels}} &
	\multicolumn{1}{c}{\multirow{1}{*}{963 labels}} \\
    & \multicolumn{1}{c}{\multirow{1}{*}{~~20 images}}
    & \multicolumn{1}{c}{\multirow{1}{*}{~~50 images}}
    & \multicolumn{1}{c}{\multirow{1}{*}{~~100 images}}
    & \multicolumn{1}{c}{\multirow{1}{*}{~~200 images}}
    & \multicolumn{1}{c}{\multirow{1}{*}{~~963 images}} \\
    \midrule
    HorizonNet
                & $5.68 \pm 0.52$
				& $5.03 \pm 0.45$
				& $4.75 \pm 0.06$
				& $3.17 \pm 0.17$
				& $1.97 \pm 0.03$ \\
	SSLayout360
                & $5.52 \pm 0.48$
				& $4.35 \pm 0.36$
				& $4.49 \pm 0.28$
				& $3.03 \pm 0.53$
				& $2.03 \pm 0.19$ \\
    \bottomrule
	\end{tabular}
	}
\end{minipage}
\begin{minipage}[t]{0.5\textwidth}
\centering
\resizebox{\columnwidth}{!}{
	\begin{tabular} {lrrrrrr}
	\toprule
	\multicolumn{2}{l}{\textbf{Stanford 2D-3D}}
	& \multicolumn{3}{c}{3D IoU (\%) $\uparrow$} \\
	\cmidrule{2-6}
	\multicolumn{1}{l}{\multirow{2}{*}{Method}} &
	\multicolumn{1}{c}{\multirow{1}{*}{20 labels}} &
	\multicolumn{1}{c}{\multirow{1}{*}{50 labels}} &
	\multicolumn{1}{c}{\multirow{1}{*}{100 labels}} &
	\multicolumn{1}{c}{\multirow{1}{*}{200 labels}} &
	\multicolumn{1}{c}{\multirow{1}{*}{916 labels}} \\
    & \multicolumn{1}{c}{\multirow{1}{*}{~~20 images}}
    & \multicolumn{1}{c}{\multirow{1}{*}{~~50 images}}
    & \multicolumn{1}{c}{\multirow{1}{*}{~~100 images}}
    & \multicolumn{1}{c}{\multirow{1}{*}{~~200 images}}
    & \multicolumn{1}{c}{\multirow{1}{*}{~~916 images}} \\
    \midrule
    HorizonNet
                & $62.20 \pm 3.98$
                & $68.27 \pm 1.45$
                & $69.94 \pm 3.64$
                & $74.95 \pm 3.69$
				& $82.79 \pm 0.90$ \\
	SSLayout360
                & $69.19 \pm 2.34$
				& $70.66 \pm 1.62$
				& $74.21 \pm 1.33$
				& $77.75 \pm 1.30$
				& $84.54 \pm 0.59$ \\
	\end{tabular}
	}
\resizebox{\columnwidth}{!}{
	\begin{tabular} {lrrrrrr}
	\toprule
	& \multicolumn{5}{c}{Corner Error (\%) $\downarrow$} \\
	\cmidrule{2-6}
	\multicolumn{1}{l}{\multirow{2}{*}{Method}} &
	\multicolumn{1}{c}{\multirow{1}{*}{20 labels}} &
	\multicolumn{1}{c}{\multirow{1}{*}{50 labels}} &
	\multicolumn{1}{c}{\multirow{1}{*}{100 labels}} &
	\multicolumn{1}{c}{\multirow{1}{*}{200 labels}} &
	\multicolumn{1}{c}{\multirow{1}{*}{916 labels}} \\
    & \multicolumn{1}{c}{\multirow{1}{*}{~~20 images}}
    & \multicolumn{1}{c}{\multirow{1}{*}{~~50 images}}
    & \multicolumn{1}{c}{\multirow{1}{*}{~~100 images}}
    & \multicolumn{1}{c}{\multirow{1}{*}{~~200 images}}
    & \multicolumn{1}{c}{\multirow{1}{*}{~~916 images}} \\
    \midrule
    HorizonNet
                & $2.70 \pm 0.60$
                & $1.64 \pm 0.12$
                & $1.66 \pm 0.20$
                & $1.50 \pm 0.18$
				& $0.64 \pm 0.02$ \\
	SSLayout360
                & $2.19 \pm 0.61$
                & $1.79 \pm 0.42$
                & $1.43 \pm 0.16$
                & $1.13 \pm 0.15$
				& $0.63 \pm 0.02$ \\
	\end{tabular}
	}
\resizebox{\columnwidth}{!}{
	\begin{tabular} {lrrrrrr}
	\toprule
	& \multicolumn{5}{c}{Pixel Error (\%) $\downarrow$} \\
	\cmidrule{2-6}
	\multicolumn{1}{l}{\multirow{2}{*}{Method}} &
	\multicolumn{1}{c}{\multirow{1}{*}{20 labels}} &
	\multicolumn{1}{c}{\multirow{1}{*}{50 labels}} &
	\multicolumn{1}{c}{\multirow{1}{*}{100 labels}} &
	\multicolumn{1}{c}{\multirow{1}{*}{200 labels}} &
	\multicolumn{1}{c}{\multirow{1}{*}{916 labels}} \\
    & \multicolumn{1}{c}{\multirow{1}{*}{~~20 images}}
    & \multicolumn{1}{c}{\multirow{1}{*}{~~50 images}}
    & \multicolumn{1}{c}{\multirow{1}{*}{~~100 images}}
    & \multicolumn{1}{c}{\multirow{1}{*}{~~200 images}}
    & \multicolumn{1}{c}{\multirow{1}{*}{~~916 images}} \\
    \midrule
    HorizonNet
                & $5.03 \pm 0.51$
                & $3.95 \pm 0.22$
                & $3.77 \pm 0.39$
                & $3.69 \pm 0.26$
				& $2.13 \pm 0.05$ \\
	SSLayout360
                & $4.24 \pm 0.41$
                & $4.07 \pm 0.15$
                & $3.54 \pm 0.11$
                & $3.27 \pm 0.35$
				& $1.99 \pm 0.04$ \\
    \bottomrule
	\end{tabular}
	} \smallskip
\end{minipage}
{\caption{Quantitative cuboid layout results, without the use of unlabeled data, evaluated on the \textbf{PanoContext (left)} and \textbf{Stanford 2D-3D (right)} test sets averaged over four independent runs with different random seeds. Number format: mean value $\pm$ standard deviation.}
\label{cuboid-no-unlabeled}}
\end{table*}

\renewcommand{\thetable}{C2}
\begin{table}[t]
\centering
\resizebox{\columnwidth}{!}{
	\begin{tabular} {lrrrrrr}
	\toprule
	\multicolumn{2}{l}{\textbf{MatterportLayout (4-18 Corners)}}
	& \multicolumn{3}{c}{~~3D IoU (\%) $\uparrow$} \\
	\cmidrule{2-6}
	\multicolumn{1}{l}{\multirow{2}{*}{Method}} &
	\multicolumn{1}{c}{\multirow{1}{*}{50 labels}} &
	\multicolumn{1}{c}{\multirow{1}{*}{100 labels}} &
	\multicolumn{1}{c}{\multirow{1}{*}{200 labels}} &
	\multicolumn{1}{c}{\multirow{1}{*}{400 labels}} &
	\multicolumn{1}{c}{\multirow{1}{*}{1,650 labels}} \\
    & \multicolumn{1}{c}{\multirow{1}{*}{~~50 images}}
    & \multicolumn{1}{c}{\multirow{1}{*}{~~100 images}}
    & \multicolumn{1}{c}{\multirow{1}{*}{~~200 images}}
    & \multicolumn{1}{c}{\multirow{1}{*}{~~400 images}}
    & \multicolumn{1}{c}{\multirow{1}{*}{~~1,650 images}} \\
    \midrule
    HorizonNet
                & $63.44 \pm 0.56$
				& $68.79 \pm 0.49$
				& $72.25 \pm 0.50$
				& $74.46 \pm 0.35$
				& $79.12 \pm 0.37$ \\
	SSLayout360
                & $63.21 \pm 0.73$
				& $68.71 \pm 0.22$
				& $73.01 \pm 0.24$
				& $75.38 \pm 0.72$
				& $80.06 \pm 0.26$ \\
	\end{tabular}
	}
\resizebox{\columnwidth}{!}{
	\begin{tabular} {lrrrrrr}
	\toprule
	& \multicolumn{5}{c}{2D IoU (\%) $\uparrow$} \\
	\cmidrule{2-6}
	\multicolumn{1}{l}{\multirow{2}{*}{Method}} &
	\multicolumn{1}{c}{\multirow{1}{*}{50 labels}} &
	\multicolumn{1}{c}{\multirow{1}{*}{100 labels}} &
	\multicolumn{1}{c}{\multirow{1}{*}{200 labels}} &
	\multicolumn{1}{c}{\multirow{1}{*}{400 labels}} &
	\multicolumn{1}{c}{\multirow{1}{*}{1,650 labels}} \\
    & \multicolumn{1}{c}{\multirow{1}{*}{~~50 images}}
    & \multicolumn{1}{c}{\multirow{1}{*}{~~100 images}}
    & \multicolumn{1}{c}{\multirow{1}{*}{~~200 images}}
    & \multicolumn{1}{c}{\multirow{1}{*}{~~400 images}}
    & \multicolumn{1}{c}{\multirow{1}{*}{~~1,650 images}} \\
    \midrule
    HorizonNet
                & $67.17 \pm 0.65$
				& $72.06 \pm 0.49$
				& $75.16 \pm 0.53$
				& $77.15 \pm 0.36$
				& $81.54 \pm 0.31$ \\
	SSLayout360
                & $66.76 \pm 0.74$
				& $71.83 \pm 0.24$
				& $75.78 \pm 0.26$
				& $77.99 \pm 0.70$
				& $82.35 \pm 0.29$ \\
	\end{tabular}
	}
\resizebox{\columnwidth}{!}{
	\begin{tabular} {lrrrrrr}
	\toprule
	& \multicolumn{5}{c}{$\delta_1$ $\uparrow$} \\
	\cmidrule{2-6}
	\multicolumn{1}{l}{\multirow{2}{*}{Method}} &
	\multicolumn{1}{c}{\multirow{1}{*}{50 labels}} &
	\multicolumn{1}{c}{\multirow{1}{*}{100 labels}} &
	\multicolumn{1}{c}{\multirow{1}{*}{200 labels}} &
	\multicolumn{1}{c}{\multirow{1}{*}{400 labels}} &
	\multicolumn{1}{c}{\multirow{1}{*}{1,650 labels}} \\
    & \multicolumn{1}{c}{\multirow{1}{*}{~~50 images}}
    & \multicolumn{1}{c}{\multirow{1}{*}{~~100 images}}
    & \multicolumn{1}{c}{\multirow{1}{*}{~~200 images}}
    & \multicolumn{1}{c}{\multirow{1}{*}{~~400 images}}
    & \multicolumn{1}{c}{\multirow{1}{*}{~~1,650 images}} \\
    \midrule
    HorizonNet
                & $0.76 \pm 0.01$
				& $0.84 \pm 0.01$
				& $0.89 \pm 0.01$
				& $0.91 \pm 0.01$
				& $0.94 \pm 0.01$ \\
	SSLayout360
                & $0.77 \pm 0.01$
				& $0.85 \pm 0.01$
				& $0.89 \pm 0.01$
				& $0.91 \pm 0.01$
				& $0.95 \pm 0.01$ \\
	\end{tabular}
	}
\resizebox{\columnwidth}{!}{
	\begin{tabular} {lrrrrrr}
	\toprule
	& \multicolumn{5}{c}{RMSE $\downarrow$} \\
	\cmidrule{2-6}
	\multicolumn{1}{l}{\multirow{2}{*}{Method}} &
	\multicolumn{1}{c}{\multirow{1}{*}{50 labels}} &
	\multicolumn{1}{c}{\multirow{1}{*}{100 labels}} &
	\multicolumn{1}{c}{\multirow{1}{*}{200 labels}} &
	\multicolumn{1}{c}{\multirow{1}{*}{400 labels}} &
	\multicolumn{1}{c}{\multirow{1}{*}{1,650 labels}} \\
    & \multicolumn{1}{c}{\multirow{1}{*}{~~50 images}}
    & \multicolumn{1}{c}{\multirow{1}{*}{~~100 images}}
    & \multicolumn{1}{c}{\multirow{1}{*}{~~200 images}}
    & \multicolumn{1}{c}{\multirow{1}{*}{~~400 images}}
    & \multicolumn{1}{c}{\multirow{1}{*}{~~1,650 images}} \\
    \midrule
    HorizonNet
                & $0.41 \pm 0.01$
				& $0.34 \pm 0.01$
				& $0.30 \pm 0.01$
				& $0.28 \pm 0.01$
				& $0.23 \pm 0.01$ \\
	SSLayout360
                & $0.41 \pm 0.01$
				& $0.34 \pm 0.01$
				& $0.29 \pm 0.01$
				& $0.27 \pm 0.02$
				& $0.23 \pm 0.01$ \\
    \bottomrule
	\end{tabular}
	} \smallskip
{\caption{Quantitative non-cuboid layout results, without the use of unlabeled data, evaluated on the MatterportLayout test set with 4-18 corners averaged over four independent runs with different random seeds. Number format: mean value $\pm$ standard deviation.}
\label{non-cuboid-no-unlabeled}}
\end{table}

\section{SSLayout360 without Unlabeled Data}
\label{no-unlabeled}
Tables \ref{cuboid-no-unlabeled} -- \ref{non-cuboid-no-unlabeled} compare the fully supervised HorizonNet with SSLayout360 without the use of unlabeled data. This setting corresponds to fully supervised learning with consistency regularization. In this setting, SSLayout360 only uses the available labeled training examples as the source of ``unlabeled data'' to generate unsupervised proxy targets for enforcing the consistency constraint. We observe that for PanoContext and Stanford 2D-3D in Table~\ref{cuboid-no-unlabeled}, SSLayout360 without unlabeled data produces slightly better results than the supervised HorizonNet counterpart across most settings and metrics. Similarly, for MatterportLayout in Table~\ref{non-cuboid-no-unlabeled}, SSLayout360 gives competitive or slightly better results than the supervised HorizonNet counterpart. Our findings from these experiments corroborate previous SSL literature that regularization can slightly improve supervised learning without unlabeled data \cite{vat2,tran-ssl,s4l}. In scenarios with additional unlabeled data, SSLayout360 provides a significant boost in accuracy performance over the supervised baselines.

\section{Additional MatterportLayout Results}
\label{more-mp3d}
We report supervised and semi-supervised MatterportLayout results for rooms having 4, 6, 8, and 10 or more corners in Tables \ref{4-6-corners} -- \ref{8-10+-corners}. These results show that the effective use of unlabeled data, in combination with labeled data, improves room layout estimation with increasing scene complexity across most settings and metrics under consideration.

\section{Qualitative 3D Layout Reconstruction}
\label{3d-layout}
Figures \ref{qualitative01} -- \ref{qualitative03} present qualitative 3D layout reconstruction results on select MatterportLayout test instances with increasing layout complexity, ranging between 6 and 12 corners, using the post-processing algorithm proposed by Sun \etal of HorizonNet \cite{horizonnet}. These results are obtained from our SSLayout360 model trained on 100 labeled and 4,000 unlabeled images. Using only 6\% of the available 1,650 labels, our SSLayout360 model is able to produce convincing 3D layout reconstruction of complex indoor scenes.

\renewcommand{\thetable}{D1}
\begin{table*}[ht]
\begin{minipage}[t]{0.5\textwidth}
\centering
\resizebox{\columnwidth}{!}{
	\begin{tabular} {lrrrrrr}
	\toprule
	\multicolumn{2}{l}{\textbf{MatterportLayout (4 Corners)}}
	& \multicolumn{3}{c}{~~3D IoU (\%) $\uparrow$} \\
	\cmidrule{2-6}
	\multicolumn{1}{l}{\multirow{2}{*}{Method}} &
	\multicolumn{1}{c}{\multirow{1}{*}{~~~~50 labels}} &
	\multicolumn{1}{c}{\multirow{1}{*}{~~~100 labels}} &
	\multicolumn{1}{c}{\multirow{1}{*}{~~~200 labels}} &
	\multicolumn{1}{c}{\multirow{1}{*}{~~~400 labels}} &
	\multicolumn{1}{c}{\multirow{1}{*}{~~1,650 labels}} \\
    & \multicolumn{1}{c}{\multirow{1}{*}{~~1,837 images}}
    & \multicolumn{1}{c}{\multirow{1}{*}{~~1,837 images}}
    & \multicolumn{1}{c}{\multirow{1}{*}{~~1,837 images}}
    & \multicolumn{1}{c}{\multirow{1}{*}{~~1,837 images}}
    & \multicolumn{1}{c}{\multirow{1}{*}{~~~~1,837 images}} \\
    \midrule
    HorizonNet
                & $63.21 \pm 0.59$
				& $69.09 \pm 0.53$
				& $73.62 \pm 0.57$
				& $76.10 \pm 0.52$
				& $81.41 \pm 0.58$ \\
	SSLayout360
                & \bfseries 68.07 $\pm$ 0.23
				& \bfseries 73.91 $\pm$ 0.44
				& \bfseries 77.31 $\pm$ 0.40
				& \bfseries 79.37 $\pm$ 0.41
				& \bfseries 82.81 $\pm$ 0.72 \\
	\end{tabular}
	}
\resizebox{\columnwidth}{!}{
	\begin{tabular} {lrrrrrr}
	\toprule
	& \multicolumn{5}{c}{2D IoU (\%) $\uparrow$} \\
	\cmidrule{2-6}
	\multicolumn{1}{l}{\multirow{2}{*}{Method}} &
	\multicolumn{1}{c}{\multirow{1}{*}{~~~~50 labels}} &
	\multicolumn{1}{c}{\multirow{1}{*}{~~~100 labels}} &
	\multicolumn{1}{c}{\multirow{1}{*}{~~~200 labels}} &
	\multicolumn{1}{c}{\multirow{1}{*}{~~~400 labels}} &
	\multicolumn{1}{c}{\multirow{1}{*}{~~1,650 labels}} \\
    & \multicolumn{1}{c}{\multirow{1}{*}{~~1,837 images}}
    & \multicolumn{1}{c}{\multirow{1}{*}{~~1,837 images}}
    & \multicolumn{1}{c}{\multirow{1}{*}{~~1,837 images}}
    & \multicolumn{1}{c}{\multirow{1}{*}{~~1,837 images}}
    & \multicolumn{1}{c}{\multirow{1}{*}{~~~~1,837 images}} \\
    \midrule
    HorizonNet
                & $67.22 \pm 0.68$
				& $72.63 \pm 0.52$
				& $76.78 \pm 0.61$
				& $78.99 \pm 0.48$
				& $84.02 \pm 0.48$ \\
	SSLayout360
                & \bfseries 71.99 $\pm$ 0.28
				& \bfseries 77.31 $\pm$ 0.42
				& \bfseries 80.28 $\pm$ 0.40
				& \bfseries 82.18 $\pm$ 0.40
				& \bfseries 85.22 $\pm$ 0.80 \\
	\end{tabular}
	}
\resizebox{\columnwidth}{!}{
	\begin{tabular} {lrrrrrr}
	\toprule
	& \multicolumn{5}{c}{$\delta_1$ $\uparrow$} \\
	\cmidrule{2-6}
	\multicolumn{1}{l}{\multirow{2}{*}{Method}} &
	\multicolumn{1}{c}{\multirow{1}{*}{~~~~50 labels}} &
	\multicolumn{1}{c}{\multirow{1}{*}{~~~100 labels}} &
	\multicolumn{1}{c}{\multirow{1}{*}{~~~200 labels}} &
	\multicolumn{1}{c}{\multirow{1}{*}{~~~400 labels}} &
	\multicolumn{1}{c}{\multirow{1}{*}{~~1,650 labels}} \\
    & \multicolumn{1}{c}{\multirow{1}{*}{~~1,837 images}}
    & \multicolumn{1}{c}{\multirow{1}{*}{~~1,837 images}}
    & \multicolumn{1}{c}{\multirow{1}{*}{~~1,837 images}}
    & \multicolumn{1}{c}{\multirow{1}{*}{~~1,837 images}}
    & \multicolumn{1}{c}{\multirow{1}{*}{~~~~1,837 images}} \\
    \midrule
    HorizonNet
                & $0.75 \pm 0.01$
				& $0.83 \pm 0.01$
				& $0.88 \pm 0.01$
				& $0.91 \pm 0.01$
				& \bfseries 0.95 $\pm$ 0.01 \\
	SSLayout360
                & \bfseries 0.80 $\pm$ 0.01
				& \bfseries 0.88 $\pm$ 0.01
				& \bfseries 0.91 $\pm$ 0.01
				& \bfseries 0.93 $\pm$ 0.01
				& \bfseries 0.96 $\pm$ 0.01 \\
	\end{tabular}
	}
\resizebox{\columnwidth}{!}{
	\begin{tabular} {lrrrrrr}
	\toprule
	& \multicolumn{5}{c}{RMSE $\downarrow$} \\
	\cmidrule{2-6}
	\multicolumn{1}{l}{\multirow{2}{*}{Method}} &
	\multicolumn{1}{c}{\multirow{1}{*}{~~~~50 labels}} &
	\multicolumn{1}{c}{\multirow{1}{*}{~~~100 labels}} &
	\multicolumn{1}{c}{\multirow{1}{*}{~~~200 labels}} &
	\multicolumn{1}{c}{\multirow{1}{*}{~~~400 labels}} &
	\multicolumn{1}{c}{\multirow{1}{*}{~~1,650 labels}} \\
    & \multicolumn{1}{c}{\multirow{1}{*}{~~1,837 images}}
    & \multicolumn{1}{c}{\multirow{1}{*}{~~1,837 images}}
    & \multicolumn{1}{c}{\multirow{1}{*}{~~1,837 images}}
    & \multicolumn{1}{c}{\multirow{1}{*}{~~1,837 images}}
    & \multicolumn{1}{c}{\multirow{1}{*}{~~~~1,837 images}} \\
    \midrule
    HorizonNet
                & $0.40 \pm 0.01$
				& $0.31 \pm 0.01$
				& $0.27 \pm 0.02$
				& $0.25 \pm 0.02$
				& \bfseries 0.18 $\pm$ 0.01 \\
	SSLayout360
                & \bfseries 0.33 $\pm$ 0.01
				& \bfseries 0.26 $\pm$ 0.01
				& \bfseries 0.23 $\pm$ 0.01
				& \bfseries 0.20 $\pm$ 0.01
				& \bfseries 0.17 $\pm$ 0.01 \\
    \bottomrule
	\end{tabular}
	}
\end{minipage}
\begin{minipage}[t]{0.5\textwidth}
\centering
\resizebox{\columnwidth}{!}{
	\begin{tabular} {lrrrrrr}
	\toprule
	\multicolumn{2}{l}{\textbf{MatterportLayout (6 Corners)}}
	& \multicolumn{3}{c}{~~3D IoU (\%) $\uparrow$} \\
	\cmidrule{2-6}
	\multicolumn{1}{l}{\multirow{2}{*}{Method}} &
	\multicolumn{1}{c}{\multirow{1}{*}{~~~~50 labels}} &
	\multicolumn{1}{c}{\multirow{1}{*}{~~~100 labels}} &
	\multicolumn{1}{c}{\multirow{1}{*}{~~~200 labels}} &
	\multicolumn{1}{c}{\multirow{1}{*}{~~~400 labels}} &
	\multicolumn{1}{c}{\multirow{1}{*}{~~1,650 labels}} \\
    & \multicolumn{1}{c}{\multirow{1}{*}{~~1,837 images}}
    & \multicolumn{1}{c}{\multirow{1}{*}{~~1,837 images}}
    & \multicolumn{1}{c}{\multirow{1}{*}{~~1,837 images}}
    & \multicolumn{1}{c}{\multirow{1}{*}{~~1,837 images}}
    & \multicolumn{1}{c}{\multirow{1}{*}{~~~~1,837 images}} \\
    \midrule
    HorizonNet
                & $67.67 \pm 0.60$
				& $73.31 \pm 0.82$
				& $77.23 \pm 0.58$
				& $78.04 \pm 0.34$
				& $81.93 \pm 0.55$ \\
	SSLayout360
                & \bfseries 70.48 $\pm$ 0.35
				& \bfseries 76.38 $\pm$ 0.58
				& \bfseries 79.55 $\pm$ 0.52
				& \bfseries 80.32 $\pm$ 0.24
				& \bfseries 82.82 $\pm$ 0.68 \\
	\end{tabular}
	}
\resizebox{\columnwidth}{!}{
	\begin{tabular} {lrrrrrr}
	\toprule
	& \multicolumn{5}{c}{2D IoU (\%) $\uparrow$} \\
	\cmidrule{2-6}
	\multicolumn{1}{l}{\multirow{2}{*}{Method}} &
	\multicolumn{1}{c}{\multirow{1}{*}{~~~~50 labels}} &
	\multicolumn{1}{c}{\multirow{1}{*}{~~~100 labels}} &
	\multicolumn{1}{c}{\multirow{1}{*}{~~~200 labels}} &
	\multicolumn{1}{c}{\multirow{1}{*}{~~~400 labels}} &
	\multicolumn{1}{c}{\multirow{1}{*}{~~1,650 labels}} \\
    & \multicolumn{1}{c}{\multirow{1}{*}{~~1,837 images}}
    & \multicolumn{1}{c}{\multirow{1}{*}{~~1,837 images}}
    & \multicolumn{1}{c}{\multirow{1}{*}{~~1,837 images}}
    & \multicolumn{1}{c}{\multirow{1}{*}{~~1,837 images}}
    & \multicolumn{1}{c}{\multirow{1}{*}{~~~~1,837 images}} \\
    \midrule
    HorizonNet
                & $71.92 \pm 0.63$
				& $76.68 \pm 0.90$
				& $80.05 \pm 0.57$
				& $80.62 \pm 0.30$
				& $84.40 \pm 0.60$ \\
	SSLayout360
                & \bfseries 74.57 $\pm$ 0.37
				& \bfseries 79.67 $\pm$ 0.60
				& \bfseries 82.12 $\pm$ 0.60
				& \bfseries 82.86 $\pm$ 0.28
				& \bfseries 85.05 $\pm$ 0.81 \\
	\end{tabular}
	}
\resizebox{\columnwidth}{!}{
	\begin{tabular} {lrrrrrr}
	\toprule
	& \multicolumn{5}{c}{$\delta_1$ $\uparrow$} \\
	\cmidrule{2-6}
	\multicolumn{1}{l}{\multirow{2}{*}{Method}} &
	\multicolumn{1}{c}{\multirow{1}{*}{~~~~50 labels}} &
	\multicolumn{1}{c}{\multirow{1}{*}{~~~100 labels}} &
	\multicolumn{1}{c}{\multirow{1}{*}{~~~200 labels}} &
	\multicolumn{1}{c}{\multirow{1}{*}{~~~400 labels}} &
	\multicolumn{1}{c}{\multirow{1}{*}{~~1,650 labels}} \\
    & \multicolumn{1}{c}{\multirow{1}{*}{~~1,837 images}}
    & \multicolumn{1}{c}{\multirow{1}{*}{~~1,837 images}}
    & \multicolumn{1}{c}{\multirow{1}{*}{~~1,837 images}}
    & \multicolumn{1}{c}{\multirow{1}{*}{~~1,837 images}}
    & \multicolumn{1}{c}{\multirow{1}{*}{~~~~1,837 images}} \\
    \midrule
    HorizonNet
                & $0.76 \pm 0.01$
				& $0.87 \pm 0.01$
				& $0.93 \pm 0.01$
				& $0.94 \pm 0.01$
				& \bfseries 0.95 $\pm$ 0.02 \\
	SSLayout360
                & \bfseries 0.79 $\pm$ 0.01
				& \bfseries 0.90 $\pm$ 0.01
				& \bfseries 0.95 $\pm$ 0.01
				& \bfseries 0.95 $\pm$ 0.01
				& \bfseries 0.96 $\pm$ 0.01 \\
	\end{tabular}
	}
\resizebox{\columnwidth}{!}{
	\begin{tabular} {lrrrrrr}
	\toprule
	& \multicolumn{5}{c}{RMSE $\downarrow$} \\
	\cmidrule{2-6}
	\multicolumn{1}{l}{\multirow{2}{*}{Method}} &
	\multicolumn{1}{c}{\multirow{1}{*}{~~~~50 labels}} &
	\multicolumn{1}{c}{\multirow{1}{*}{~~~100 labels}} &
	\multicolumn{1}{c}{\multirow{1}{*}{~~~200 labels}} &
	\multicolumn{1}{c}{\multirow{1}{*}{~~~400 labels}} &
	\multicolumn{1}{c}{\multirow{1}{*}{~~1,650 labels}} \\
    & \multicolumn{1}{c}{\multirow{1}{*}{~~1,837 images}}
    & \multicolumn{1}{c}{\multirow{1}{*}{~~1,837 images}}
    & \multicolumn{1}{c}{\multirow{1}{*}{~~1,837 images}}
    & \multicolumn{1}{c}{\multirow{1}{*}{~~1,837 images}}
    & \multicolumn{1}{c}{\multirow{1}{*}{~~~~1,837 images}} \\
    \midrule
    HorizonNet
                & $0.35 \pm 0.01$
				& $0.28 \pm 0.02$
				& $0.24 \pm 0.01$
				& $0.23 \pm 0.01$
				& \bfseries 0.21 $\pm$ 0.03 \\
	SSLayout360
                & \bfseries 0.31 $\pm$ 0.01
				& \bfseries 0.24 $\pm$ 0.01
				& \bfseries 0.21 $\pm$ 0.01
				& \bfseries 0.21 $\pm$ 0.01
				& \bfseries 0.20 $\pm$ 0.02 \\
    \bottomrule
	\end{tabular}
	} \smallskip
\end{minipage}
{\caption{Quantitative layout results for \textbf{4 corners (left)} and \textbf{6 corners (right)} evaluated on the MatterportLayout test set averaged over four independent runs with different random seeds. Number format: mean value $\pm$ standard deviation.}
\label{4-6-corners}}
\end{table*}

\renewcommand{\thetable}{D2}
\begin{table*}[ht]
\begin{minipage}[t]{0.5\textwidth}
\centering
\resizebox{\columnwidth}{!}{
	\begin{tabular} {lrrrrrr}
	\toprule
	\multicolumn{2}{l}{\textbf{MatterportLayout (8 Corners)}}
	& \multicolumn{3}{c}{~~3D IoU (\%) $\uparrow$} \\
	\cmidrule{2-6}
	\multicolumn{1}{l}{\multirow{2}{*}{Method}} &
	\multicolumn{1}{c}{\multirow{1}{*}{~~~~50 labels}} &
	\multicolumn{1}{c}{\multirow{1}{*}{~~~100 labels}} &
	\multicolumn{1}{c}{\multirow{1}{*}{~~~200 labels}} &
	\multicolumn{1}{c}{\multirow{1}{*}{~~~400 labels}} &
	\multicolumn{1}{c}{\multirow{1}{*}{~~1,650 labels}} \\
    & \multicolumn{1}{c}{\multirow{1}{*}{~~1,837 images}}
    & \multicolumn{1}{c}{\multirow{1}{*}{~~1,837 images}}
    & \multicolumn{1}{c}{\multirow{1}{*}{~~1,837 images}}
    & \multicolumn{1}{c}{\multirow{1}{*}{~~1,837 images}}
    & \multicolumn{1}{c}{\multirow{1}{*}{~~~~1,837 images}} \\
    \midrule
    HorizonNet
                & $63.14 \pm 1.20$
				& $67.97 \pm 0.39$
				& $68.30 \pm 0.56$
				& $70.30 \pm 0.91$
				& \bfseries 74.54 $\pm$ 0.59 \\
	SSLayout360
                & \bfseries 65.97 $\pm$ 0.13
				& \bfseries 68.39 $\pm$ 0.63
				& \bfseries 69.96 $\pm$ 0.42
				& \bfseries 72.05 $\pm$ 0.82
				& \bfseries 74.51 $\pm$ 0.54 \\
	\end{tabular}
	}
\resizebox{\columnwidth}{!}{
	\begin{tabular} {lrrrrrr}
	\toprule
	& \multicolumn{5}{c}{2D IoU (\%) $\uparrow$} \\
	\cmidrule{2-6}
	\multicolumn{1}{l}{\multirow{2}{*}{Method}} &
	\multicolumn{1}{c}{\multirow{1}{*}{~~~~50 labels}} &
	\multicolumn{1}{c}{\multirow{1}{*}{~~~100 labels}} &
	\multicolumn{1}{c}{\multirow{1}{*}{~~~200 labels}} &
	\multicolumn{1}{c}{\multirow{1}{*}{~~~400 labels}} &
	\multicolumn{1}{c}{\multirow{1}{*}{~~1,650 labels}} \\
    & \multicolumn{1}{c}{\multirow{1}{*}{~~1,837 images}}
    & \multicolumn{1}{c}{\multirow{1}{*}{~~1,837 images}}
    & \multicolumn{1}{c}{\multirow{1}{*}{~~1,837 images}}
    & \multicolumn{1}{c}{\multirow{1}{*}{~~1,837 images}}
    & \multicolumn{1}{c}{\multirow{1}{*}{~~~~1,837 images}} \\
    \midrule
    HorizonNet
                & $66.29 \pm 1.31$
				& \bfseries 70.87 $\pm$ 0.34
				& $70.87 \pm 0.67$
				& $72.60 \pm 0.92$
				& \bfseries 76.58 $\pm$ 0.58 \\
	SSLayout360
                & \bfseries 68.94 $\pm$ 0.16
				& \bfseries 70.86 $\pm$ 0.62
				& \bfseries 72.38 $\pm$ 0.42
				& \bfseries 74.34 $\pm$ 0.80
				& \bfseries 76.31 $\pm$ 0.63 \\
	\end{tabular}
	}
\resizebox{\columnwidth}{!}{
	\begin{tabular} {lrrrrrr}
	\toprule
	& \multicolumn{5}{c}{$\delta_1$ $\uparrow$} \\
	\cmidrule{2-6}
	\multicolumn{1}{l}{\multirow{2}{*}{Method}} &
	\multicolumn{1}{c}{\multirow{1}{*}{~~~~50 labels}} &
	\multicolumn{1}{c}{\multirow{1}{*}{~~~100 labels}} &
	\multicolumn{1}{c}{\multirow{1}{*}{~~~200 labels}} &
	\multicolumn{1}{c}{\multirow{1}{*}{~~~400 labels}} &
	\multicolumn{1}{c}{\multirow{1}{*}{~~1,650 labels}} \\
    & \multicolumn{1}{c}{\multirow{1}{*}{~~1,837 images}}
    & \multicolumn{1}{c}{\multirow{1}{*}{~~1,837 images}}
    & \multicolumn{1}{c}{\multirow{1}{*}{~~1,837 images}}
    & \multicolumn{1}{c}{\multirow{1}{*}{~~1,837 images}}
    & \multicolumn{1}{c}{\multirow{1}{*}{~~~~1,837 images}} \\
    \midrule
    HorizonNet
                & $0.79 \pm 0.01$
				& $0.87 \pm 0.01$
				& $0.89 \pm 0.01$
				& $0.90 \pm 0.01$
				& \bfseries 0.94 $\pm$ 0.01 \\
	SSLayout360
                & \bfseries 0.83 $\pm$ 0.01
				& \bfseries 0.90 $\pm$ 0.01
				& \bfseries 0.90 $\pm$ 0.01
				& \bfseries 0.91 $\pm$ 0.01
				& \bfseries 0.94 $\pm$ 0.01 \\
	\end{tabular}
	}
\resizebox{\columnwidth}{!}{
	\begin{tabular} {lrrrrrr}
	\toprule
	& \multicolumn{5}{c}{RMSE $\downarrow$} \\
	\cmidrule{2-6}
	\multicolumn{1}{l}{\multirow{2}{*}{Method}} &
	\multicolumn{1}{c}{\multirow{1}{*}{~~~~50 labels}} &
	\multicolumn{1}{c}{\multirow{1}{*}{~~~100 labels}} &
	\multicolumn{1}{c}{\multirow{1}{*}{~~~200 labels}} &
	\multicolumn{1}{c}{\multirow{1}{*}{~~~400 labels}} &
	\multicolumn{1}{c}{\multirow{1}{*}{~~1,650 labels}} \\
    & \multicolumn{1}{c}{\multirow{1}{*}{~~1,837 images}}
    & \multicolumn{1}{c}{\multirow{1}{*}{~~1,837 images}}
    & \multicolumn{1}{c}{\multirow{1}{*}{~~1,837 images}}
    & \multicolumn{1}{c}{\multirow{1}{*}{~~1,837 images}}
    & \multicolumn{1}{c}{\multirow{1}{*}{~~~~1,837 images}} \\
    \midrule
    HorizonNet
                & $0.41 \pm 0.02$
				& \bfseries 0.34 $\pm$ 0.01
				& \bfseries 0.33 $\pm$ 0.01
				& \bfseries 0.33 $\pm$ 0.02
				& \bfseries 0.29 $\pm$ 0.01 \\
	SSLayout360
                & \bfseries 0.36 $\pm$ 0.01
				& \bfseries 0.34 $\pm$ 0.01
				& \bfseries 0.32 $\pm$ 0.02
				& \bfseries 0.31 $\pm$ 0.01
				& \bfseries 0.29 $\pm$ 0.01 \\
    \bottomrule
	\end{tabular}
	}
\end{minipage}
\begin{minipage}[t]{0.5\textwidth}
\centering
\resizebox{\columnwidth}{!}{
	\begin{tabular} {lrrrrrr}
	\toprule
	\multicolumn{2}{l}{\textbf{MatterportLayout (10$+$ Corners)}}
	& \multicolumn{3}{c}{~~3D IoU (\%) $\uparrow$} \\
	\cmidrule{2-6}
	\multicolumn{1}{l}{\multirow{2}{*}{Method}} &
	\multicolumn{1}{c}{\multirow{1}{*}{~~~~50 labels}} &
	\multicolumn{1}{c}{\multirow{1}{*}{~~~100 labels}} &
	\multicolumn{1}{c}{\multirow{1}{*}{~~~200 labels}} &
	\multicolumn{1}{c}{\multirow{1}{*}{~~~400 labels}} &
	\multicolumn{1}{c}{\multirow{1}{*}{~~1,650 labels}} \\
    & \multicolumn{1}{c}{\multirow{1}{*}{~~1,837 images}}
    & \multicolumn{1}{c}{\multirow{1}{*}{~~1,837 images}}
    & \multicolumn{1}{c}{\multirow{1}{*}{~~1,837 images}}
    & \multicolumn{1}{c}{\multirow{1}{*}{~~1,837 images}}
    & \multicolumn{1}{c}{\multirow{1}{*}{~~~~1,837 images}} \\
    \midrule
    HorizonNet
                & $57.82 \pm 1.11$
				& $60.52 \pm 0.80$
				& $61.49 \pm 1.23$
				& \bfseries 64.91 $\pm$ 1.99
				& $67.90 \pm 0.69$ \\
	SSLayout360
                & \bfseries 60.53 $\pm$ 1.71
				& \bfseries 62.33 $\pm$ 1.14
				& \bfseries 64.15 $\pm$ 0.95
				& \bfseries 66.52 $\pm$ 1.13
				& \bfseries 70.24 $\pm$ 0.98 \\
	\end{tabular}
	}
\resizebox{\columnwidth}{!}{
	\begin{tabular} {lrrrrrr}
	\toprule
	& \multicolumn{5}{c}{2D IoU (\%) $\uparrow$} \\
	\cmidrule{2-6}
	\multicolumn{1}{l}{\multirow{2}{*}{Method}} &
	\multicolumn{1}{c}{\multirow{1}{*}{~~~~50 labels}} &
	\multicolumn{1}{c}{\multirow{1}{*}{~~~100 labels}} &
	\multicolumn{1}{c}{\multirow{1}{*}{~~~200 labels}} &
	\multicolumn{1}{c}{\multirow{1}{*}{~~~400 labels}} &
	\multicolumn{1}{c}{\multirow{1}{*}{~~1,650 labels}} \\
    & \multicolumn{1}{c}{\multirow{1}{*}{~~1,837 images}}
    & \multicolumn{1}{c}{\multirow{1}{*}{~~1,837 images}}
    & \multicolumn{1}{c}{\multirow{1}{*}{~~1,837 images}}
    & \multicolumn{1}{c}{\multirow{1}{*}{~~1,837 images}}
    & \multicolumn{1}{c}{\multirow{1}{*}{~~~~1,837 images}} \\
    \midrule
    HorizonNet
                & $59.85 \pm 1.22$
				& $62.56 \pm 0.79$
				& $63.59 \pm 1.15$
				& \bfseries 67.19 $\pm$ 2.13
				& $69.75 \pm 0.61$ \\
	SSLayout360
                & \bfseries 62.49 $\pm$ 1.75
				& \bfseries 64.28 $\pm$ 1.17
				& \bfseries 66.13 $\pm$ 0.89
				& \bfseries 68.42 $\pm$ 1.12
				& \bfseries 71.98 $\pm$ 1.02 \\
	\end{tabular}
	}
\resizebox{\columnwidth}{!}{
	\begin{tabular} {lrrrrrr}
	\toprule
	& \multicolumn{5}{c}{$\delta_1$ $\uparrow$} \\
	\cmidrule{2-6}
	\multicolumn{1}{l}{\multirow{2}{*}{Method}} &
	\multicolumn{1}{c}{\multirow{1}{*}{~~~~50 labels}} &
	\multicolumn{1}{c}{\multirow{1}{*}{~~~100 labels}} &
	\multicolumn{1}{c}{\multirow{1}{*}{~~~200 labels}} &
	\multicolumn{1}{c}{\multirow{1}{*}{~~~400 labels}} &
	\multicolumn{1}{c}{\multirow{1}{*}{~~1,650 labels}} \\
    & \multicolumn{1}{c}{\multirow{1}{*}{~~1,837 images}}
    & \multicolumn{1}{c}{\multirow{1}{*}{~~1,837 images}}
    & \multicolumn{1}{c}{\multirow{1}{*}{~~1,837 images}}
    & \multicolumn{1}{c}{\multirow{1}{*}{~~1,837 images}}
    & \multicolumn{1}{c}{\multirow{1}{*}{~~~~1,837 images}} \\
    \midrule
    HorizonNet
                & $0.80 \pm 0.01$
				& $0.84 \pm 0.01$
				& $0.85 \pm 0.01$
				& $0.86 \pm 0.01$
				& $0.88 \pm 0.01$ \\
	SSLayout360
                & \bfseries 0.82 $\pm$ 0.01
				& \bfseries 0.87 $\pm$ 0.01
				& \bfseries 0.87 $\pm$ 0.01
				& \bfseries 0.89 $\pm$ 0.01
				& \bfseries 0.90 $\pm$ 0.01 \\
	\end{tabular}
	}
\resizebox{\columnwidth}{!}{
	\begin{tabular} {lrrrrrr}
	\toprule
	& \multicolumn{5}{c}{RMSE $\downarrow$} \\
	\cmidrule{2-6}
	\multicolumn{1}{l}{\multirow{2}{*}{Method}} &
	\multicolumn{1}{c}{\multirow{1}{*}{~~~~50 labels}} &
	\multicolumn{1}{c}{\multirow{1}{*}{~~~100 labels}} &
	\multicolumn{1}{c}{\multirow{1}{*}{~~~200 labels}} &
	\multicolumn{1}{c}{\multirow{1}{*}{~~~400 labels}} &
	\multicolumn{1}{c}{\multirow{1}{*}{~~1,650 labels}} \\
    & \multicolumn{1}{c}{\multirow{1}{*}{~~1,837 images}}
    & \multicolumn{1}{c}{\multirow{1}{*}{~~1,837 images}}
    & \multicolumn{1}{c}{\multirow{1}{*}{~~1,837 images}}
    & \multicolumn{1}{c}{\multirow{1}{*}{~~1,837 images}}
    & \multicolumn{1}{c}{\multirow{1}{*}{~~~~1,837 images}} \\
    \midrule
    HorizonNet
                & $0.56 \pm 0.02$
				& $0.53 \pm 0.02$
				& \bfseries 0.50 $\pm$ 0.03
				& \bfseries 0.47 $\pm$ 0.03
				& \bfseries 0.44 $\pm$ 0.03 \\
	SSLayout360
                & \bfseries 0.52 $\pm$ 0.03
				& \bfseries 0.49 $\pm$ 0.01
				& \bfseries 0.49 $\pm$ 0.02
				& \bfseries 0.46 $\pm$ 0.03
				& \bfseries 0.41 $\pm$ 0.02 \\
    \bottomrule
	\end{tabular}
	} \smallskip
\end{minipage}
{\caption{Quantitative layout results for \textbf{8 corners (left)} and \textbf{10+ corners (right)} evaluated on the MatterportLayout test set averaged over four independent runs with different random seeds. Number format: mean value $\pm$ standard deviation.}
\label{8-10+-corners}}
\end{table*}

\renewcommand{\thefigure}{E1}
\begin{figure*}[t]
    \begin{minipage}[t]{0.4\textwidth}
        \subfloat{%
            \includegraphics[width=\columnwidth]{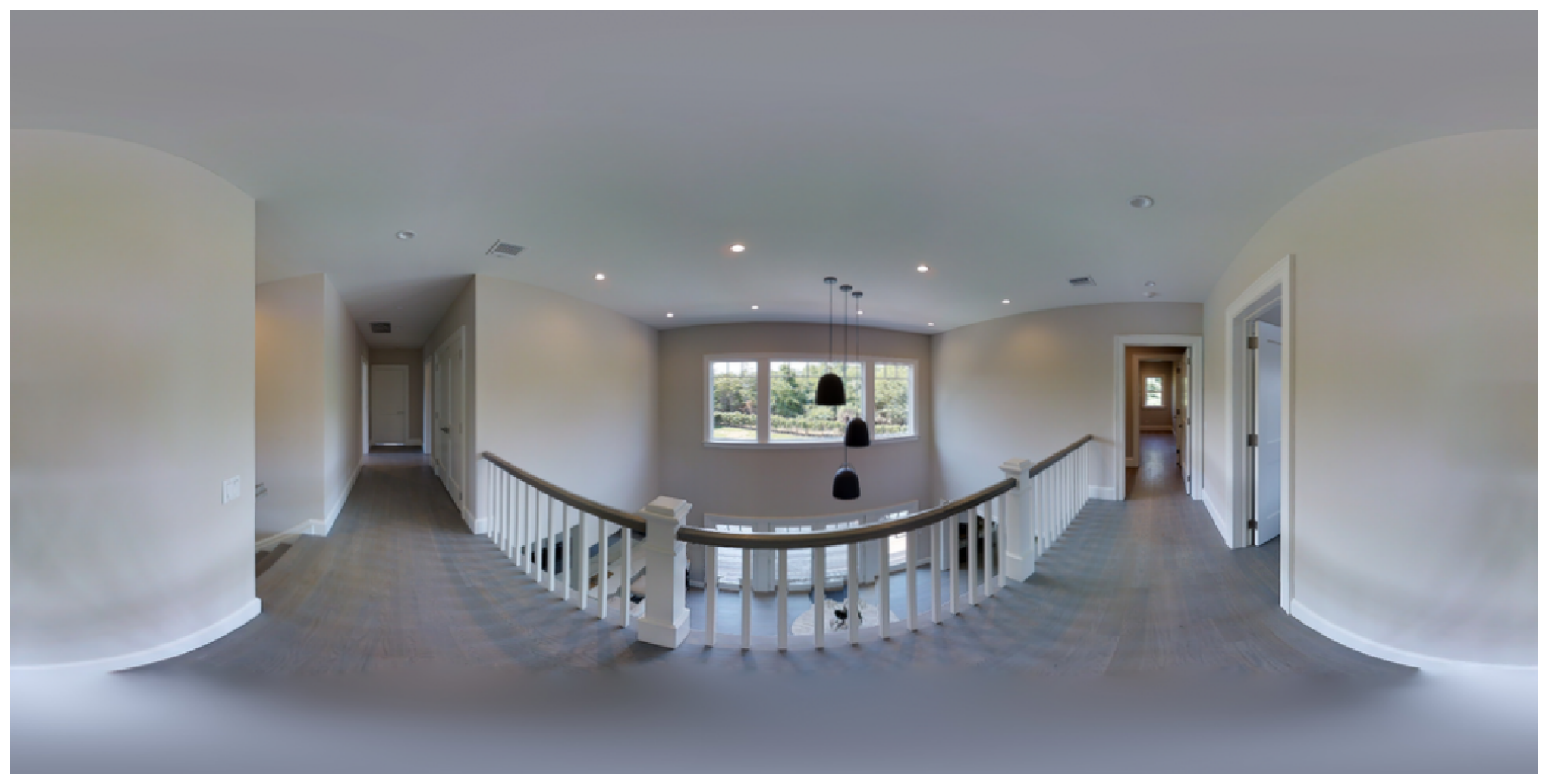}
        }
        \\[-3ex]
        \subfloat{%
            \includegraphics[width=\columnwidth]{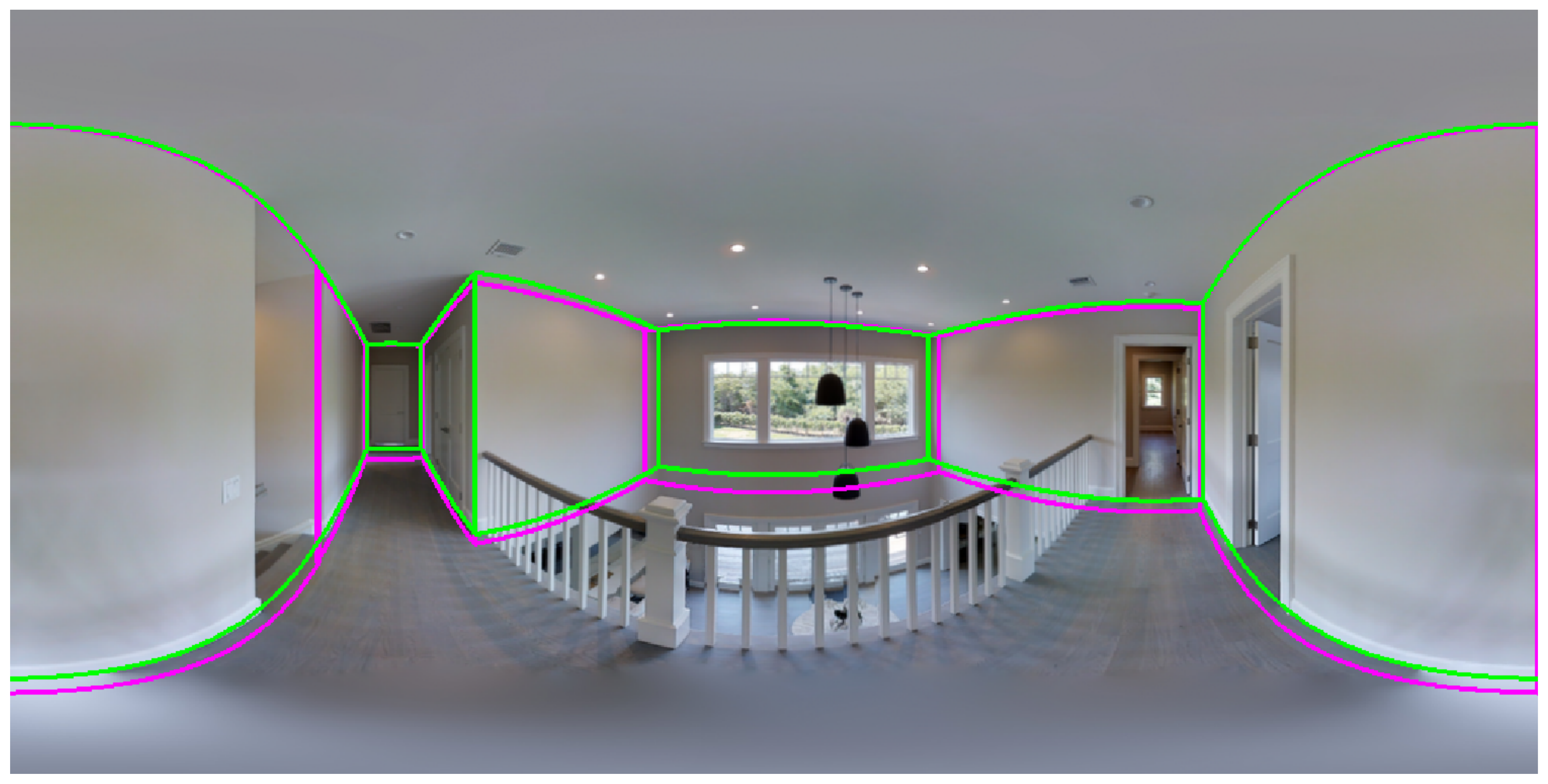}
        }
    \end{minipage}%
    \hfill
    \begin{minipage}[t]{0.5\textwidth}
        \subfloat{%
            \includegraphics[height=0.75\textwidth]{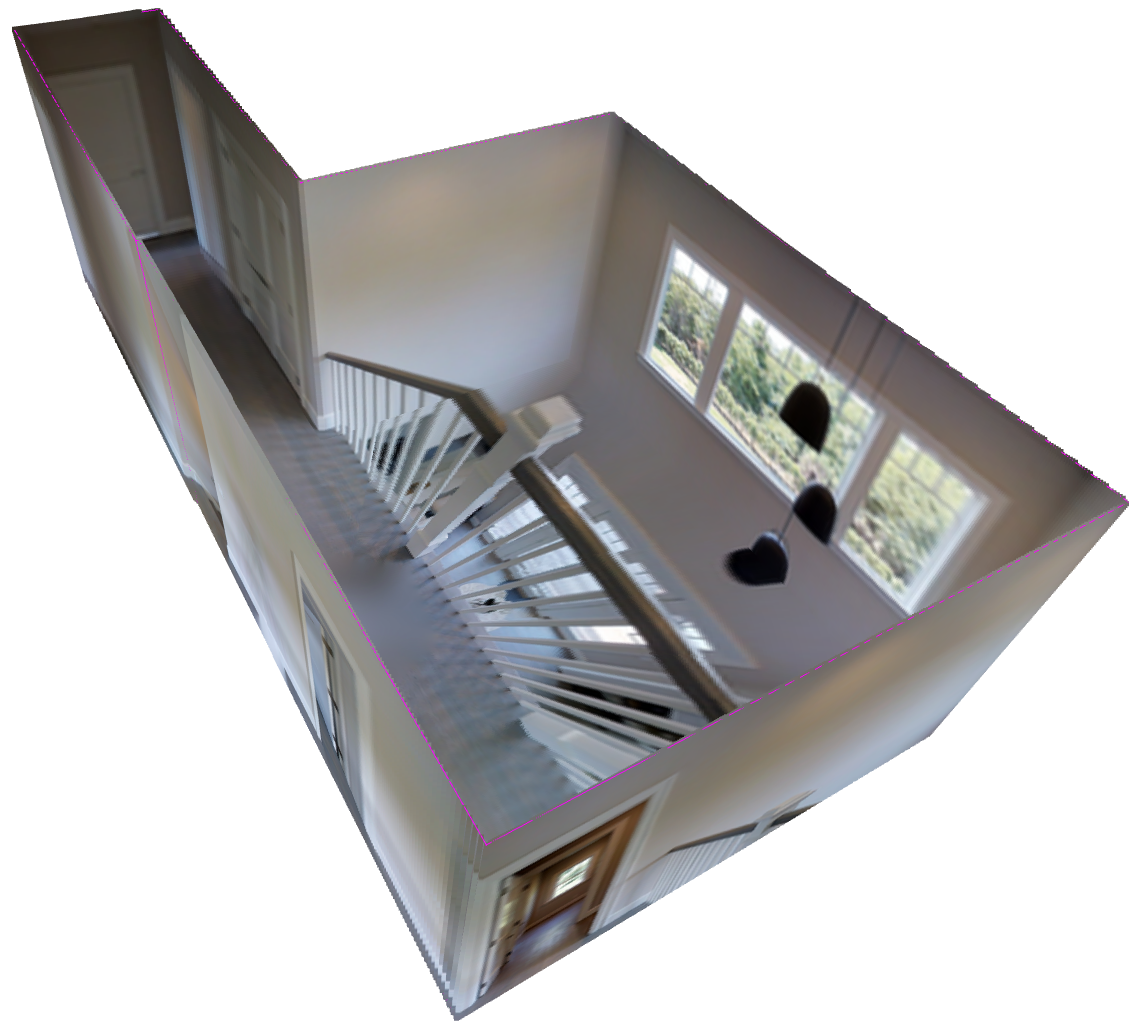}
        }
    \end{minipage}%
    \\[1ex]
    \begin{minipage}[t]{0.4\textwidth}
        \subfloat{%
            \includegraphics[width=\columnwidth]{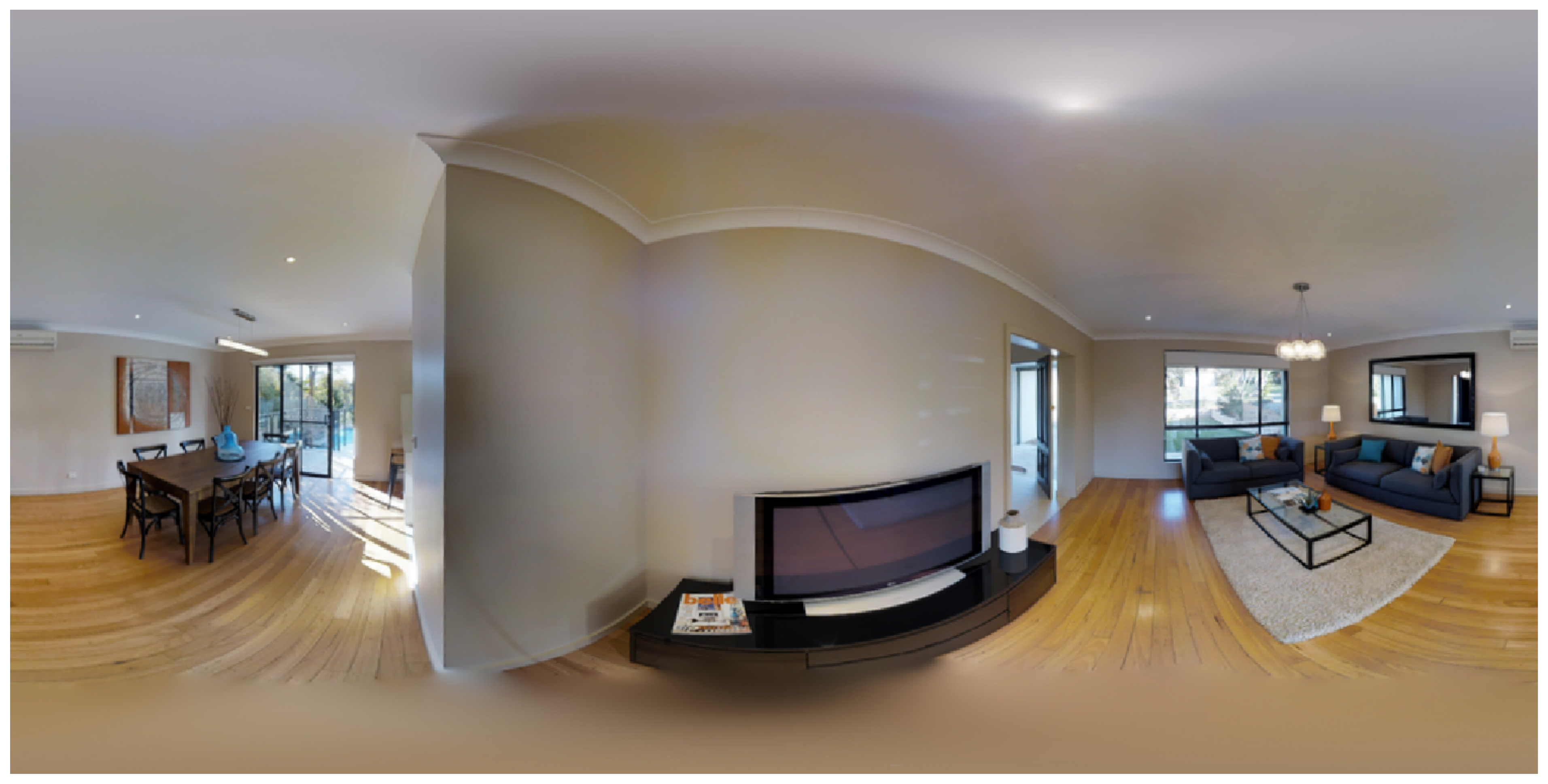}
        }
        \\[-3ex]
        \subfloat{%
            \includegraphics[width=\columnwidth]{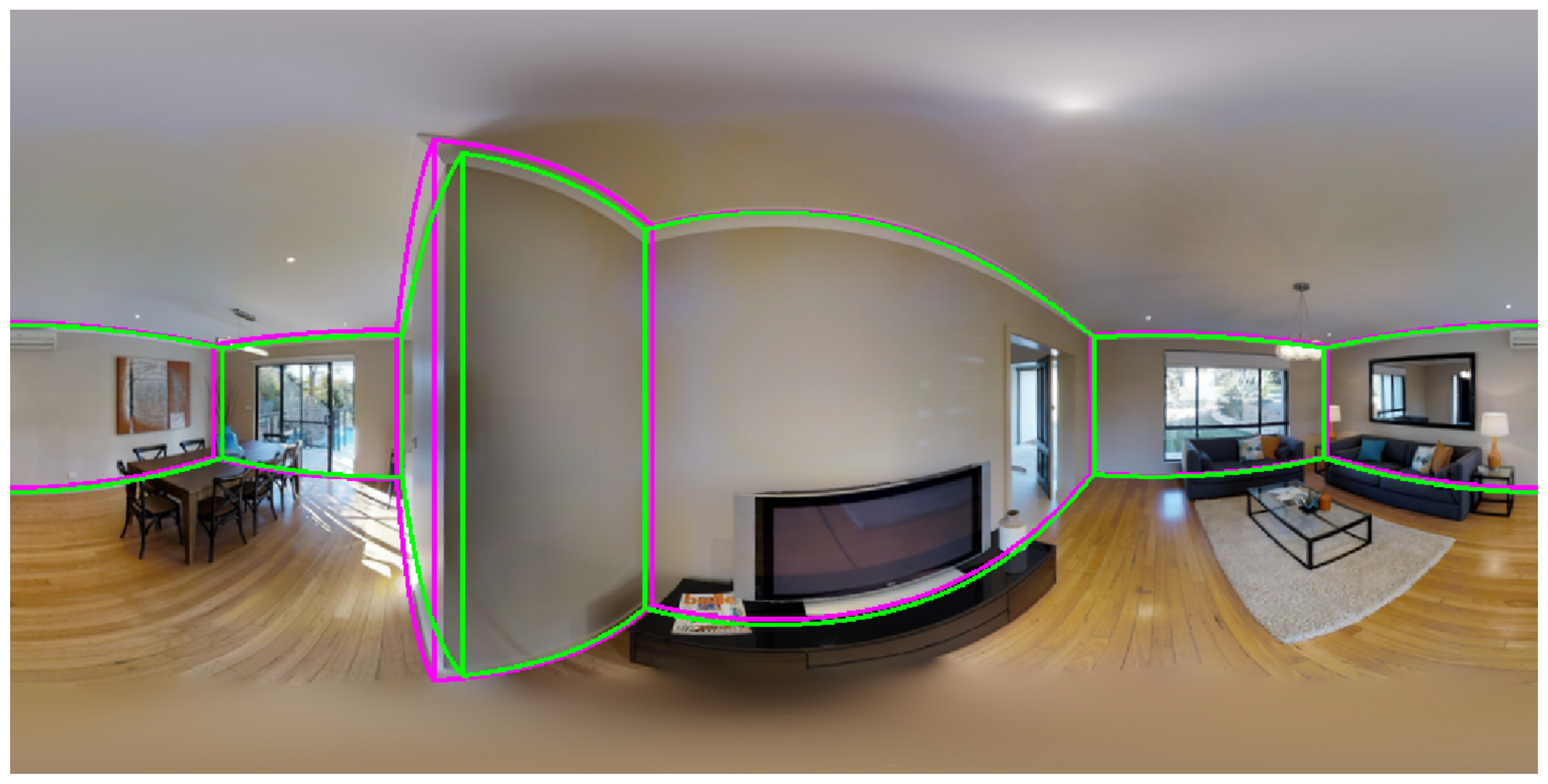}
        }
    \end{minipage}%
    \hfill
    \begin{minipage}[t]{0.5\textwidth}
        \subfloat{%
            \includegraphics[height=0.75\textwidth]{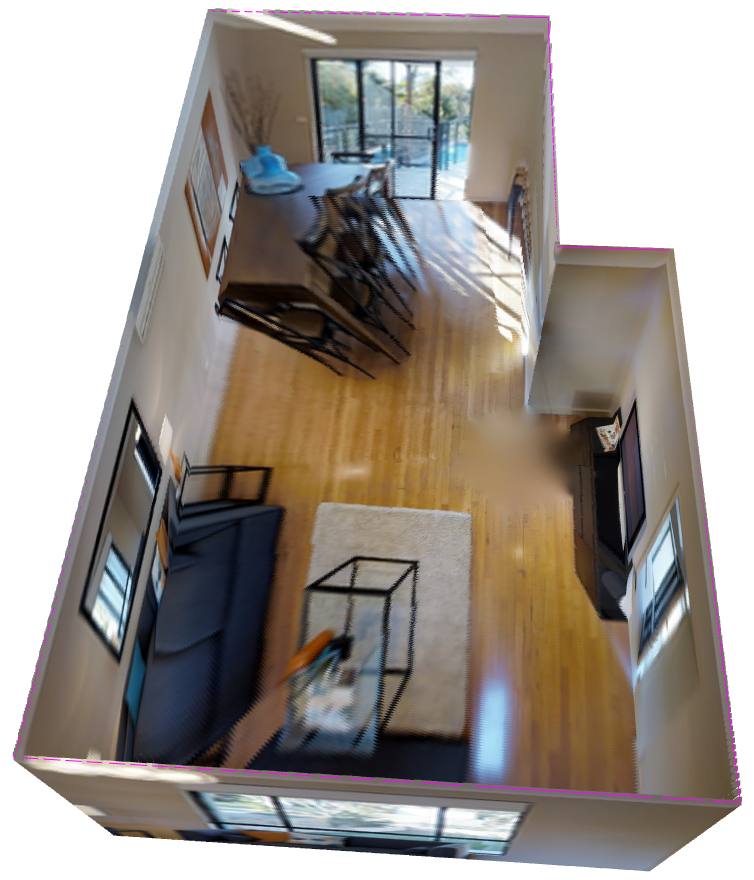}
        }
    \end{minipage}%
    \\[1ex]
    \begin{minipage}[t]{0.4\textwidth}
        \subfloat{%
            \includegraphics[width=\columnwidth]{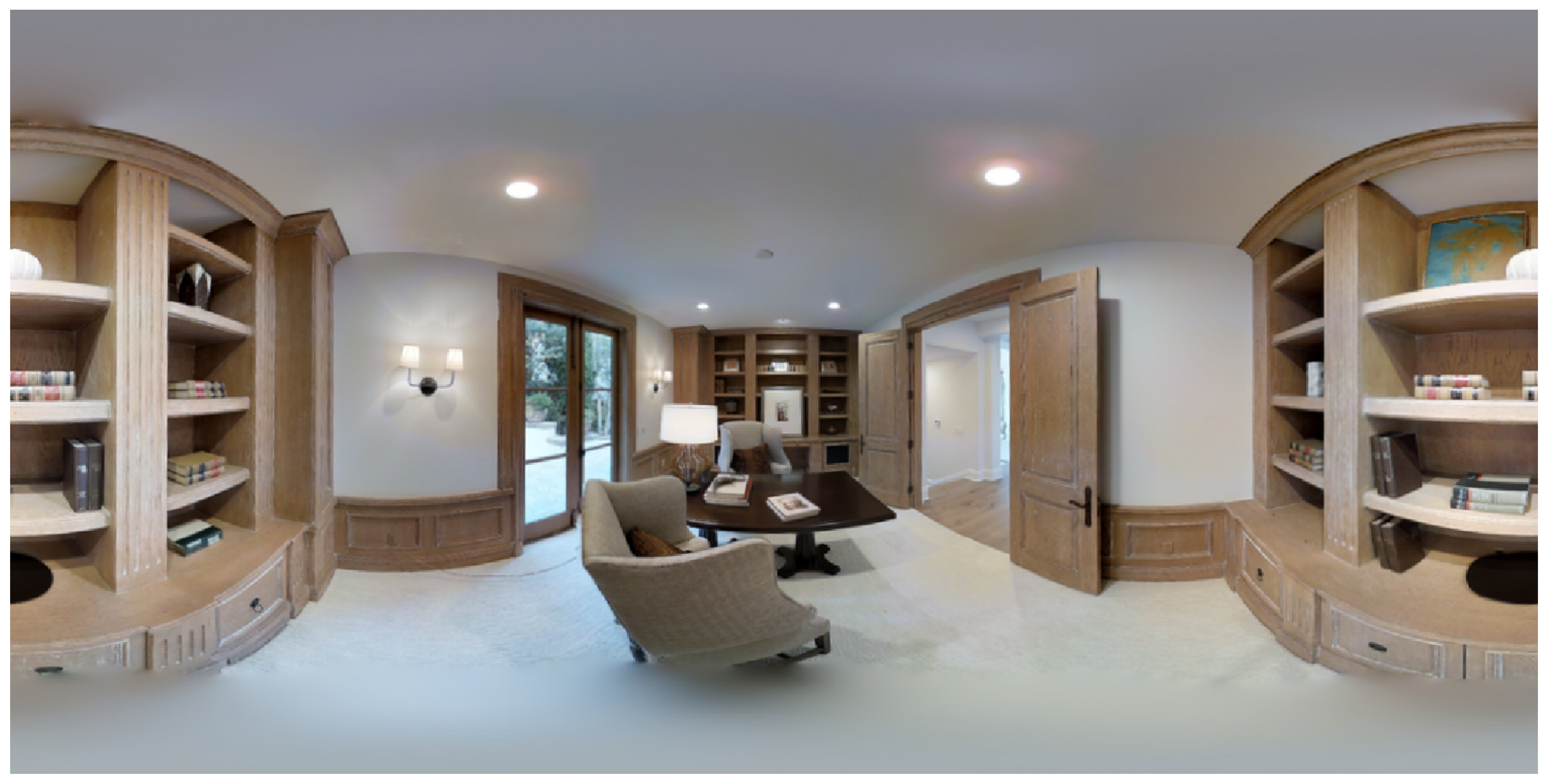}
        }
        \\[-3ex]
        \subfloat{%
            \includegraphics[width=\columnwidth]{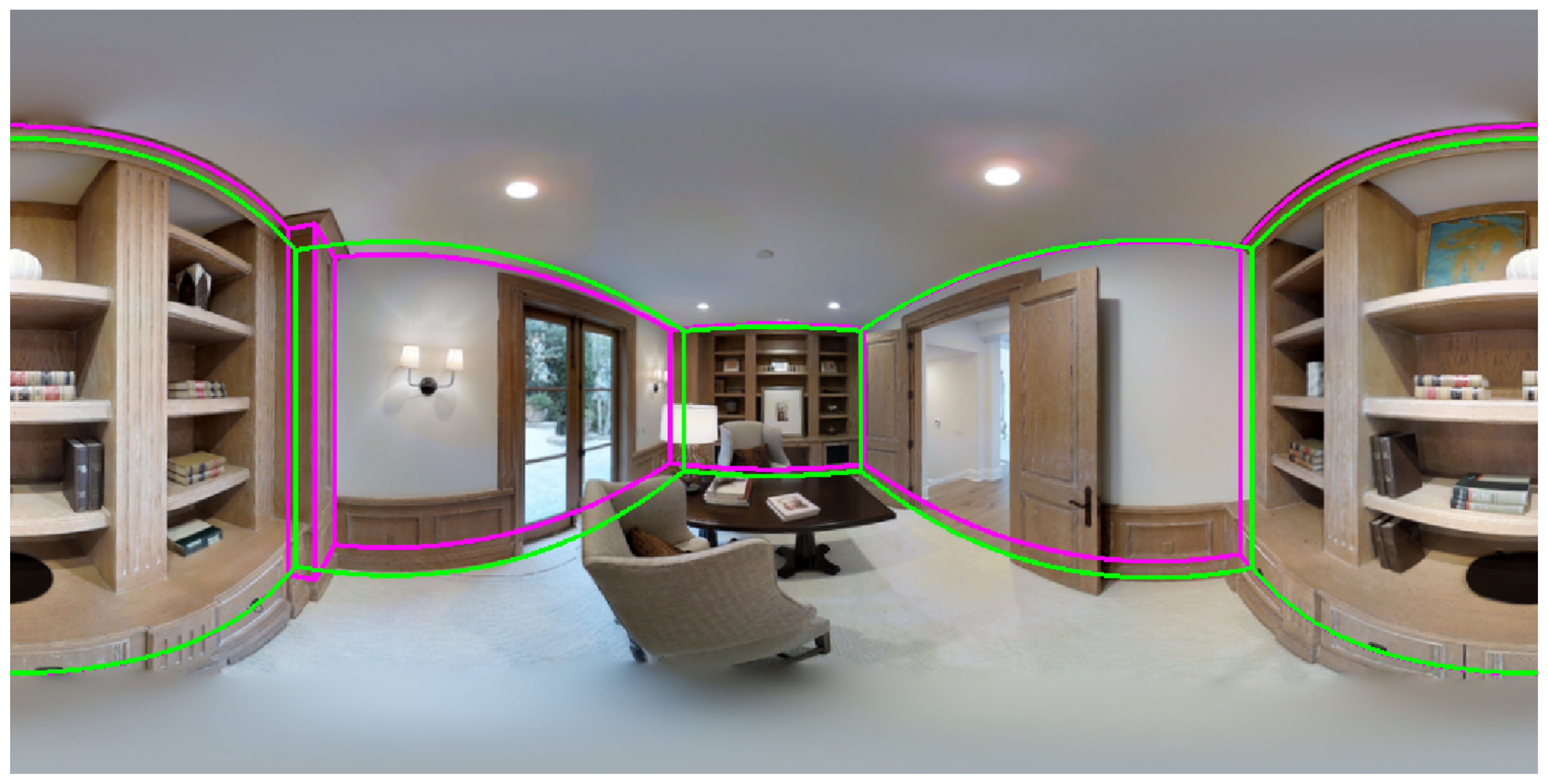}
        }
    \end{minipage}%
    \hfill
    \begin{minipage}[t]{0.5\textwidth}
        \subfloat{%
            \includegraphics[height=0.75\textwidth]{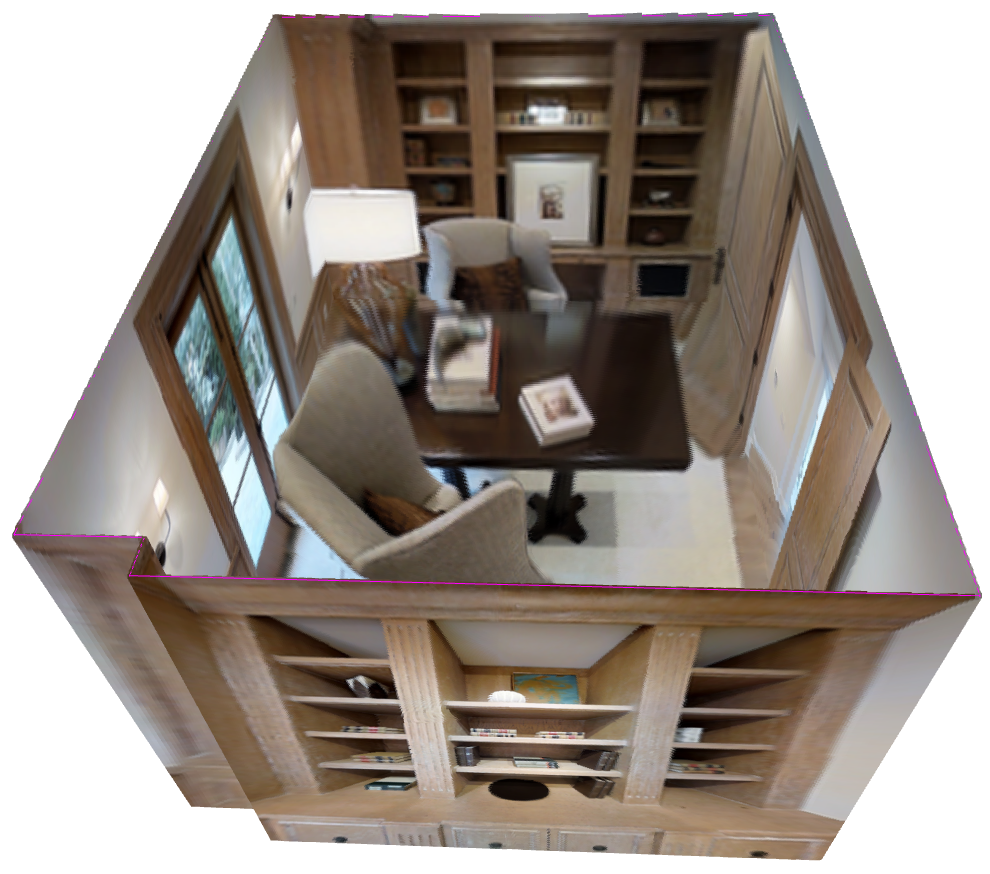}
        }
    \end{minipage}%
    \caption{Qualitative layout estimation of exemplar MatterportLayout test instances. \textbf{Left:} A comparison of ground truth layout (green lines) with our SSLayout360 model (magenta lines) trained on 100 labeled and 4,000 unlabeled images under equirectangular view. \textbf{Right:} 3D layout reconstruction. Best viewed electronically.}
    \label{qualitative01}
\end{figure*}

\renewcommand{\thefigure}{E2}
\begin{figure*}[t]
    \begin{minipage}[t]{0.4\textwidth}
        \subfloat{%
            \includegraphics[width=\columnwidth]{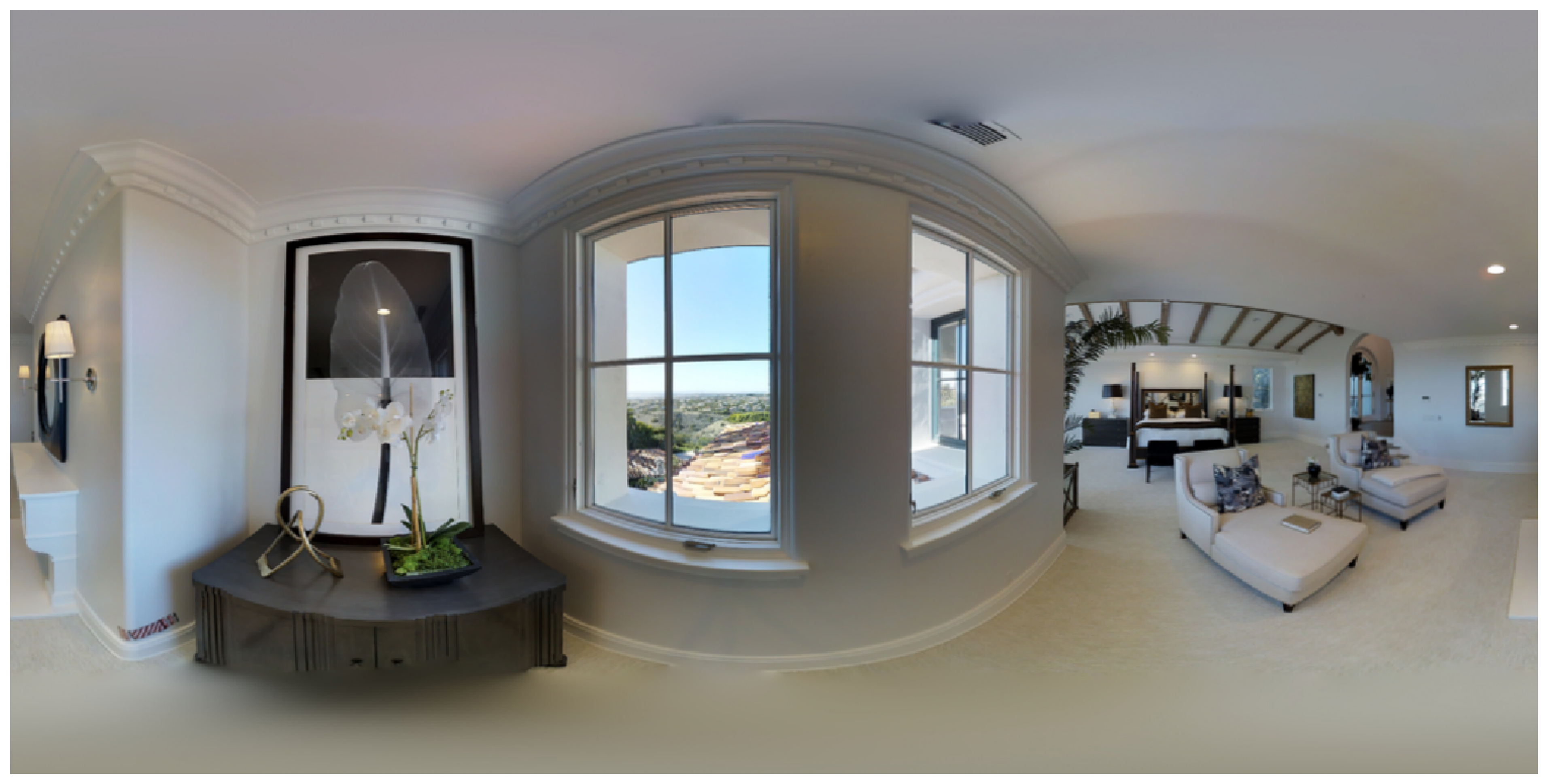}
        }
        \\[-3ex]
        \subfloat{%
            \includegraphics[width=\columnwidth]{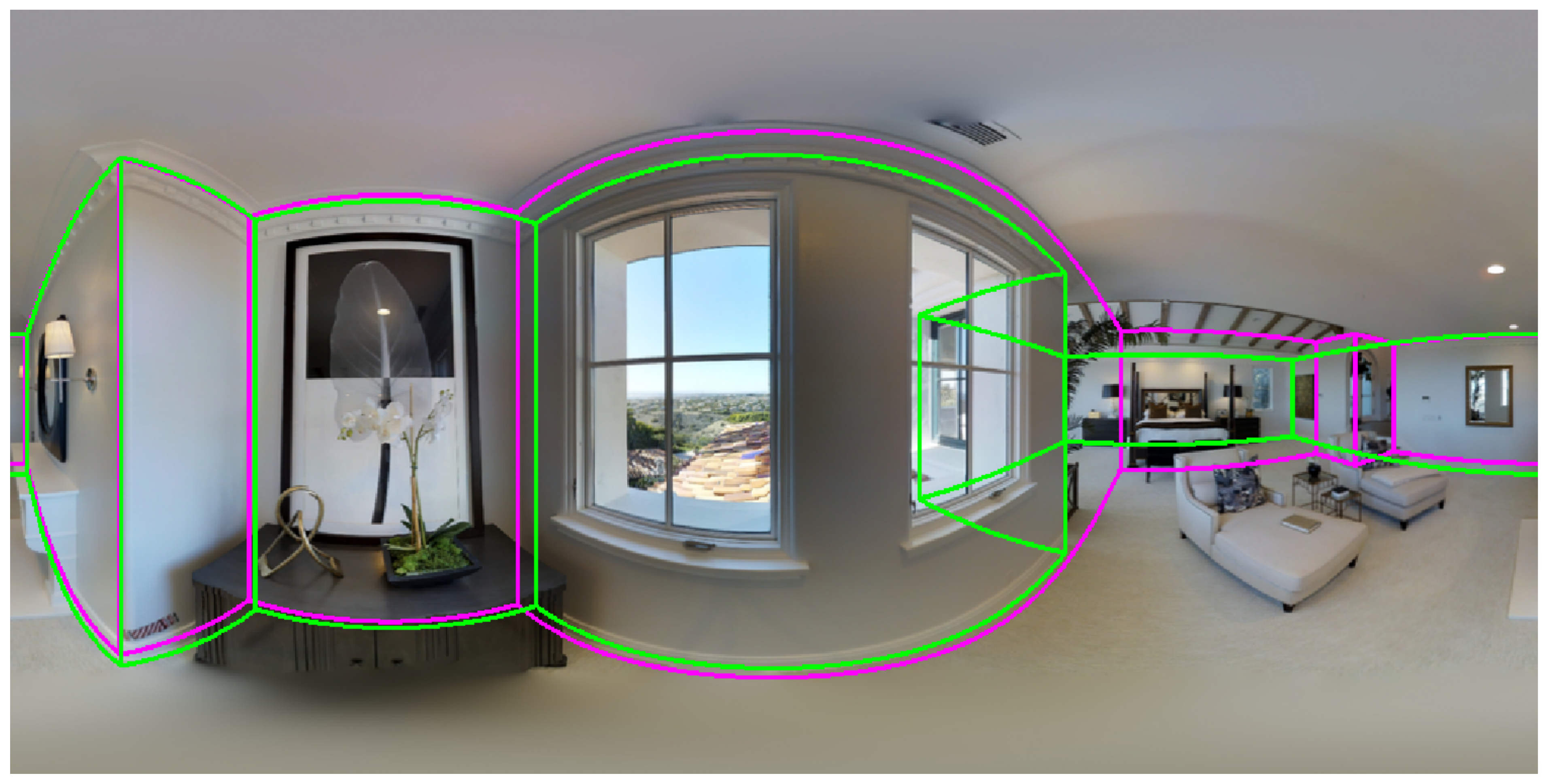}
        }
    \end{minipage}%
    \hfill
    \begin{minipage}[t]{0.5\textwidth}
        \subfloat{%
            \includegraphics[height=0.75\textwidth]{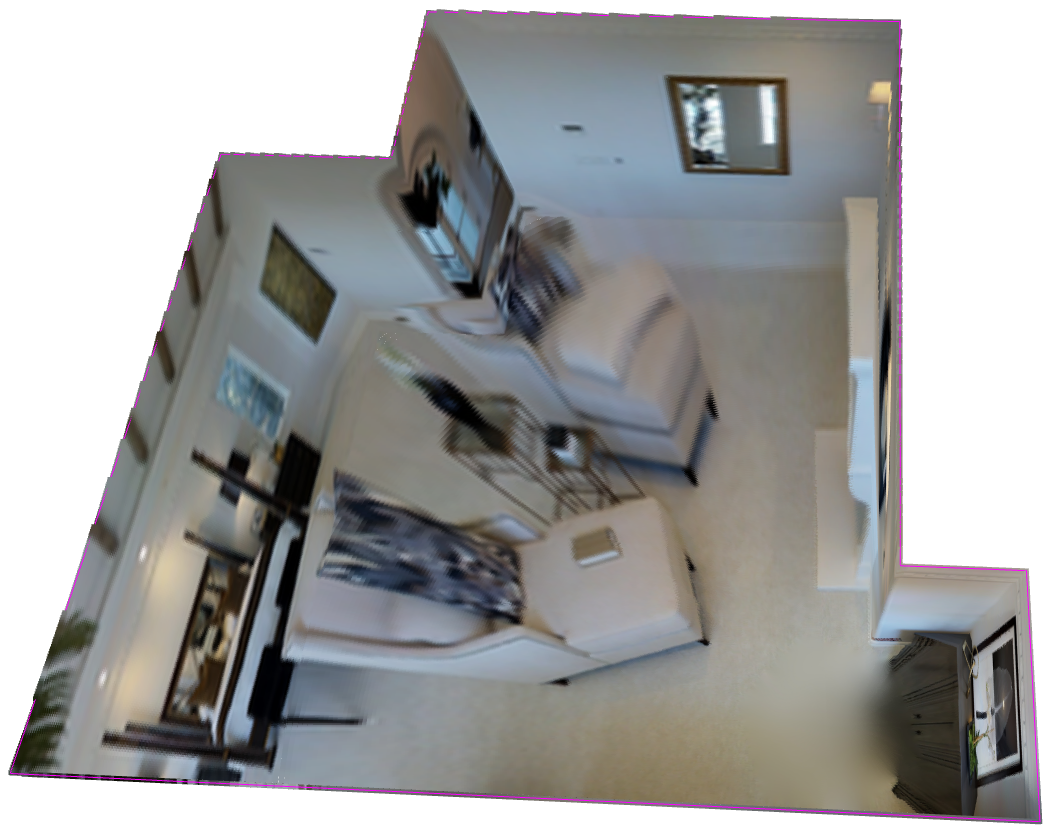}
        }
    \end{minipage}%
    \\[1ex]
    \begin{minipage}[t]{0.4\textwidth}
        \subfloat{%
            \includegraphics[width=\columnwidth]{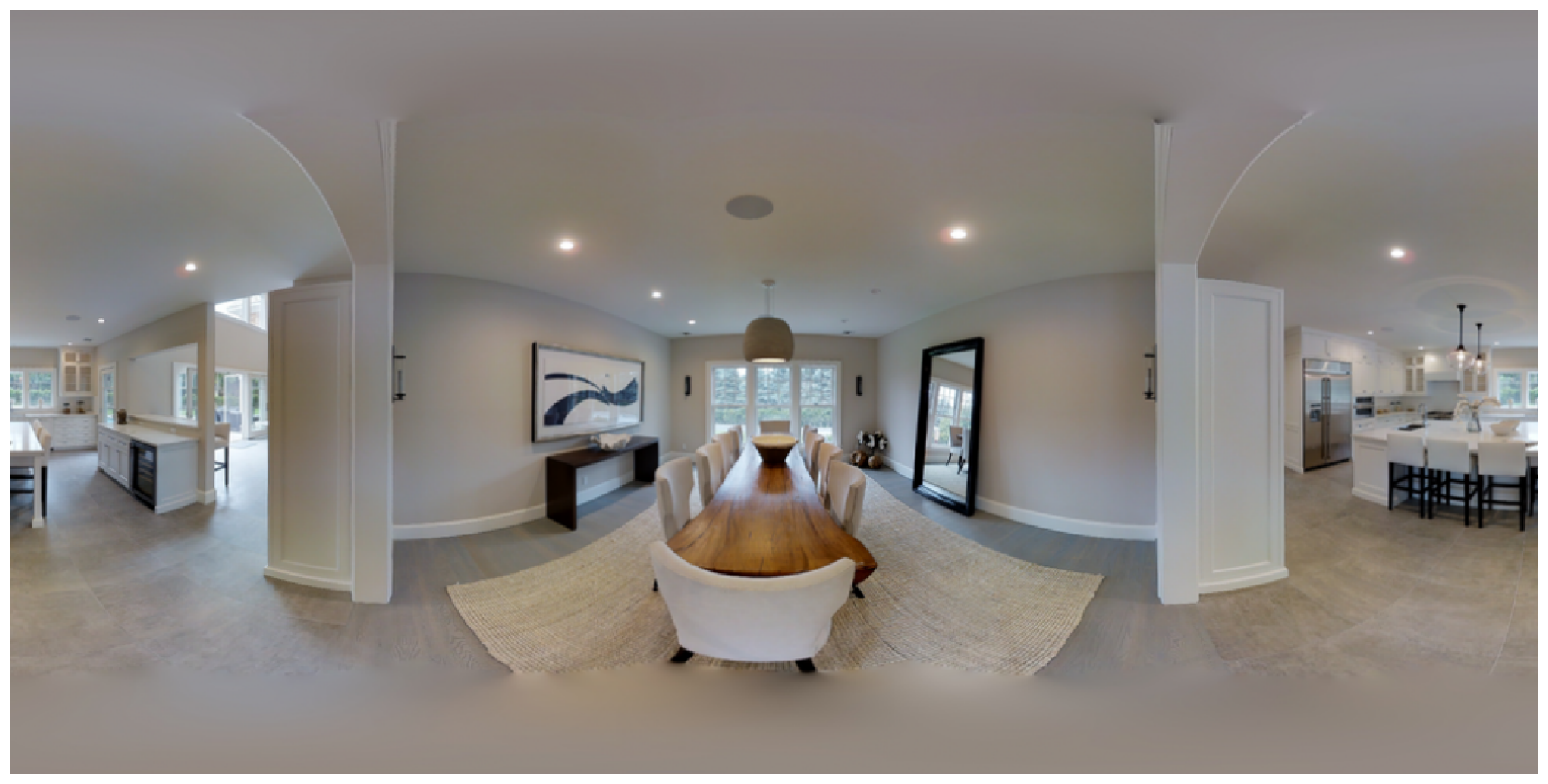}
        }
        \\[-3ex]
        \subfloat{%
            \includegraphics[width=\columnwidth]{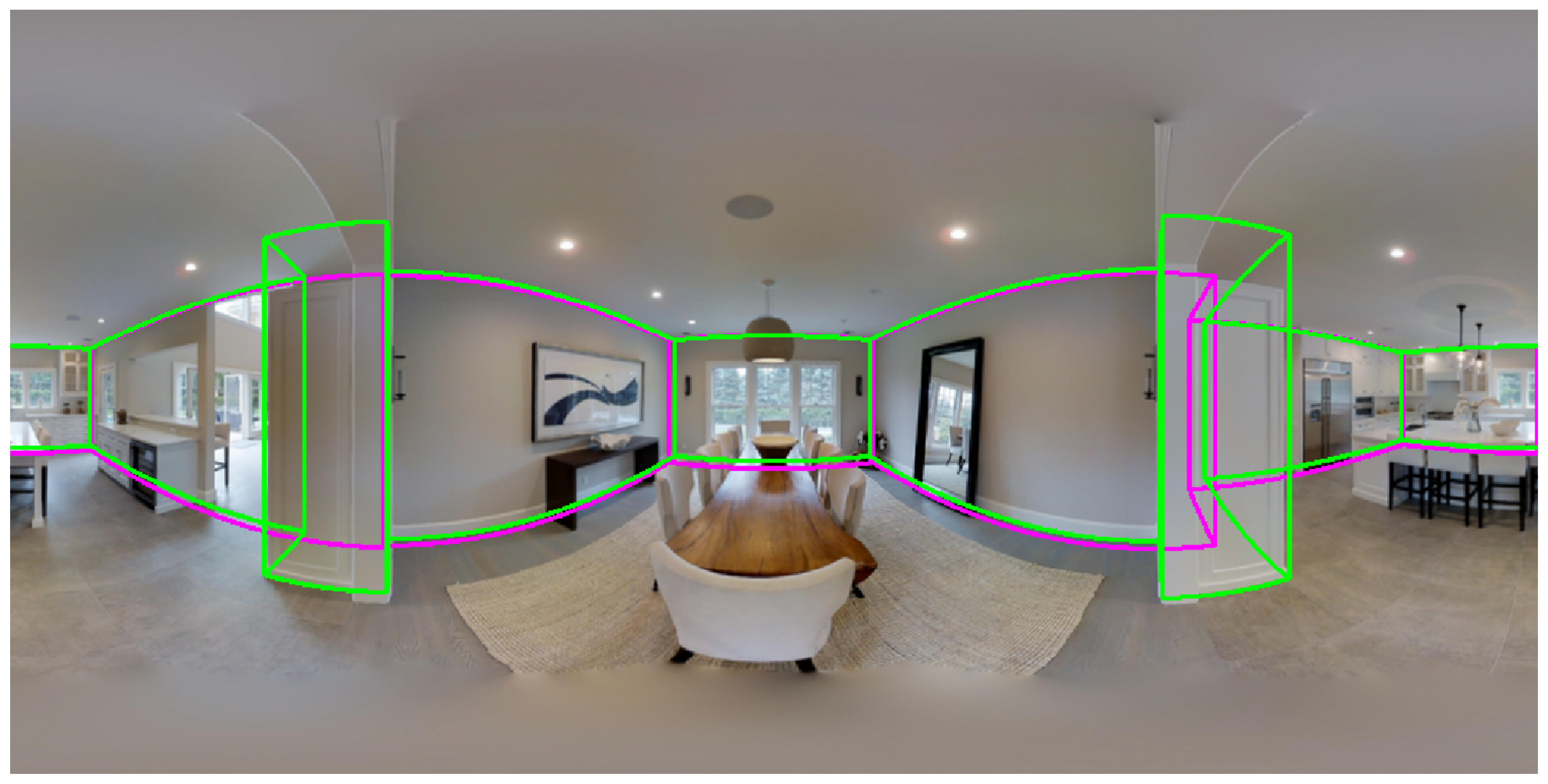}
        }
    \end{minipage}%
    \hfill
    \begin{minipage}[t]{0.5\textwidth}
        \subfloat{%
            \includegraphics[height=0.75\textwidth]{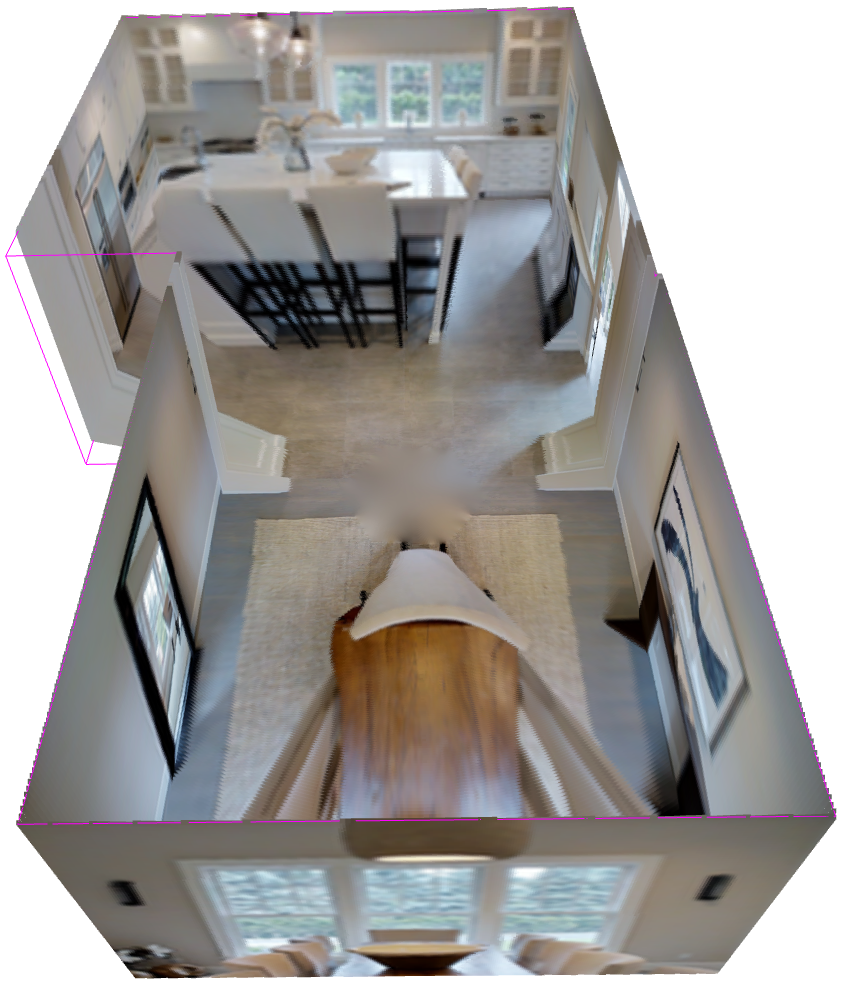}
        }
    \end{minipage}%
    \\[1ex]
    \begin{minipage}[t]{0.4\textwidth}
        \subfloat{%
            \includegraphics[width=\columnwidth]{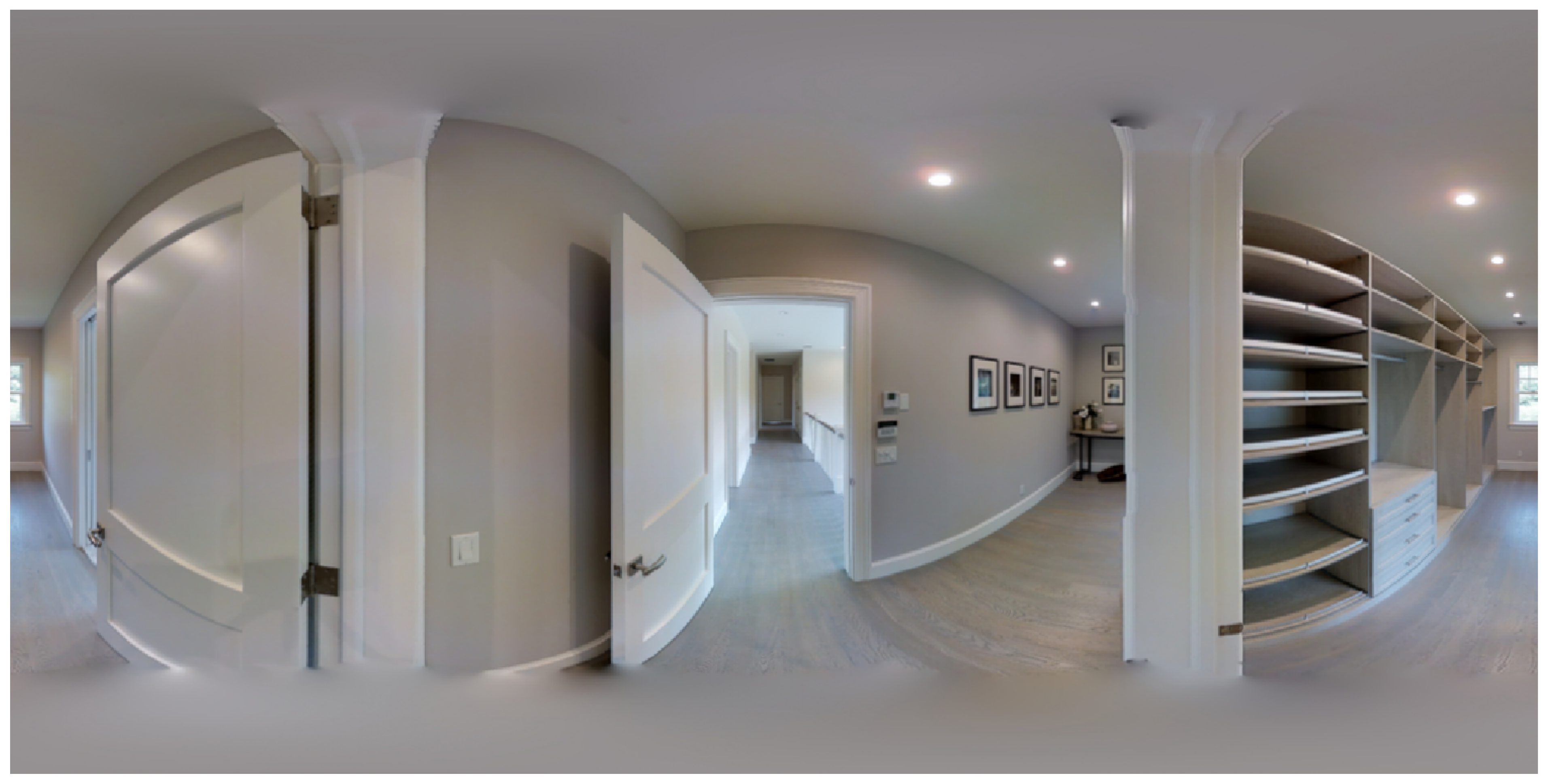}
        }
        \\[-3ex]
        \subfloat{%
            \includegraphics[width=\columnwidth]{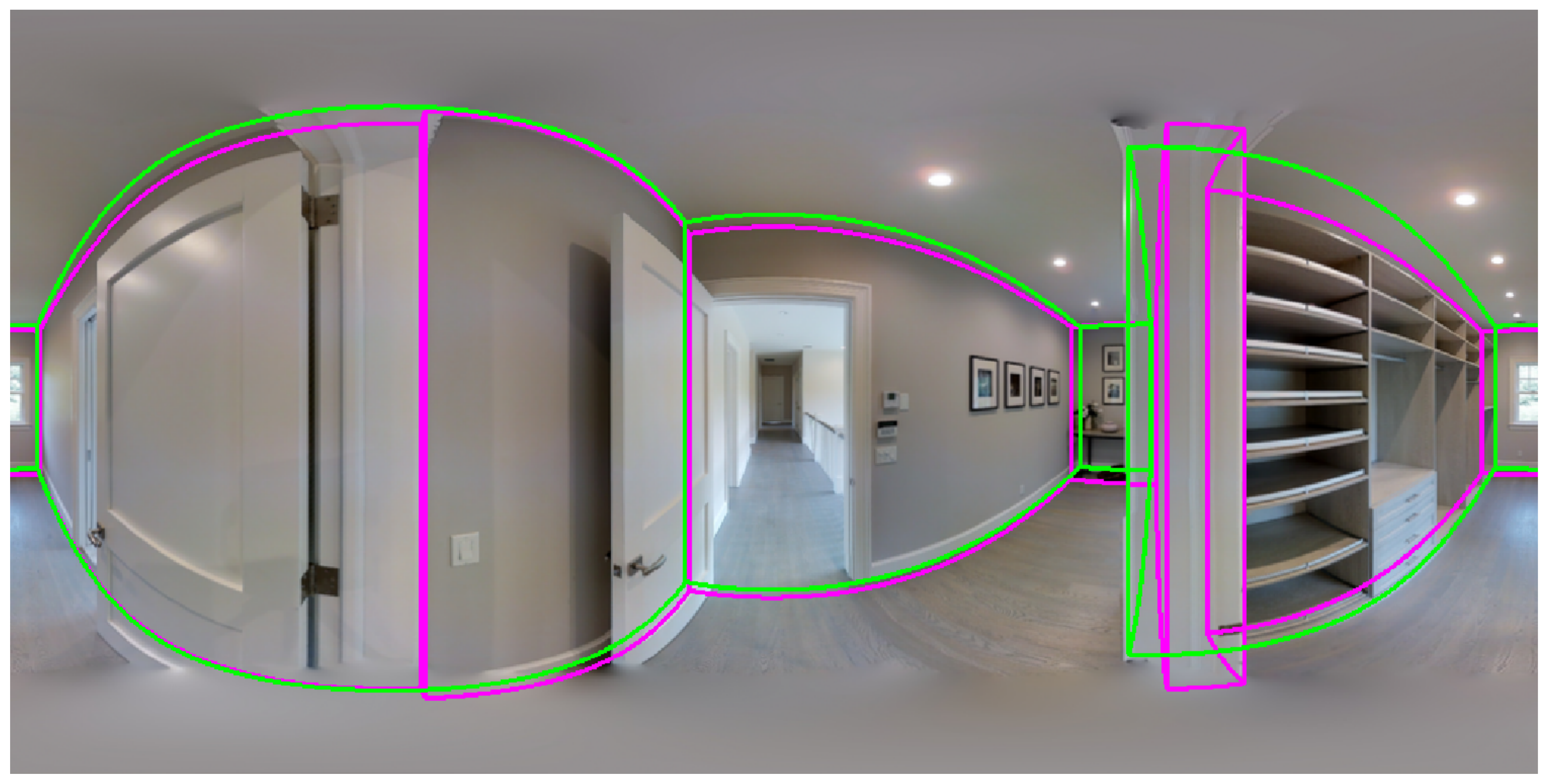}
        }
    \end{minipage}%
    \hfill
    \begin{minipage}[t]{0.5\textwidth}
        \subfloat{%
            \includegraphics[height=0.75\textwidth]{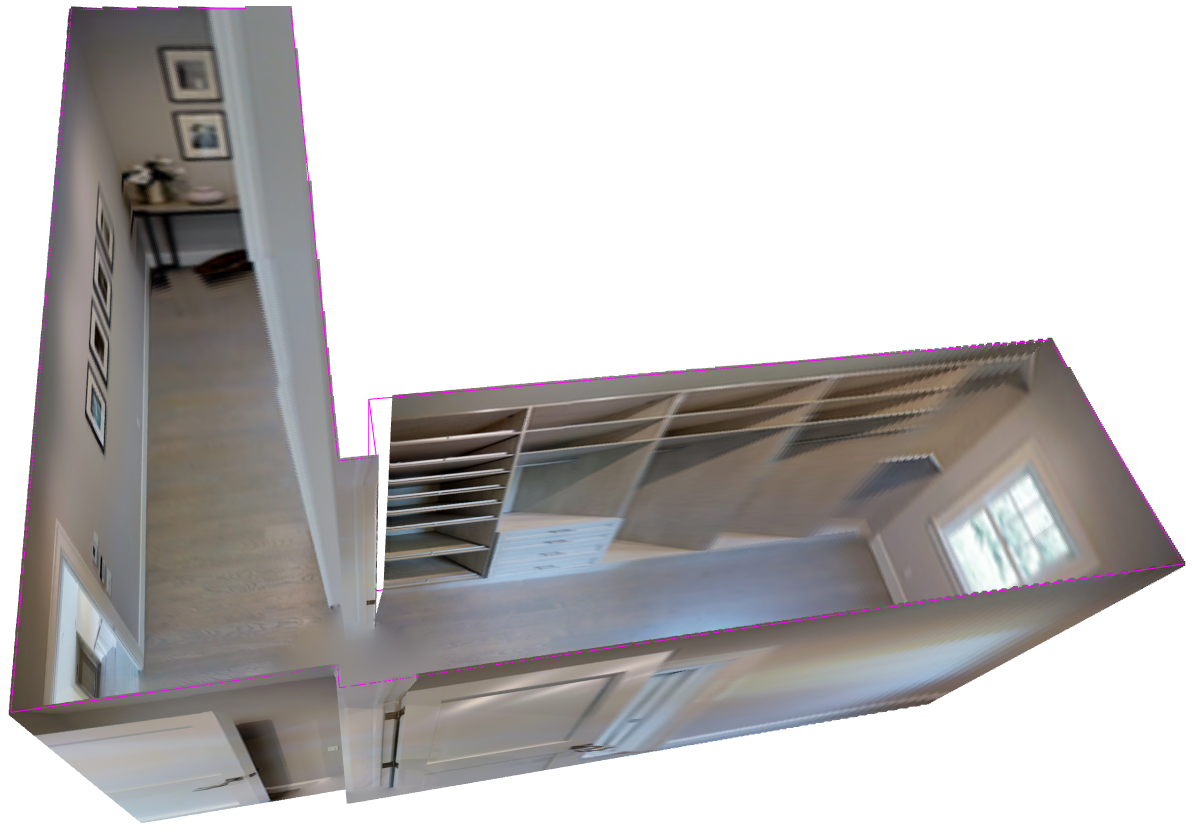}
        }
    \end{minipage}%
    \caption{Qualitative layout estimation of exemplar MatterportLayout test instances. \textbf{Left:} A comparison of ground truth layout (green lines) with our SSLayout360 model (magenta lines) trained on 100 labeled and 4,000 unlabeled images under equirectangular view. \textbf{Right:} 3D layout reconstruction. The transparent regions denote walls hidden from the camera field-of-view. Best viewed electronically.}
    \label{qualitative02}
\end{figure*}

\renewcommand{\thefigure}{E3}
\begin{figure*}[t]
    \begin{minipage}[t]{0.4\textwidth}
        \subfloat{%
            \includegraphics[width=\columnwidth]{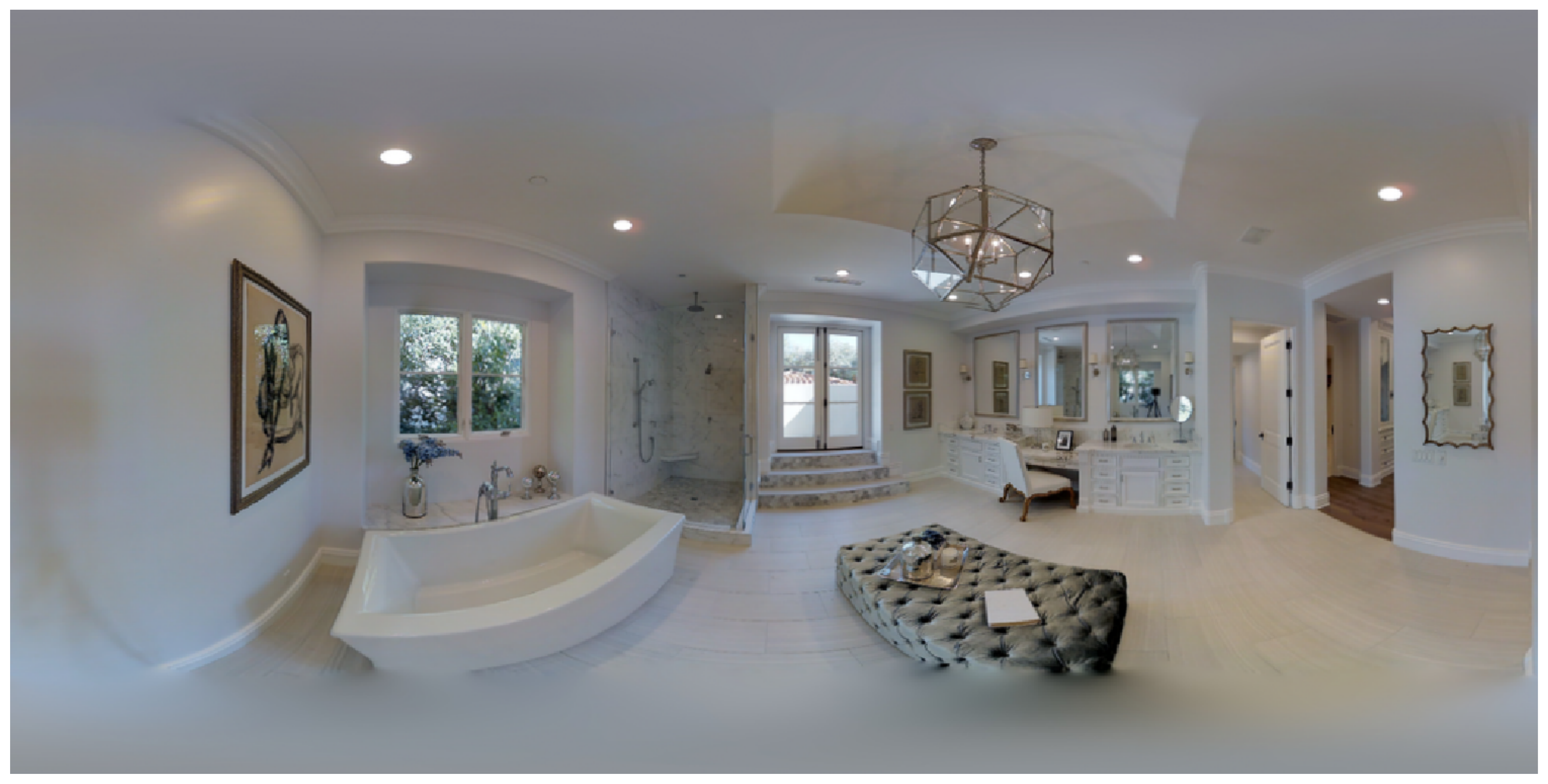}
        }
        \\[-3ex]
        \subfloat{%
            \includegraphics[width=\columnwidth]{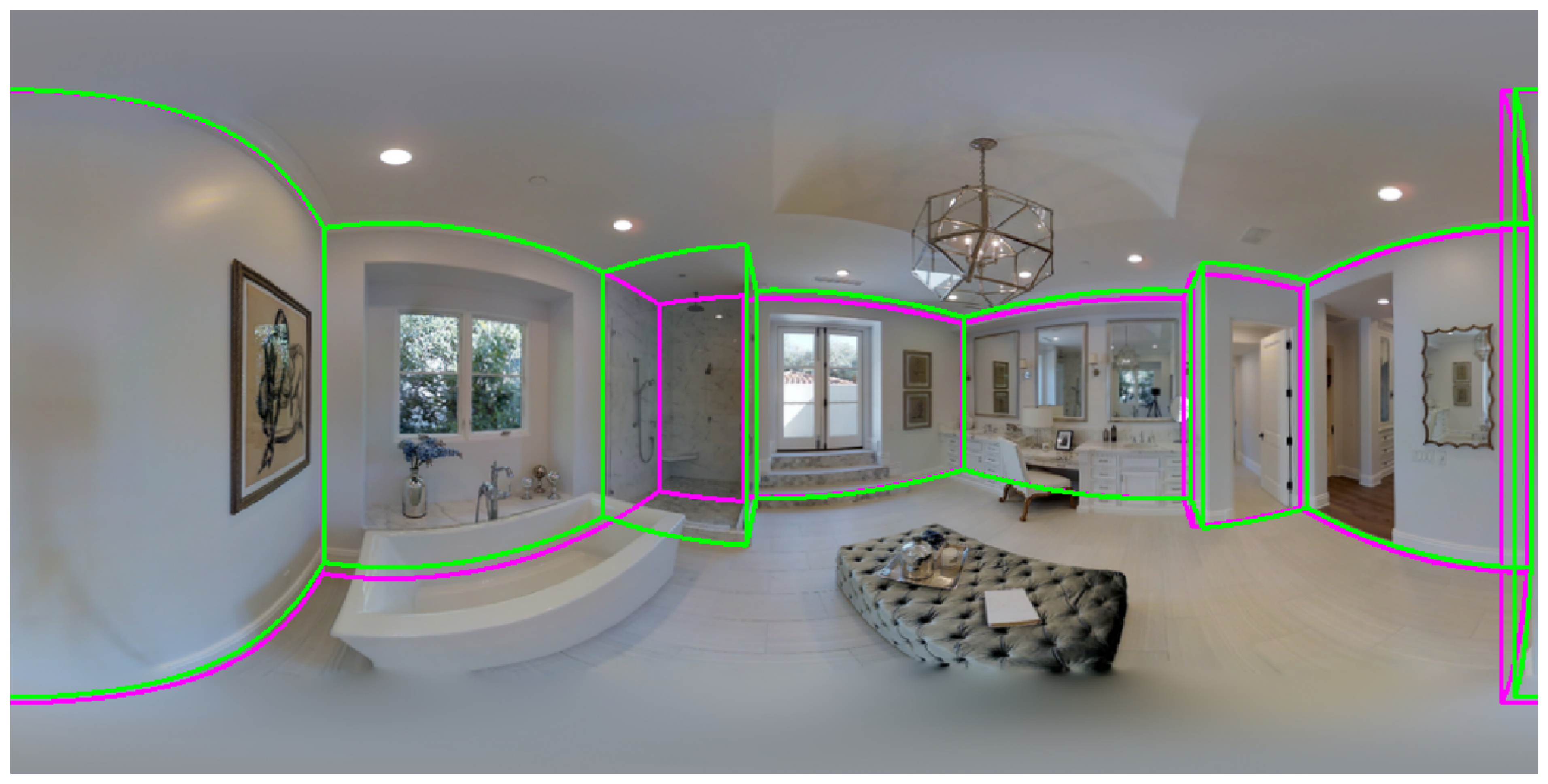}
        }
    \end{minipage}%
    \hfill
    \begin{minipage}[t]{0.5\textwidth}
        \subfloat{%
            \includegraphics[height=0.75\textwidth]{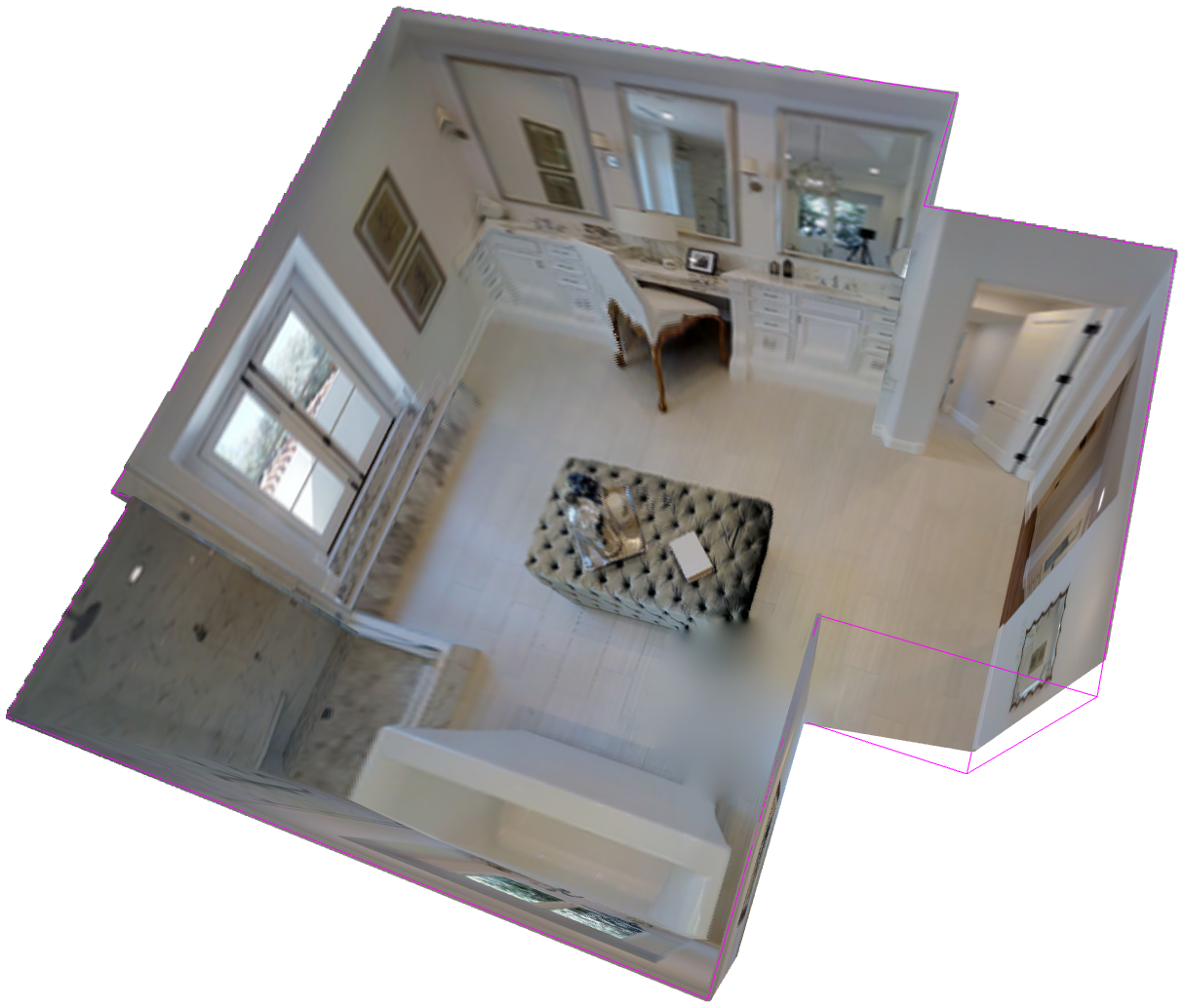}
        }
    \end{minipage}%
    \\[1ex]
    \begin{minipage}[t]{0.4\textwidth}
        \subfloat{%
            \includegraphics[width=\columnwidth]{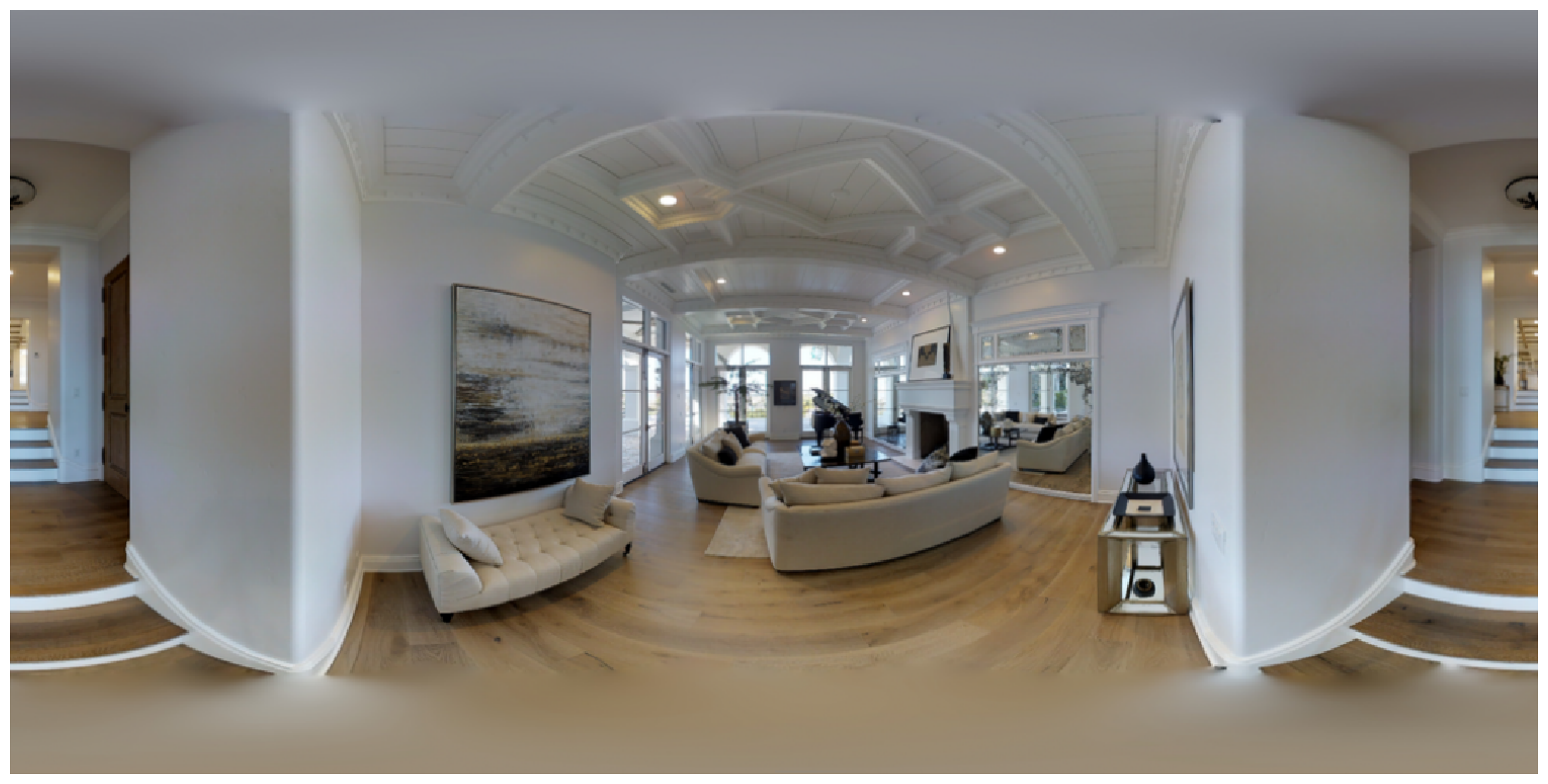}
        }
        \\[-3ex]
        \subfloat{%
            \includegraphics[width=\columnwidth]{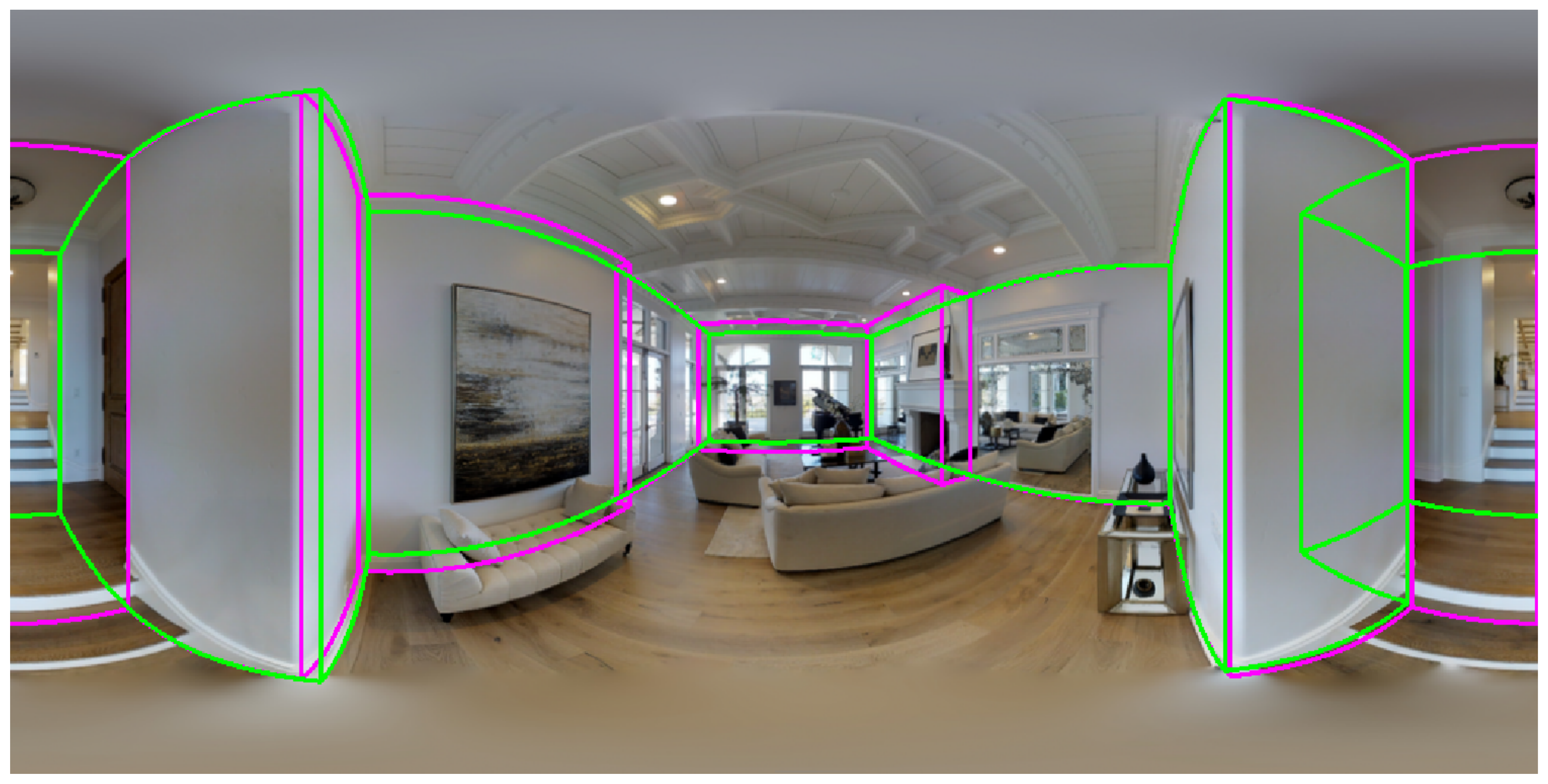}
        }
    \end{minipage}%
    \hfill
    \begin{minipage}[t]{0.5\textwidth}
        \subfloat{%
            \includegraphics[height=0.75\textwidth]{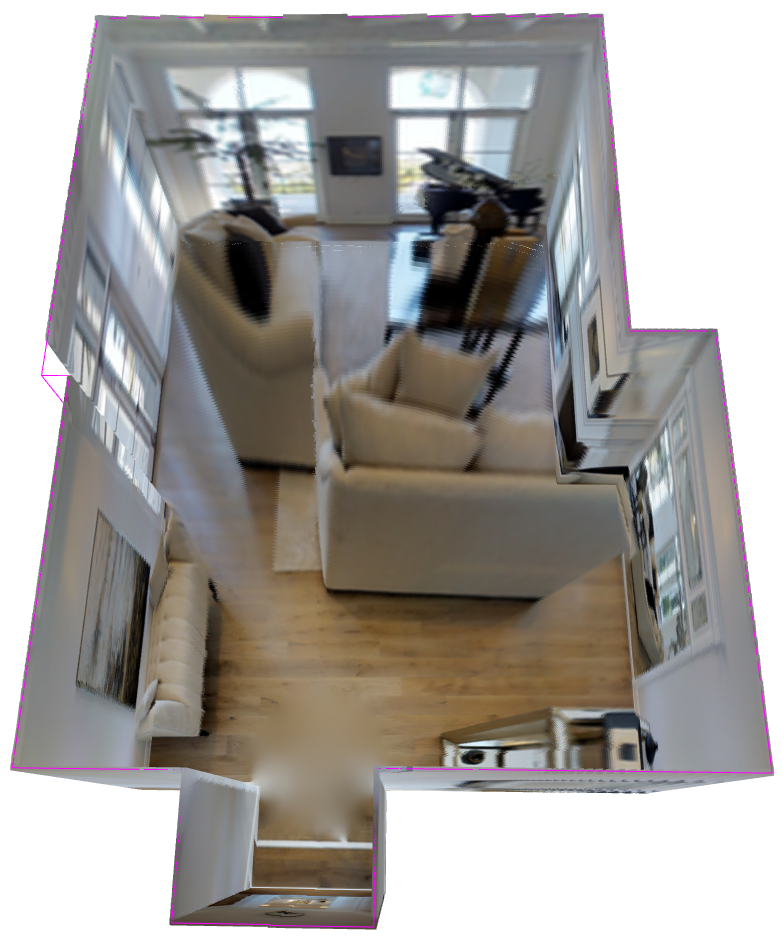}
        }
    \end{minipage}%
    \\[1ex]
    \begin{minipage}[t]{0.4\textwidth}
        \subfloat{%
            \includegraphics[width=\columnwidth]{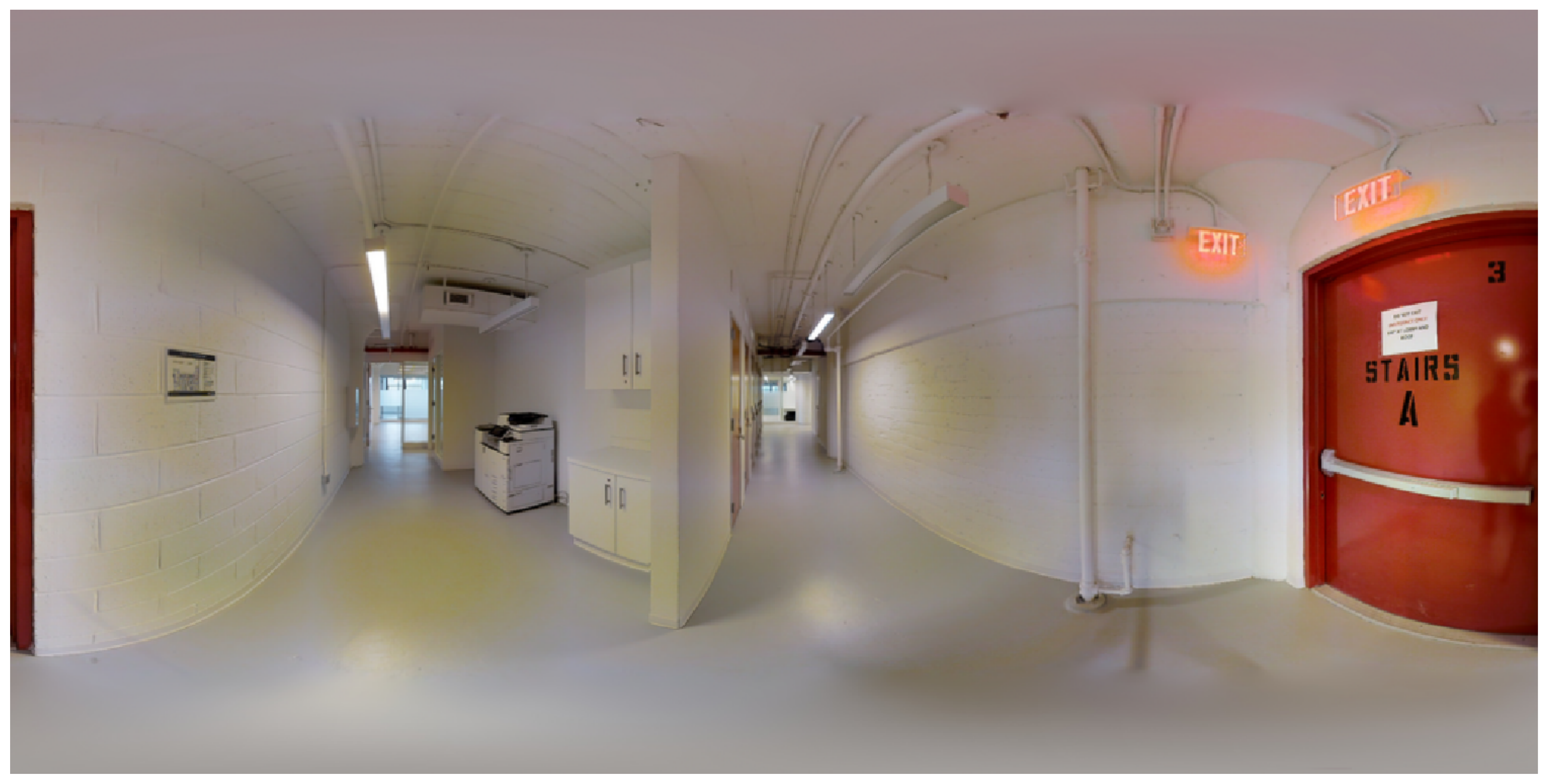}
        }
        \\[-3ex]
        \subfloat{%
            \includegraphics[width=\columnwidth]{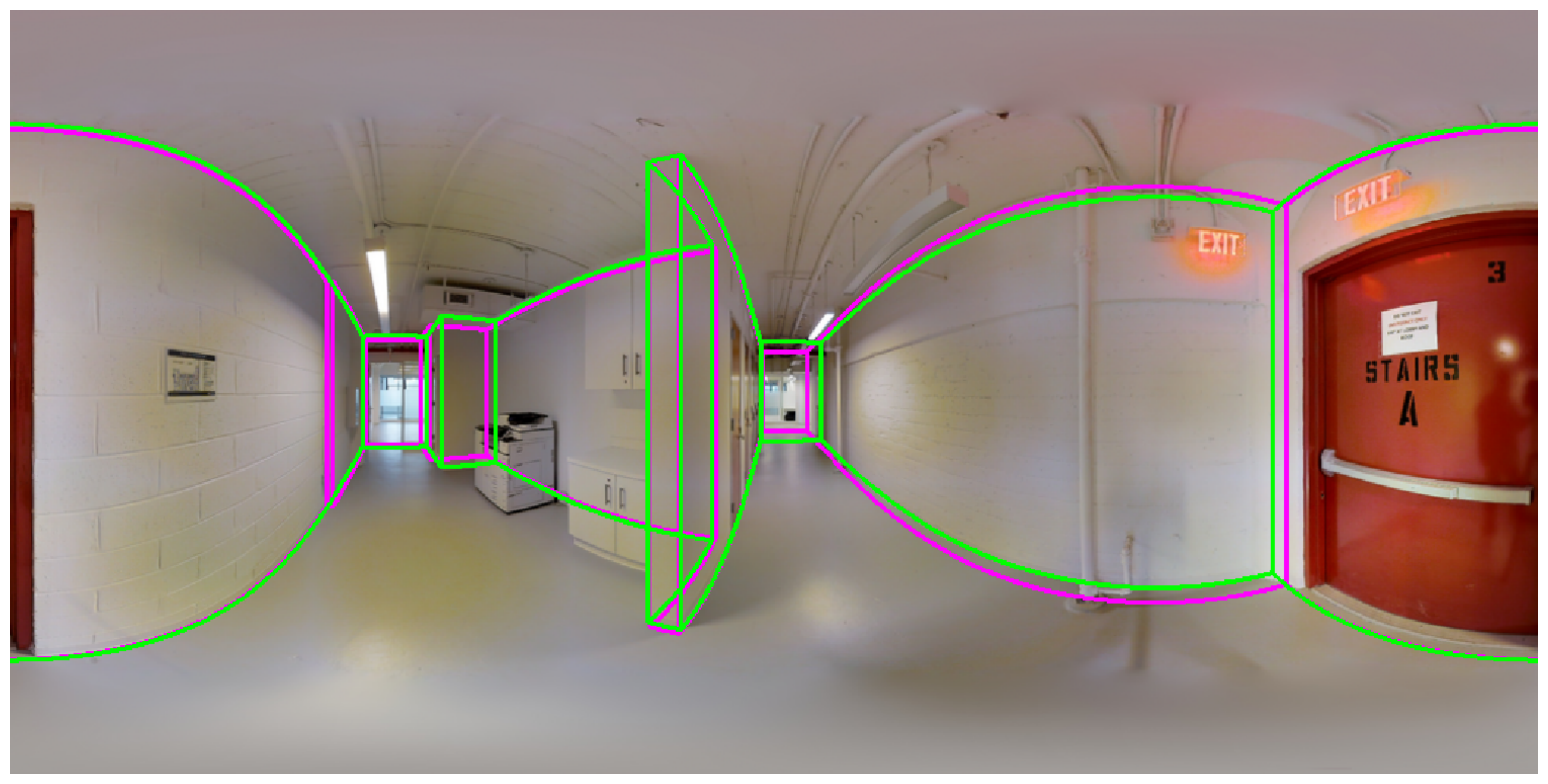}
        }
    \end{minipage}%
    \hfill
    \begin{minipage}[t]{0.5\textwidth}
        \subfloat{%
            \includegraphics[height=0.75\textwidth]{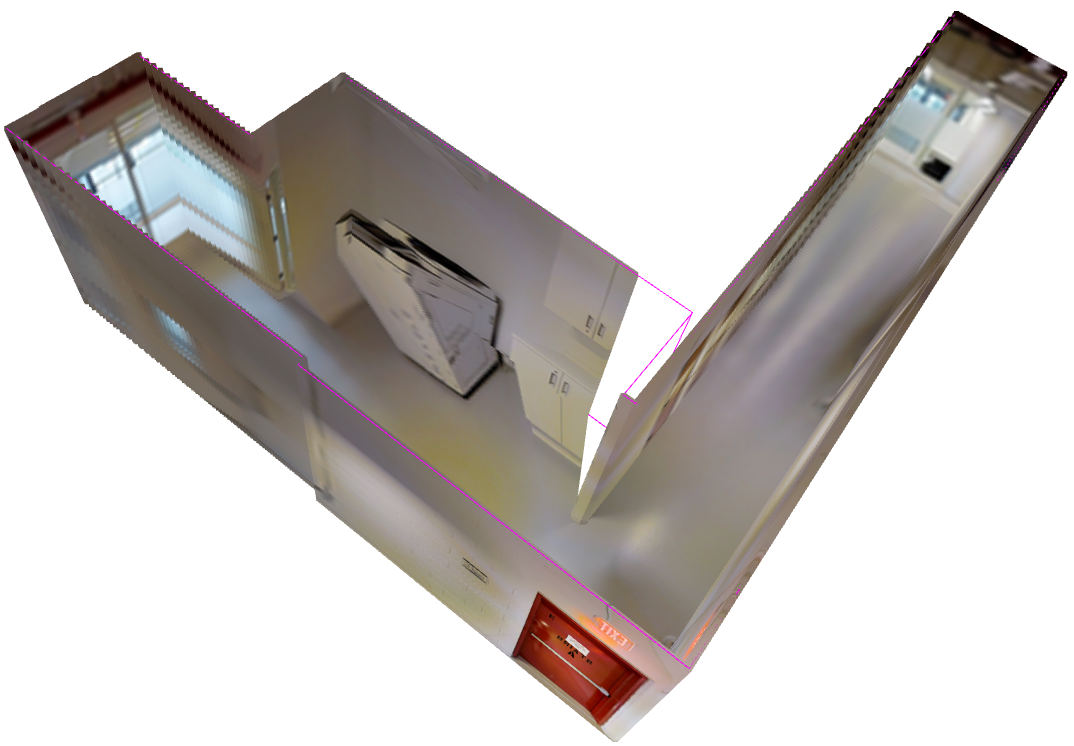}
        }
    \end{minipage}%
    \caption{Qualitative layout estimation of exemplar MatterportLayout test instances. \textbf{Left:} A comparison of ground truth layout (green lines) with our SSLayout360 model (magenta lines) trained on 100 labeled and 4,000 unlabeled images under equirectangular view. \textbf{Right:} 3D layout reconstruction. The transparent regions denote walls hidden from the camera field-of-view. Best viewed electronically.}
    \label{qualitative03}
\end{figure*}

\end{document}